\documentclass{article}

\usepackage{arxiv}

\usepackage[utf8]{inputenc} 
\usepackage[T1]{fontenc}    
\usepackage{hyperref}       
\usepackage{url}            
\usepackage{booktabs}       
\usepackage{amsfonts}       
\usepackage{amssymb}
\usepackage{nicefrac}       
\usepackage{microtype}      
\usepackage{lipsum}
\usepackage{subcaption}
\usepackage{graphicx}
\usepackage{makeidx}         
\usepackage[bottom]{footmisc}

\usepackage{glossaries}
\usepackage{array}
\usepackage{multirow}
\usepackage{color, colortbl}
\usepackage[utf8]{inputenc}
\usepackage[table]{xcolor}
\usepackage[colorinlistoftodos,prependcaption,textsize=medium]{todonotes}

\newcommand{\urlDate}{(accessed January 12, 2019)}

\definecolor{LightGray}{gray}{0.9}

\title{Nutty-based Robot Animation - Principles and Practices}

\author{
  Tiago Ribeiro\thanks{\protect\url{www.tiagoribeiro.pt}} \\
  INESC-ID \& \\Instituto Superior T\'{e}cnico\\
  University of Lisbon\\
  Portugal \\
  \texttt{me@tiagoribeiro.pt} \\
   \And
 Ana Paiva \\
  INESC-ID \& \\Instituto Superior T\'{e}cnico\\
  University of Lisbon\\
  Portugal \\
  \texttt{ana.paiva@inesc-id.pt} \\
}

\begin{document}
\maketitle

\begin{abstract}
Robot animation is a new form of character animation that extends the traditional process by allowing the animated motion to become more interactive and adaptable during interaction with users in real-world settings.
This paper reviews how this new type of character animation has evolved and been shaped from character animation principles and practices.
We outline some new paradigms that aim at allowing character animators to become robot animators, and to properly take part in the development of social robots.
One such paradigm consists of the 12 principles of robot animation, which describes general concepts that both animators and robot developers should consider in order to properly understand each other.
We also introduce the concept of Kinematronics, for specifying the controllable and programmable expressive abilities of robots, and the Nutty Workflow and Pipeline.
The Nutty Pipeline introduces the concept of the Programmable Robot Animation Engine, which allows to generate, compose and blend various types of animation sources into a final, interaction-enabled motion that can be rendered on robots in real-time during real-world interactions.
The Nutty Motion Filter is described and exemplified as a technique that allows an open-loop motion controller to apply physical limits to the motion while still allowing to tweak the shape and expressivity of the resulting motion.
Additionally, we describe some types of tools that can be developed and integrated into Nutty-based workflows and pipelines, which allow animation artists to perform an integral part of the expressive behaviour development within social robots, and thus to evolve from standard (3D) character animators, towards a full-stack type of robot animators.

\end{abstract}


\newacronym{hri}{HRI}{Human-Robot Interaction}
\newacronym{cgi}{CGI}{Computer-Graphics}
\newacronym{ai}{AI}{artificial intelligence}
\newacronym{ik}{IK}{inverse kinematics}
\newacronym[longplural={Degrees of Freedom}]{dof}{DoF}{Degree of Freedom}

\section{Introduction}
Robots becoming a new form of animated characters, which are jumping out of the big screens, powered by \gls*{ai}, and are becoming more interactive, and part of people's daily life.
They are being developed in order to be used in social applications,  in fields such as education, entertainment or assisted living.
Given the technological background required for the creation of such characters, they are being developed by roboticists, software engineers and \gls*{ai} scientists, instead of by artists.

While has been a necessary to go through a technically-oriented initial stage, 
we believe it is now time for robots and animated characters to reunite, 
by allowing artists and robot developers to work together, side by side, 
on the development of such characters. 
Animation artists have already been providing a contributing voice in the development of expressive, emotional and design traits of robots.
However they typically get little to no access to the development of the actual interactive and intelligent behaviors that are performed with humans.

The goal of our work is to establish a solid bridge between these two worlds, which are intrinsically connected, but have been evolving separately, based on different perspectives, fundamental competencies, and end-goals.
Such a connection will allow animators to take a new role as artists that are fully part of, and not just accessory, to the development of social robotic products.
The same happened upon the emerging of computer animated cartoons and in particular, of 3D animated characters.
At that time, animators exploring the new technique also felt the need to look into what had already been done during the last decades, and discover how that knowledge could be adapted for computer animation.
On that topic, Lasseter argued that the traditional principles of animation have a similar meaning across different animation medium \cite{Lasseter1987}.
Not only were those principles transferred to 3D animated characters, but new tools and methodologies were also created to support the creative and development processes.
Establishing robot animation as the new character animation medium will therefore require not only new theories, but also the integration of the technology with new tools and practices.

Bringing the illusion of life into actual robots in real-world situations, however, requires more than a set of guidelines such as the ones presented principles of character animation and of robot animation.
It requires proper technology, integration and tools that allow developers and animators to create such a robot together, and to make it act properly, according to its purpose.
In the past we have already delved into the use of character animation principles with robots \cite{Ribeiro2012}, and through the years we have been part of teams that have created various autonomous social robots in different applications (e.g.  \cite{Leite2008a,Pereira2014,Bernardo2016,Faria2016a,Petisca2016,Ribeiro2016,Alves-Oliveira2017,Correia2017,Ribeiro2017,Melo2018}).
Based on the lessons learned through such experiences, we outline various factors that have implications on the workflow and technical implementation of systems that put robot animation, as we describe it, into practice.

\section{Character Animation}
Disney's twelve principles of animation are considered by most to be the commandments of animation.
They are a result of more than 60 years of Disney productions, and were compiled into a book called 'The Illusion of Life', by Thomas and Johnston \cite{thomas1995illusion}, the last two of Disney's Nine Old Men\footnote{A group of nine animators that worked closely with Walt Disney since the debut feature Snow White and the Seven Dwarfs (1937) and onto The Fox and The Hound (1981).}.

For reference, we present a small summary of the original Twelve Principles of Animation defined in 'The Illusion of Life' \cite{thomas1995illusion}.

\begin{description}
\item[\textbf{Squash and Stretch}] states that characters should not be solid. 
The movement and liquidness of an object reflects that the object is alive, because it makes it look more organic. 
If we make a chair squash and stretch, the chair will seem alive. 
One rule of thumb is that despite them changing their form, the objects should keep the same volume while squashing and stretching.

\item[\textbf{Anticipation}] reveals the intentions of the character, so we know and understand better what they are going to do next.

\item[\textbf{Staging}] is the way of directing the viewers attention.
It is generally performed by the whole acting process, and also by camera, lights, sound and effects. 
This principle is related to making sure that the expressive intention is clear to the viewer. 
The essence of this principle is minimalism, keeping the user focused on what is relevant about the current action and plot.

\item[\textbf{Follow-Through and Overlapping Action}] are the way a character, objects or part of them inertially react to the physical world, thus making the movements seem more natural and physically correct. 
An example of Overlapping action would be hair and clothes that follow the movement of a character. 
Follow-through action is for example the inertial reaction of a character that throws a ball. 
After the throw, both the throwing arm and the whole body will slightly swing and tumble along the throwing direction.

\item[\textbf{Straight Ahead Action and Pose-to-Pose}] is about the animation process. 
An animator can make a character go through a sequence of well defined poses connected by smooth in-betweenings (Pose-to-Pose action), or sequentially draw each frame of the animation without necessarily knowing where it is heading (Straight-Ahead action).

\item[\textbf{Slow In and Slow Out}] is how the motions are accelerated (or slowed down).
Characters and objects do not start or stop abruptly. 
Instead, each movement has an acceleration phase followed by a slowing down phase, unless it is clearly intended not to.
Slow out can be confused with follow-through; however, follow-through extends the action, while the slow-out finishes it smoothly. 

\item[\textbf{Arcs}] draw the trajectories of natural motions, making them feel less machine-like and more natural and organic. 
An example is a head that gazes from left to right. 
A typical robotic movement would make the head rotate only along its vertical axis. 
A natural movement will make the head slightly lean up or down towards the midpoint of the trajectory while rotating.

\item[\textbf{Secondary Action}] is an action that does not contribute directly to the expression of an action, but adds personality and life-likeness. 
An example would be breathing, blinking the eyes, or holding and scratching different parts of the body.

\item[\textbf{Timing}] is a dual principle that focuses especially on two different things. 
First, it can change how users perceive the emotion of a motion or the physical world in which the character exists. 
Second, it also relates to the story, and how the story is being told. 
It is about how the character pauses between the actions, and how it synchronizes to itself and the surroundings.

\item[\textbf{Exaggeration}] makes some features more wild and relevant, and is what makes the characters behave as cartoons, as opposite to the dull motion of humans in the real world. 
An example would be popping out the eyes when startled, or growing a huge red tomato-like head while shouting.

\item[\textbf{Solid Drawing}] is about correctly balancing volume and weight of characters and objects. 
It also warns against symmetric characters and expressions. 
Characters do not stand stiff and still, unless that is what they are intended to portray.

\item[\textbf{Appeal}] of a character is how it expresses and asserts its role, personality and relevance in a story.
It is possibly the most subjective principle, as it also relates to how the character can make the viewers believe in its story. 

\end{description}

\subsection{Animation Curves}

Animation Curves are tools that are particularly important for animators. 
An animation curve exists for each \gls*{dof} that is being animated in a character, and it shows how that specific \gls*{dof} varies over time \cite{Roberts2004}. 

\begin{figure}[htbp]
\centering
\includegraphics[width=1.0\linewidth]{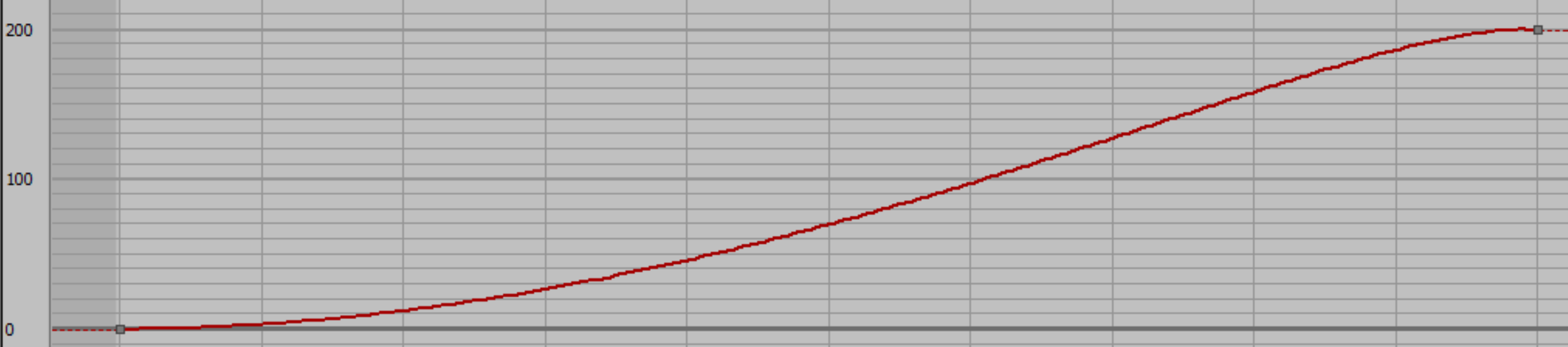}
\caption{The animation curve of the translation of a drag car accelerating until it reaches a top speed, and then decelerating until it halts. The vertical axis represents distance in generic units.}
\label{fig:curvesdragcar}
\end{figure}

Figure \ref{fig:curvesdragcar} shows the animation curve for the translation DoF of a hypothetical drag race car. 
In a drag race, the race car only drives forward at full speed. 
Because this animation curve shows the position changing over time, the speed of the car at some point of the curve is the tangent to the curve on that point (the first derivative). 
The second derivative (the rate of change of the tangent) thus represents the acceleration of the car.

By analyzing the curve, we see that the car starts by accelerating until about halfway through, when it reaches its maximum speed. We notice this because during the first part of the curve there is an accentuated concavity. 
Once the curve starts looking straight, the velocity is being kept nearly constant. 
In the end the car decelerates until it halts. 

\begin{figure}[htbp]
\centering
\includegraphics[width=1.0\linewidth]{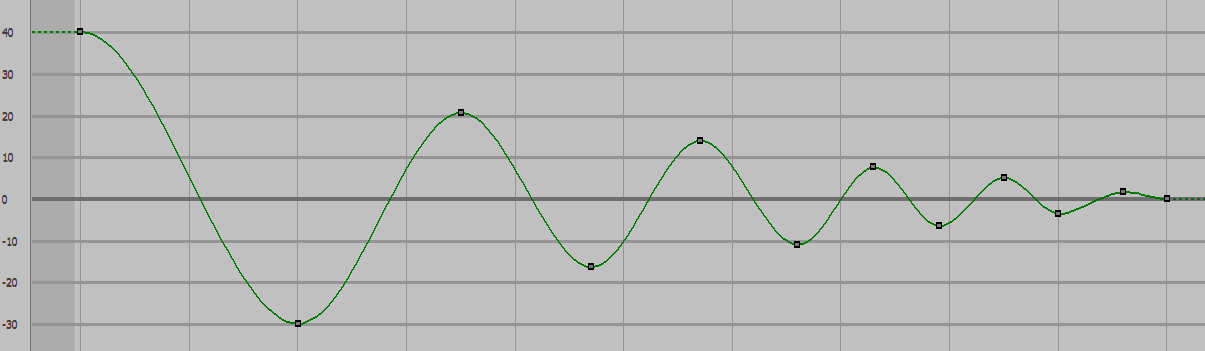}
\caption{The animation curve of the rotation of a pendulum that is dropped from 40 degrees and balances until it stops. The vertical axis represents the angle in degrees.}
\label{fig:curvespendulum}
\end{figure}

Animation curves can also be used to represent Rotation. 
Figure \ref{fig:curvespendulum} shows the animation curve of the rotation of the pivot of a pendulum that is dropped from a height of 40 degrees. 
It then balances several times while losing momentum due to friction and air resistance, until it stops.

In this curve we see some grey squares where the curve changes.
These squares are actually key-frames that were used to design the animation. 
The curve is a spline interpolation of the movement between these key-frames.

By looking at each key-frame, we see that the angle goes from 40 degrees to -30, then to about 20, and so on. 
Just like in the translation animation curve, the tangent of this curve also represents the velocity of rotation.

If we imagine the pendulum going through the lower-most position of its trajectory (which is the position in which it travels faster), that point would correspond to the 0 degrees line, thus making sense that each spline between two key-frames is steeper at this point, than closer to the key-frames.
As the pendulum loses energy and balances less, the steepness becomes lower, which reflects a lower speed, until it comes to a stop.

Animation curves therefore stand as a very important tool for representing, analyzing and adjusting animations. 
They can also be computationally processed just like a signal, in order to warp the animation and create animation effects.
More importantly, the animation curves represent a concept that both animators and engineers can understand, and can use it to connect their thoughts, requirements and obstacles.
Furthermore, they provide a technical interface that animators can use, and that can faithfully and mathematically model motion for robots.

\section{The Principles of Robot Animation}

In the context of social robotics, our understanding is that robot animation is not just about motion. 
It is about making the robot seem alive, and to convey thought and motivation while also remaining autonomously and responsive.
And because robots are physical characters, users will want to interact with them.
Therefore robot animation also becomes a robot's ability to engage in interaction with humans while conveying the illusion of life.

One of the major challenges of bringing concepts of character animation into \gls*{hri} is at the core of the typical animation process.
While in other fields, animation is directed at a specific story-line, timeline, and viewer (e.g. camera), in \gls*{hri} the animation process must consider that the flow and timeline of the story is driven by the interaction between users and the \gls*{ai},
and that the spacial dimension of the interaction is also linked to the user's own physical motion and placement.
Robot animation becomes intrinsically connected with its perception of the world and the user, given that it is not an absent character, blindly following a timeline over and over again.
This challenge is remarkable enough that character animation for robots can and should be considered a new form of animation, which builds upon and extends the current concepts and practices of both traditional and \gls*{cgi} animation and establishes a connection between these two fields and the field of robotics and \gls*{ai}.

Van Breemen initially defined animation of robots as 'The process of computing how the robot should act such that it is believable and interactive' \cite{Breemen2004a}. 
He also showed how 'Slow In/Out' could be applied to robots, although he called it Merging Logic.
We therefore complement Van Breemen's definition by stating that \textit{robot animation consists of the workflow and processes that give a robot the ability of expressing identity, emotion and intention during autonomous interaction with human users}. 

It is important to emphasize the word \textit{autonomous}, as we don't consider robot animation to be solely the design of expressive motion for robots that can be faithfully played back (that would fall into the field of animatronics).
Instead it is about creating techniques, systems and interfaces that allow animation artists to design, specify and program \textbf{\textit{how}} the motion will be generated, shaped and composed throughout an interaction, based on the behaviour descriptions that are computed by the \gls*{ai}.

A general list of Principles of Robot Animation should also address principles related to human-robot interaction. 
In our list however, we refrain from deepening such topic that is already subject of intensive study \cite{Murphy2010,Fong2003,Baxter2016,AlvesOliveira2016}.
Instead, we have looked into principles and practices of animators throughout several decades, and analysed how the scientific community can and has been trying to merge them into robot animation.

We have noted that not all principles of traditional animation can apply to robots, and that in some cases, robots actually reveal other issues that had not initially existed in traditional animation.
Most of these differences are found due to the fact that robots a) interact with people b) in the real, physical world.

The following sections reflect our understanding of how the Principles of Robot Animation can be aligned. 
Although they are stated towards robots, the figures presented show an animated human skeleton, as an easier depiction and explanation of use.
Each principle is also demonstrated on the EMYS and the NAO robots in an online video\footnote{\label{foot:poa}\protect\url{https://vimeo.com/49122495} \urlDate}, which can be watched as a complement to provide further clarification.
The video first demonstrates each principle using the same humanoid character presented in this section, and then follows with a demonstration of each principle first using the NAO robot, and then using the EMYS robot.

\subsection{Squash and Stretch}

For robots to use this principle, it sounds like the design of the robot must include physical squashing and stretching components.
However, besides relying on the design \cite{Hoffman2014,Suguitan2019}, we can also create a squash and stretch effect by using poses and body movement. 

\begin{figure}[htbp]
\centering
\includegraphics[width=1.0\linewidth]{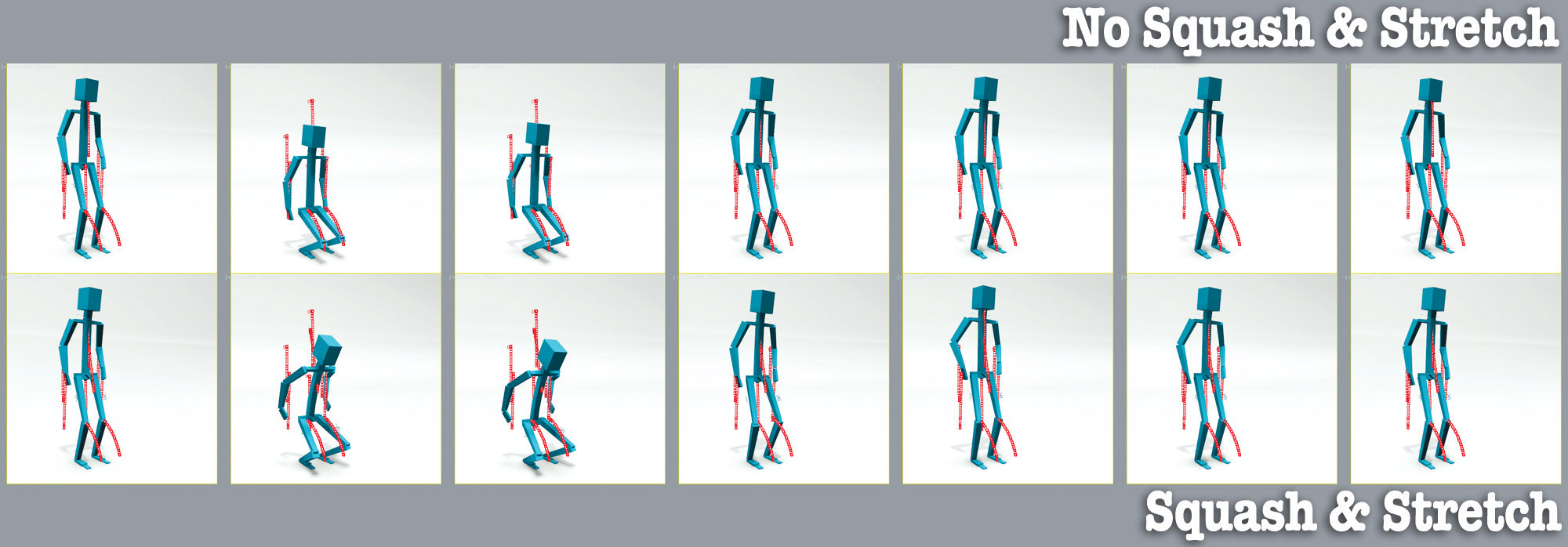}
\caption{An animation sequence denoting the principle of Squash \& Stretch. The red marks represent the trajectory of the most relevant joints.}
\label{fig:squast}
\end{figure}

\begin{figure}[htbp]
\centering
\includegraphics[width=0.8\linewidth]{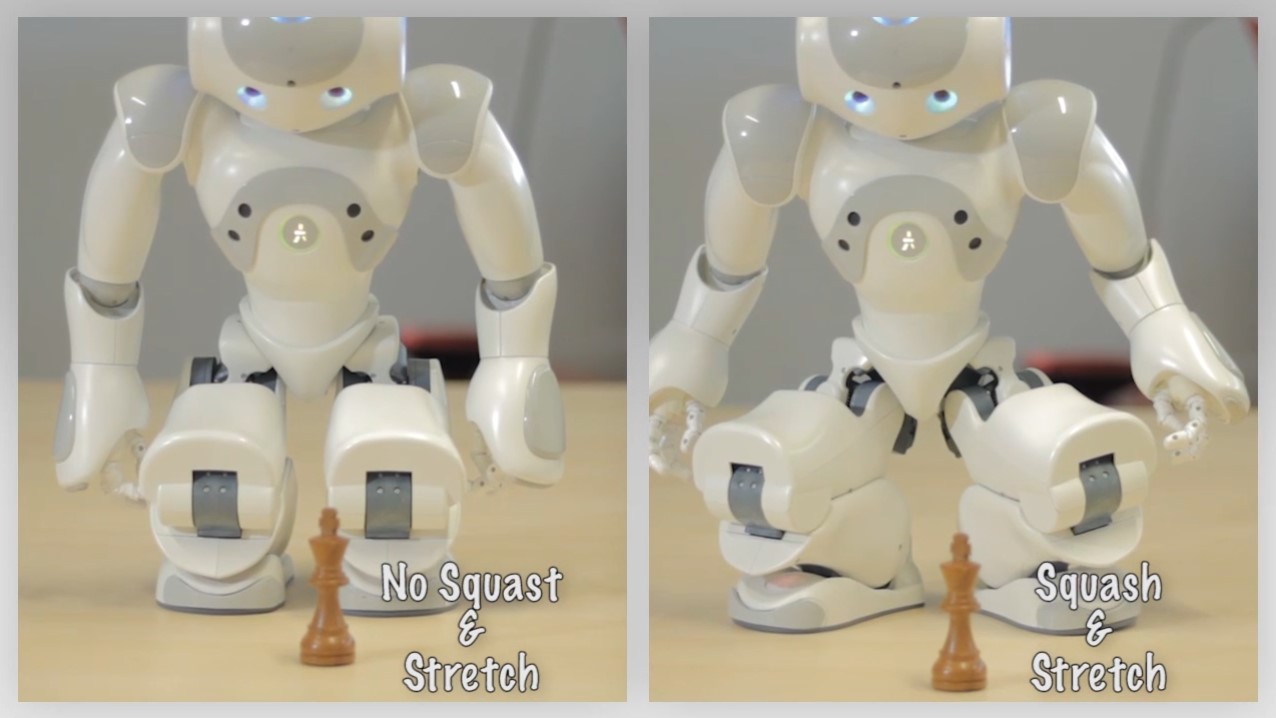}
\caption{The principle of Squash \& Stretch shown on the NAO robot.}
\label{fig:nao_squast}
\end{figure}

In Figure \ref{fig:squast} we can see how flexing arms and legs while crouching gives a totally different impression on the character.
Following the rule of constant volume, if the character is becoming shorter in height, it should become larger in length, and a humanoid robot can perform that by correctly bending its arms and legs.
Figure \ref{fig:nao_squast} presents a snapshot from the video\footnotemark[\getrefnumber{foot:poa}] illustrating how this principle looks like on the NAO robot.

\subsection{Anticipation}

Anticipating movements and actions helps viewers and users to understand what a character is going to do.
That anticipation helps the user to interpret the character or robot in a more natural and pleasing way \cite{Takayama2011}.

It is common for anticipation to be expressed by a shorter movement that reflects the opposite of the action that the character is going to perform.
A character that is going to kick a ball, will first pull back the kicking leg; in the same sense, a character that is going to punch another one will first pull back its body and arm. 
A service robot that shares a domestic or work environment with people can incorporate anticipation to mark, for example, that it is going to start to move, and in which direction, e.g., before picking up an object, or pushing a button.

\begin{figure}[htbp]
\centering
\includegraphics[width=1.0\linewidth]{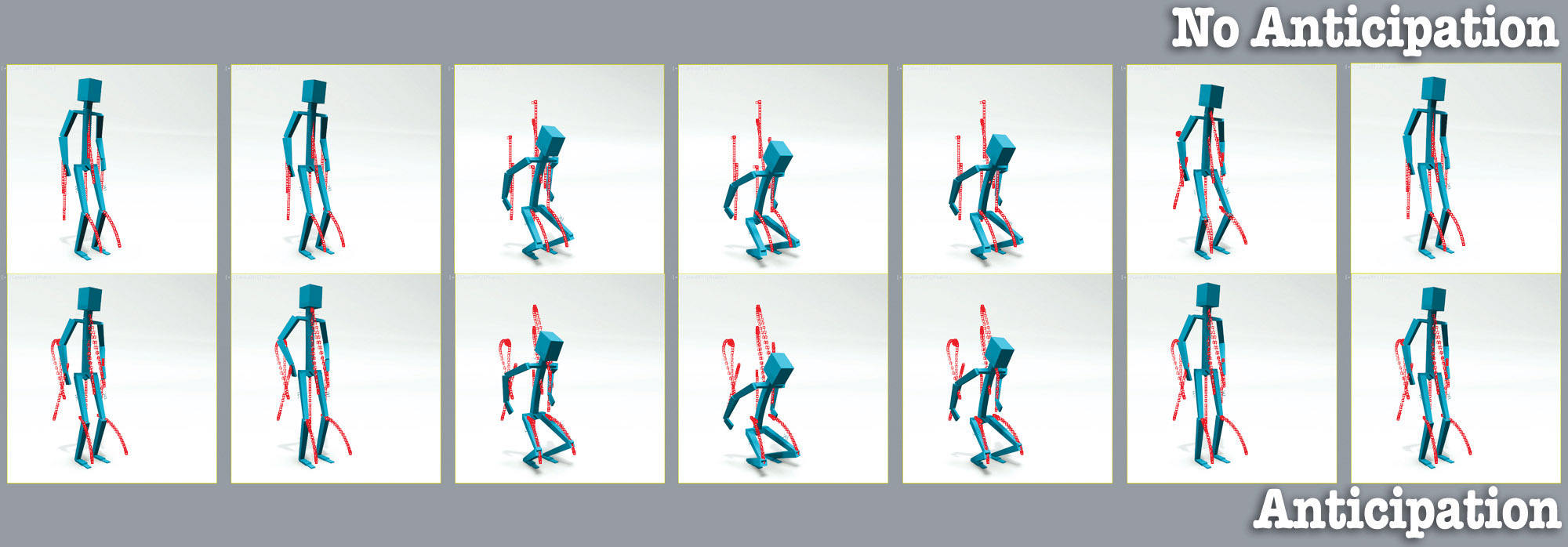}
\caption{An animation sequence denoting the principle of Anticipation. The red marks represent the trajectory of the most relevant joints.}
\label{fig:anticipation}
\end{figure}

In Figure \ref{fig:anticipation} we can see how a humanoid character that is going to crouch may first slightly stretch upwards. 

\begin{figure}[htbp]
\centering
\includegraphics[width=1.0\linewidth]{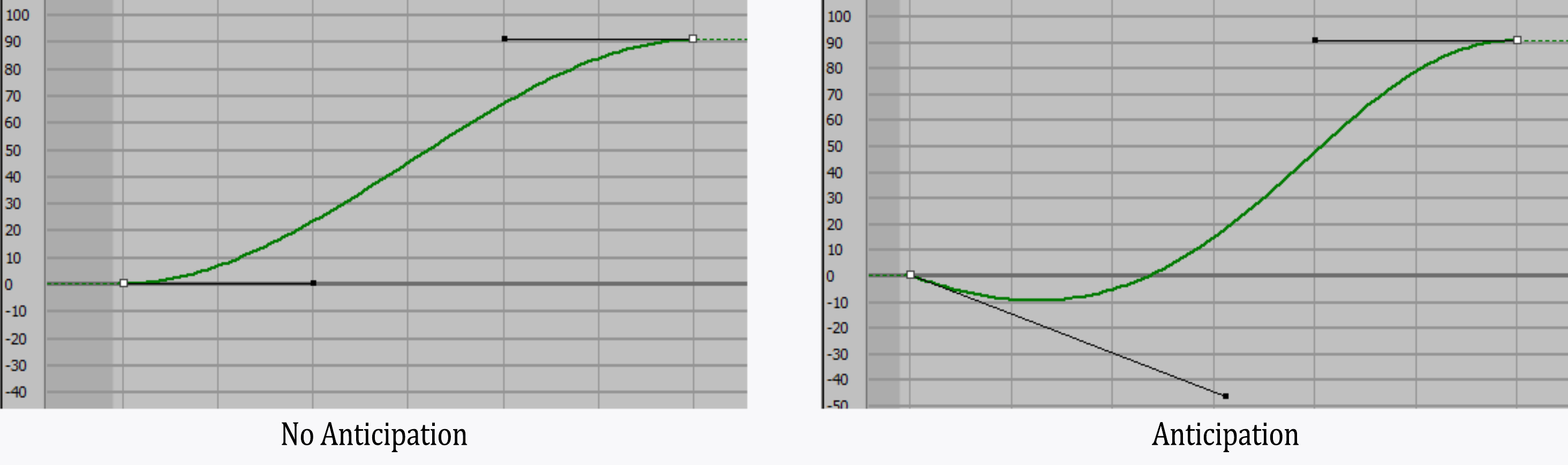}
\caption{Animation curves demonstrating anticipation. The left curve does not have anticipation; The right curve does.}
\label{fig:anticipationcurves}
\end{figure}

The concept can be better explained by looking at a simple animation curve example. 
Figure \ref{fig:anticipationcurves} shows two animation curves for a 90 degrees rotation of an object. 
On the left we see a simple animation curve, and at the start and end keyframes we see the tangent of the curve at that point.

On the right we have the same keyframes, but the tangent of the initial keyframe has been changed. 
Just by adjusting this tangent we have made the object start by slightly rotating 10 degrees backwards before performing the mentioned 90 degrees rotation, thus creating an anticipation effect.

\subsection{Intention}

This principle was formerly known as Staging in the traditional principles of animation.
In robots, staging results in several things.
First, it notes that sound and lights can carefully be used to direct the users' attention to what it is trying to communicate.
Second, if a robot is interested in, for example, picking up an object, it can show that immediately by facing such object \cite{Takayama2011}.
In either cases, the key here is showing the intention of the robot.

\begin{figure}[htbp]
\centering
\includegraphics[width=1.0\linewidth]{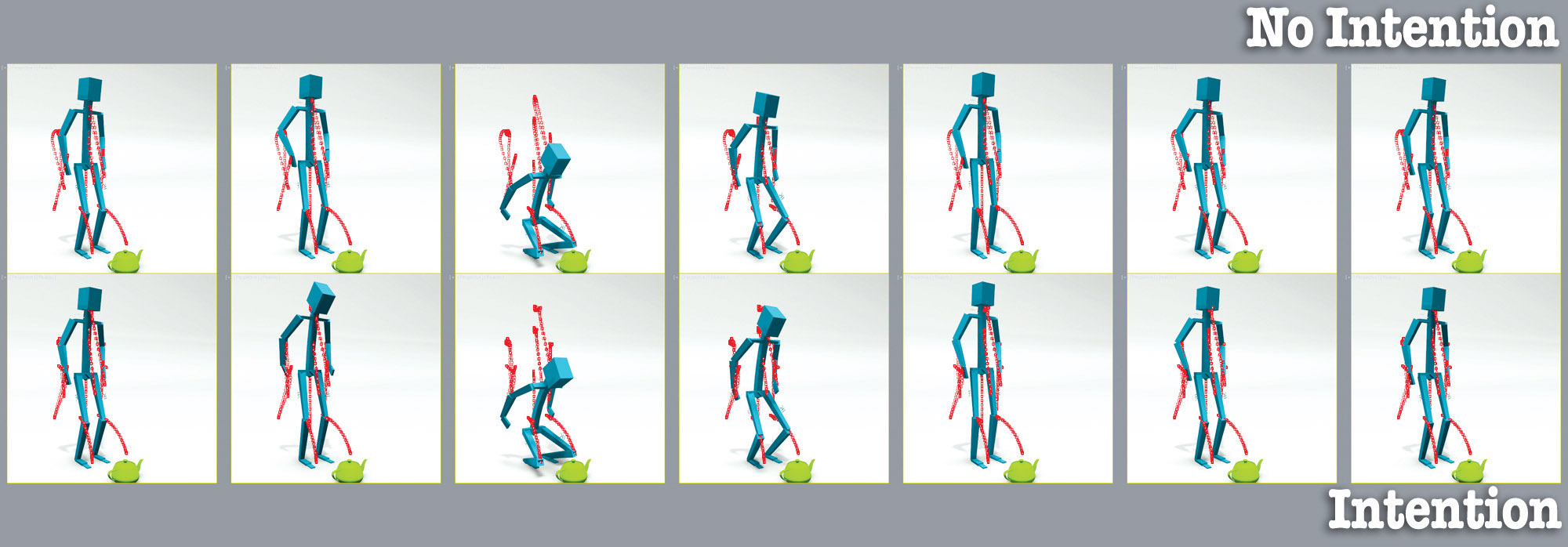}
\caption{An animation sequence denoting the principle of Intention. The red marks represent the trajectory of the most relevant joints.}
\label{fig:staging}
\end{figure}

We can see in Figure \ref{fig:staging} a simple idea of a humanoid character that is crouching over a teapot to eventually pick it up.
The character immediately looks at the teapot, so users know it is interested in it, and eventually guess that it is going to pick it up, much before the action happens.

That connects Intention with Anticipation; the difference is that while Anticipation should give clues about what the robot is going to do immediately, Intention should tell users about the purpose of all that he is doing, as a pre-action, before the actual action starts.
In a crouch-and-pick-up situation, for example, the robot will perform three actions - crouch, pick-up and stand.
We should see Anticipation for each of these actions.
The Intention, however, should reflect the overall of what the character is \textit{thinking} - it will start looking at the object even before crouching, and will start looking at the destination to where it will take the object even before starting to turn towards that direction.

\subsection{Animated, Procedural and Ad-hoc Action}

This principle was adapted from the Straight-Ahead and Pose-to-Pose action and has strong technical implications on the animation system development.
It originally talks about the method used by the animator while developing the animation.
Straight-ahead animation is used when the animator knows what he wants to do but has not yet foreseen the full sequence, so he starts on the first frame and goes on sequentially animating until the last one.
In pose-to-pose, the animator has pre-planned the animation and timing, so he knows exactly how the character should start and end, and through which poses it should go through.

In robots, this marks in the difference between playing a previously animated sequence, a procedural sequence, or an ad-hoc sequence.
As a principle of robot animation, it results in a balance between expressivity, naturalness and responsiveness.

A previously \textit{animated} sequence is self-explanatory.
It was carefully crafted by an animator using animation software, and saved to a file in order to be played-back later on.
That makes it the most common type of motion to be considered today in robot animation.
However it suffers from a lack of interactivity, as the trajectories are played-back faithfully regardless of the state of the interaction.
The motion is \textit{procedural} when it is generated and composed from a set of pre-configured motion generators (such as sine-waves).
On the other hand, it is \textit{ad-hoc} if it is fully generated in real-time, using a more sophisticated motion-planner to generate the trajectory (e.g. obstacle-avoidance; pick-and-place task).
We can say that playing an animation sequence that has previously been designed by an animator is a pose-to-pose kind of animation, while, for example, gaze-tracking a person's face by use of vision, or picking up an arbitrary object would be straight-ahead action.

A pose-to-pose motion can also contain anchor points at specific points of its trajectory (e.g. marking the beat of a gesture), so that the motion may be warped in the time-domain to allow synchronization between multiple motions. 
Those anchor-points would stand as if they were \textit{poses}, or key-frames in animation terms.
The concept of pose-to-pose can also become ambiguous in some case, such as in multi-modal synchronization, where, e.g. an ad-hoc gaze and an animated gesture should meet together at some point in time using anchor-points that define the meeting point for each of them.
In that case, the straight-ahead action, planned ad-hoc, can result in an animated sequence generated in real-time, and containing anchors placed by the planner.
From there it can be used as if it was a pose-to-pose motion to allow both motions to meet.

It currently sounds certain that the best and most expressive animations we achieve with a robot are still going to be pre-animated.
However the message here is that these different types of animation methods imply their own differences in the robotic animation system, and that such system should be developed to support them.

\begin{figure}[htbp]
\centering
\includegraphics[width=1.0\linewidth]{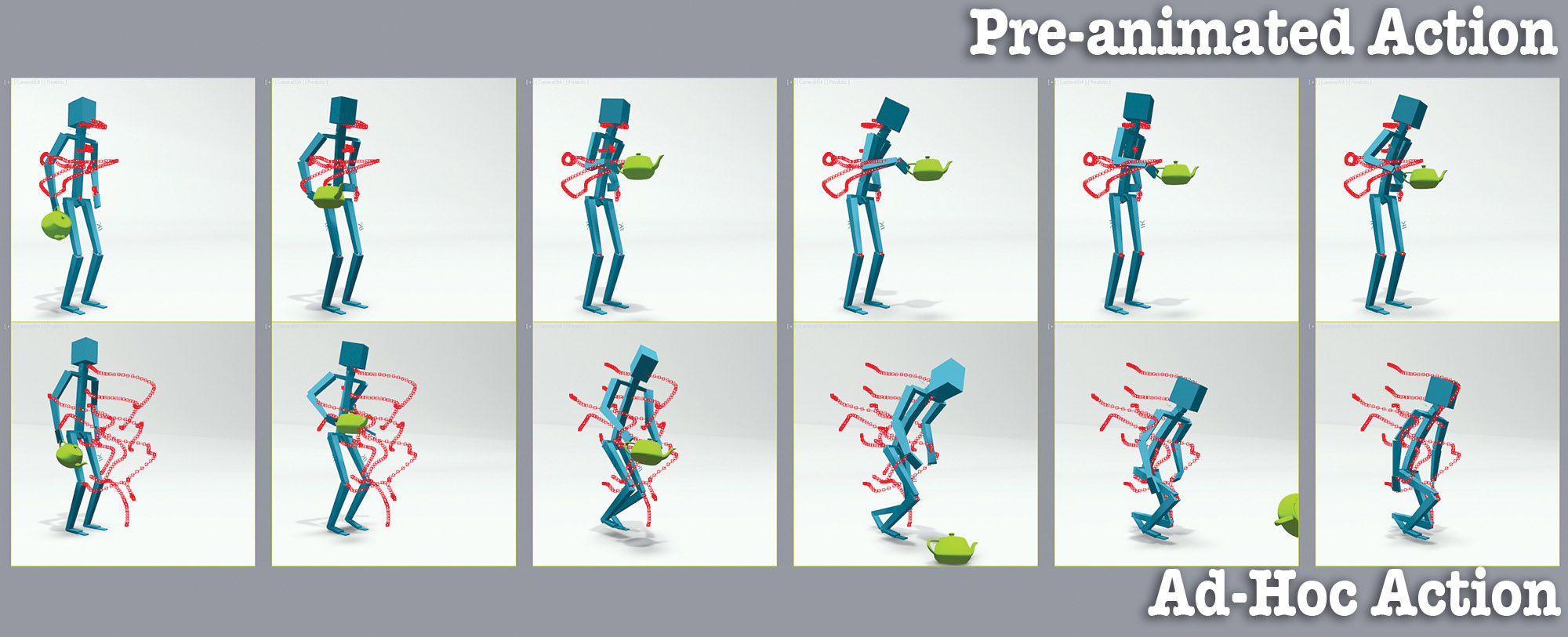}
\caption{An animation sequence denoting the principles of Pre-animated and Ad-hoc Action. The red marks represent the trajectory of the most relevant joints.}
\label{fig:straightahead}
\end{figure}

In Figure \ref{fig:straightahead} we can see on top a character performing a pre-animated and carefully designed animation, while in the bottom it is instantaneously reacting to gravity which made the teapot fall, and as such is performing an ad-hoc, straight-ahead animation.

While performing ad-hoc action, like reacting immediately to something, it might not be so important, in some cases, to guarantee principles of animation - if someone drops a cup, it would be preferable to have to robot grab it before it hits the ground, instead of planning on how to do it in a pretty way and then fail to grab it.
In another case, if a robot needs to abruptly avoid physical harm to a human, it is always preferable that the robot succeeds in whatever manner it can.
An ad-hoc motion planner therefore is likely to not contain many rules about animation principles, but act more towards functional goals (see the "Functional vs. Expressive Motion" section in \cite{Takayama2011}).

\subsection{Slow In and Slow Out}

For robot animation, Slow In and Out motion may me implemented within software in two different modalities: interpolation or motion filtering.

The former can be applied when the motion is either pre-animated, or fully planned before execution, so that the system has the full description of the trajectory points.
By tweaking the tangent type of the interpolation of the animation curve, it is possible to create accelerating and slowing down effects. 
By using a slow in and slow out tangent, the interpolation rate will slow down when approaching or leaving a key-frame.
This means that in order to keep timing unchanged, the rate of interpolation will have to accelerate towards the midpoint between two key-frames. 
Van Breemen called this Merging Logic and showed how it could be applied to the iCat \cite{Breemen2004a}.
In alternative, when the motion is generated ad-hoc, a feed-forward motion filter can be used to saturate the velocity, the acceleration and/or the jerk of the motion.

\begin{figure}[htbp]
\centering
\includegraphics[width=1.0\linewidth]{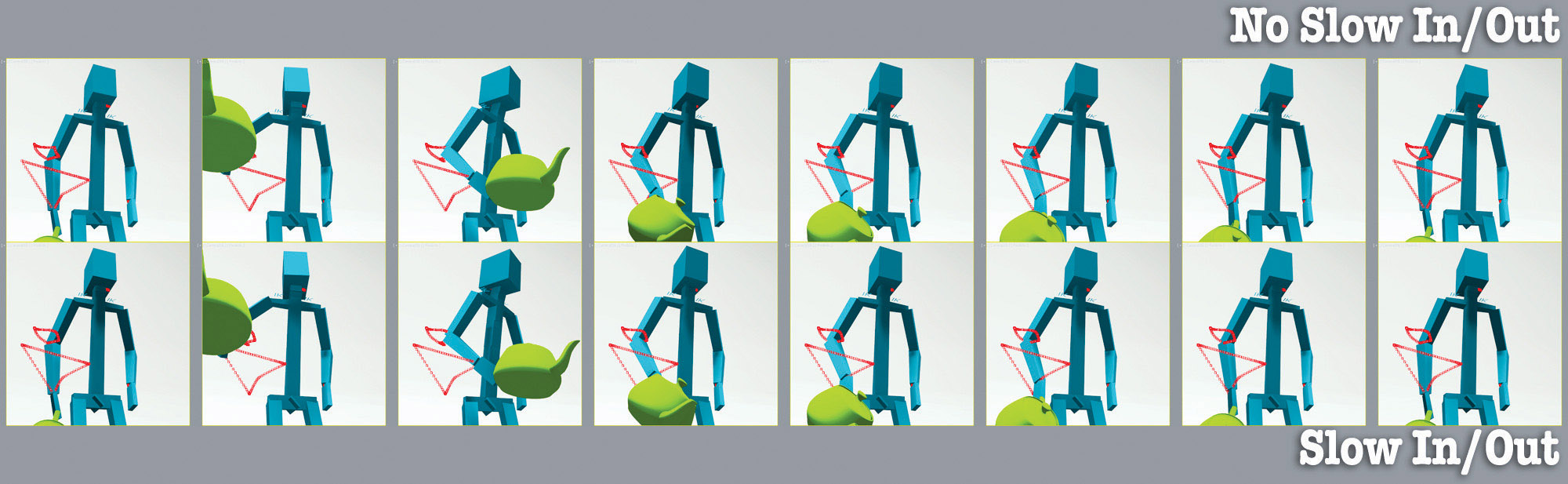}
\caption{An animation sequence denoting the principle of Slow In/Out. The red marks represent the trajectory of the most relevant joints. Notice how more frames are placed at the points of the trajectory where the motion changes in direction, in particular within the triangular-shaped portion. More spacing between points, using a fixed time-step, yields a faster motion.}
\label{fig:slowinout}
\end{figure}

A careful inspection of the red trajectories in Figure \ref{fig:slowinout} will show us the difference between the top animation and the bottom animation.
Each red dot represents an individual frame of the interpolated animation, using a fixed time-step.
We can see that in the bottom animation the spacing between the frames changes. 
It gathers more frames near the key-poses, and less between them. 
This causes the animation to have more frames on those poses, thus making it slow down while changing direction. 
Between two key poses the animation accelerates because the interpolation generated less frames there.

\begin{figure}[htbp]
\centering
\includegraphics[width=1.0\linewidth]{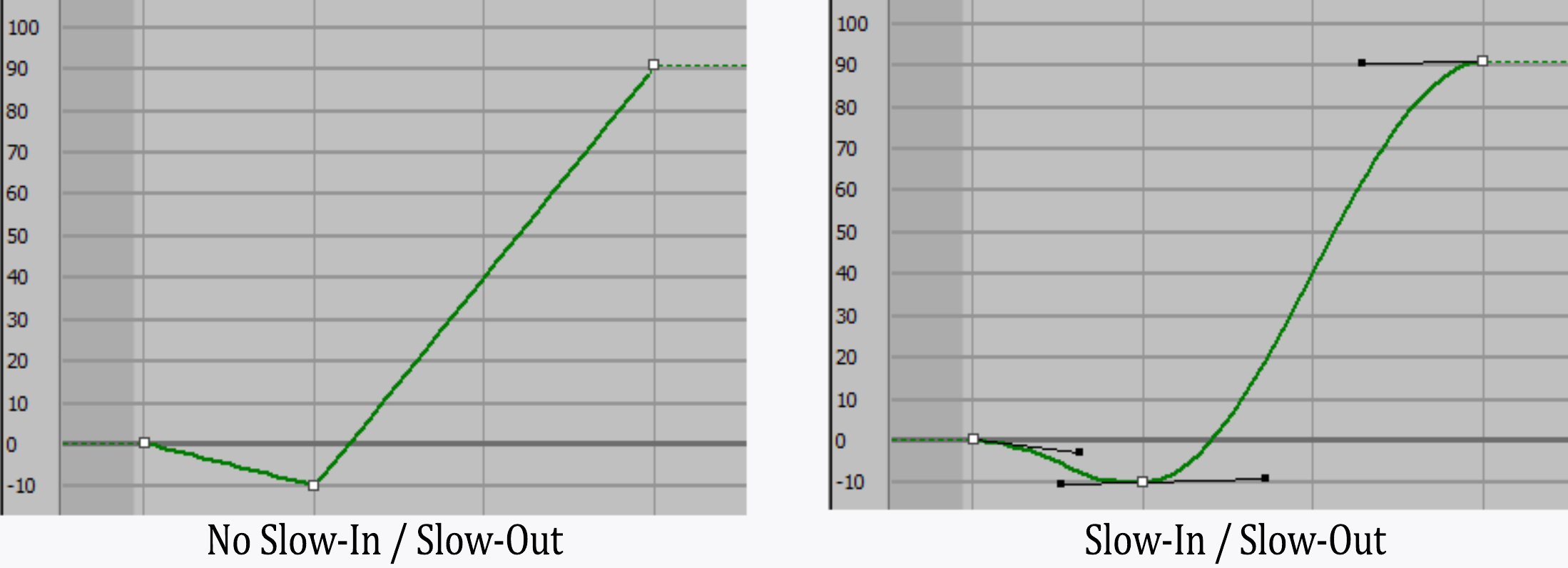}
\caption{Animation curves demonstrating Slow In and Slow-Out. The left curve does not have Slow In / Out; The right curve does.}
\label{fig:slowinoutcurves}
\end{figure}

This is more noticeable if we look at the animation curves. 
Figure \ref{fig:slowinoutcurves} shows a very simple rotation without Slow-In / Out (left) and with (right). 
In the left image we used linear tangents for the interpolation method, while in the right we used smooth spline tangents. 

We can see that with a linear interpolation, the curve looks straight, meaning that the velocity is constant during the whole movement. 
By using smooth tangents the movement both starts, stops and changes direction with some acceleration, which makes it look smoother.

\subsection{Arcs}

\begin{figure}[htbp]
\centering
\includegraphics[width=1.0\linewidth]{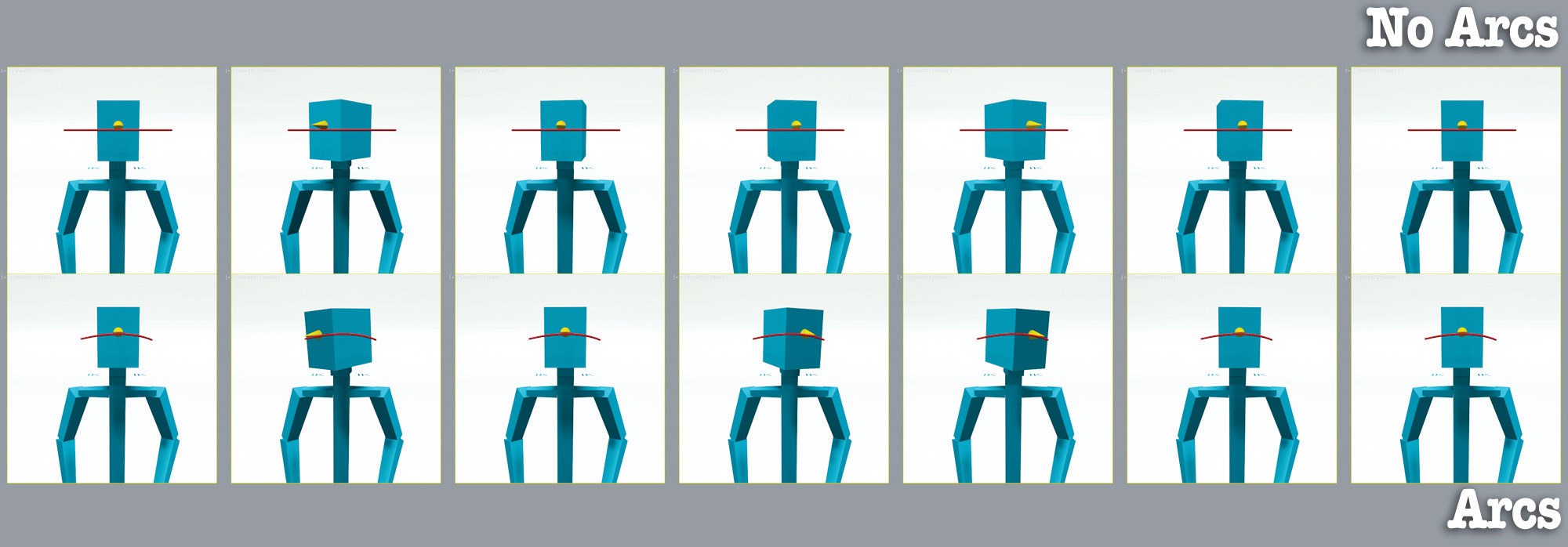}
\caption{An animation sequence denoting the principle of Arcs. The red marks represent the trajectory of the most relevant joints.}
\label{fig:arcs}
\end{figure}

Taking as example a character looking to the left and the right.
It shouldn't just perform a horizontal movement, but also some vertical movement, so that its head will be pointing slightly upwards or downwards while facing straight ahead. 
We can see that illustrated in Figure \ref{fig:arcs}.

This principle is easy to use in pre-animated motion.
However, in order to include it in an animation system, we would need to be able to know in which direction the arcs should be computed, and how wide the angle should be. 
If we have that information, then the interpolation process can be tweaked to slightly bend the trajectory towards that direction, whenever it is too straight.

What actually happens with robots is that depending on the embodiment, it might actually perform the arcs almost automatically. 
Taking as example a humanoid robot, when we create gestures for the arms, they will most likely contain arcs, due to the fact that the robot's arms are rigid, and as such, in order for the them to move around, the intrinsic mechanics will lead the hands to perform arched trajectories.
In traditional animation this principle was extremely relevant as the mechanics of the characters were not rigidly enforced as they are in robots.
Arcs still pose as an important principle to be considered in robot animation, both for pre-animated motions and also as a rule in expressive motion planners.

\begin{figure}[htbp]
\centering
\includegraphics[width=1.0\linewidth]{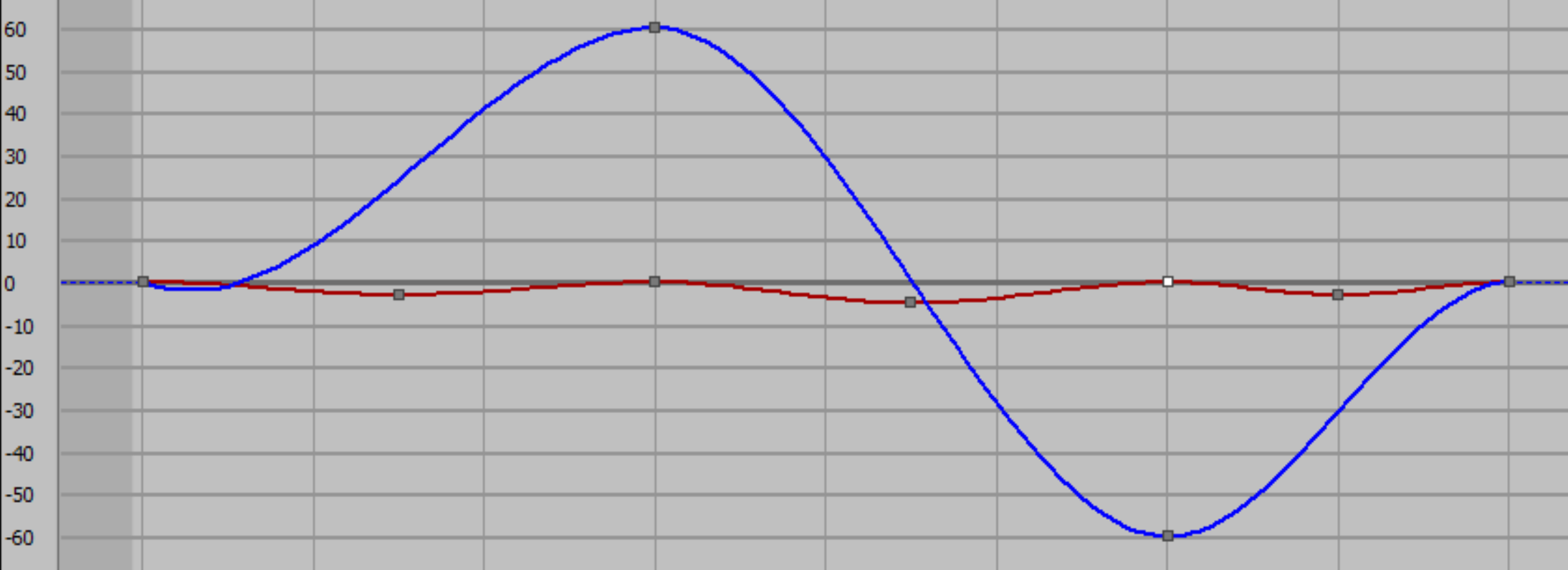}
\caption{Animation curves demonstrating Arcs. The blue curve is the Panning DoF, rotating from the rest pose, to its left (60 degrees) and then to its right (-60 degrees), and then back to rest. During this motion, the Pitch joint (red curve) slightly waves between those key-frames.}
\label{fig:arcscurves}
\end{figure}

Figure \ref{fig:arcscurves} shows a character gazing sideways.
The yellow cone represents the gazing direction at each frame.
The red curve illustrates the motion trajectory on the panning DoF (horizontally) and the Pitch DoF (vertically).
On the top motion, no movement is performed on the Pitch joint (straight line).
On the bottom motion, instead of performing only Yaw movement while looking around, the head also changes its Pitch between each keyframe of the Yaw movement.

\subsection{Exaggeration}

Exaggeration can be used to emphasize movements, expressions or actions, making them more noticeable and convincing. 
As such, it can also make robots seem more like actual characters and not just machines.

Although there are several levels of exaggeration, for robots it is interesting to look at exaggeration of actual movements.
It is actually a feature that can be implemented in animation systems by contrasting the motion signal \cite{Gielniak2012}.

\begin{figure}[htbp]
\centering
\includegraphics[width=1.0\linewidth]{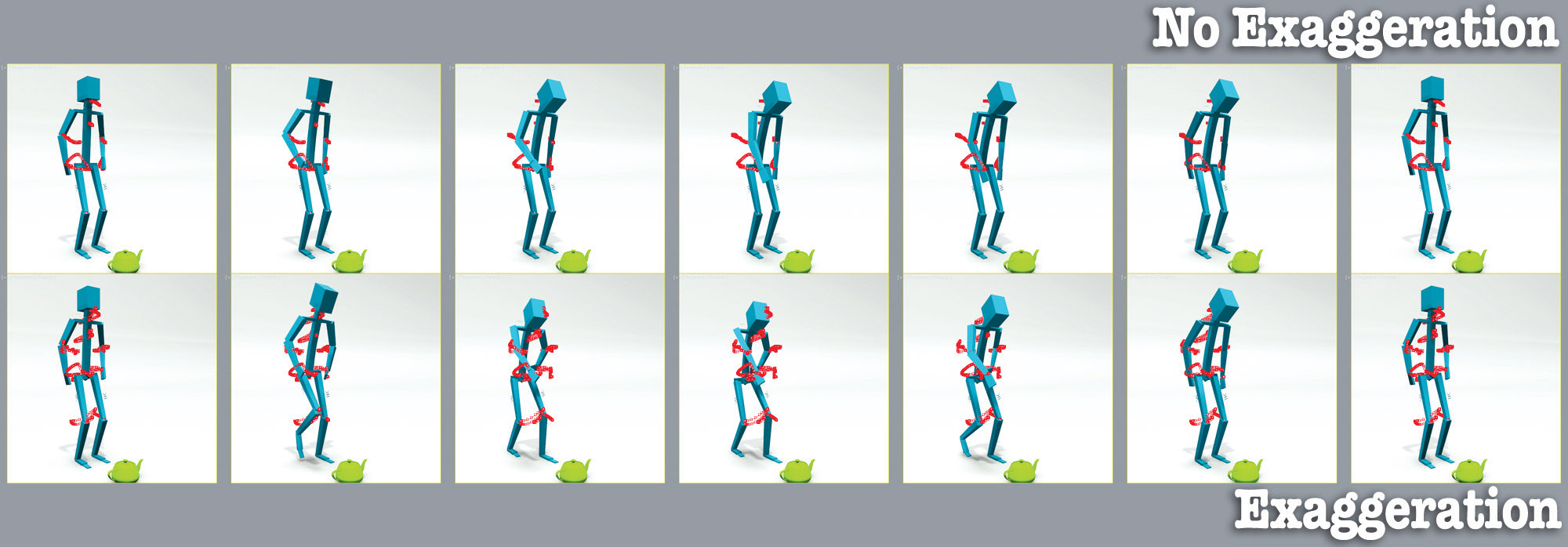}
\caption{An animation sequence denoting the principle of Exaggeration. The red marks represent the trajectory of the most relevant joints.}
\label{fig:exaggeration}
\end{figure}

Figure \ref{fig:exaggeration} shows not only an amplification of the most relevant features of an animation, but also an added feature - an 'anticipation' backward step.
This is meant to show that exaggeration can consist of more then just contrasting the signal, and that by exaggerating the anticipation we can also make the actual action seem more powerful.
Because this kind of practice may endanger the robot's surroundings and users if not correctly planned, it is recommended only within pre-animated motion, or for performance and entertainment robots in which the robot's surroundings and mechanical reach are guaranteed to be safe.

\begin{figure}[htbp]
\centering
\includegraphics[width=0.8\linewidth]{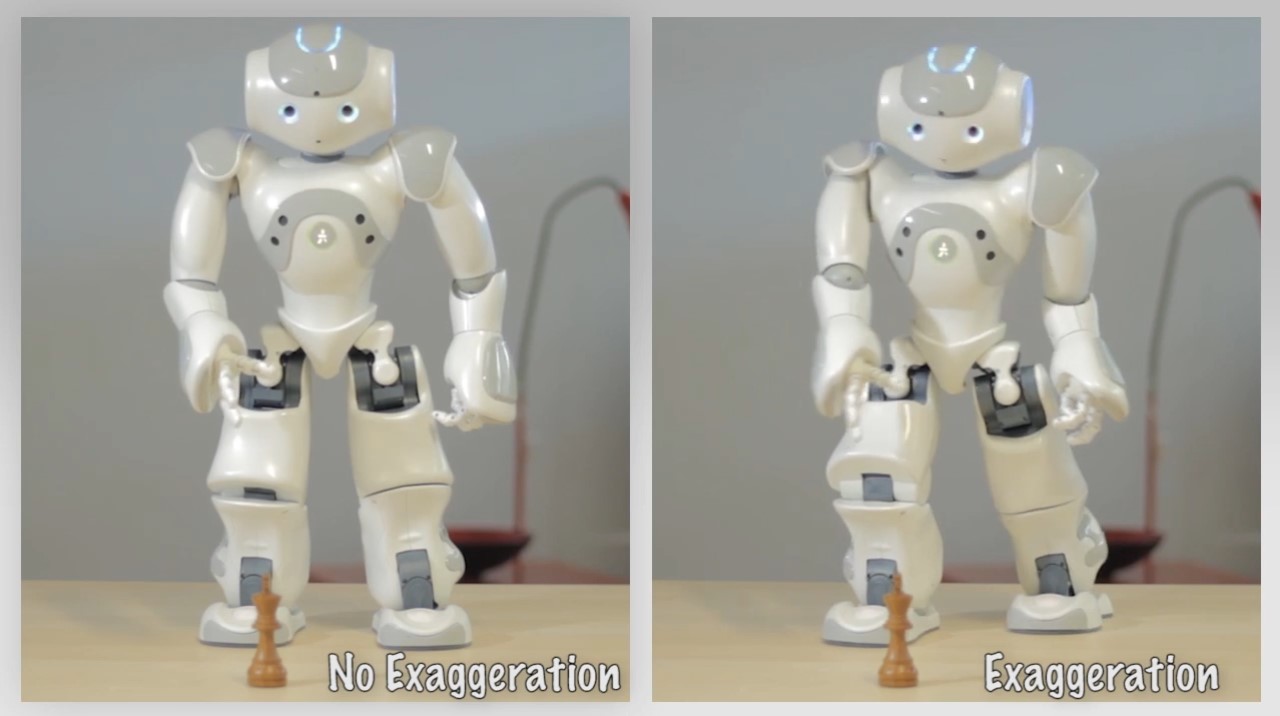}
\caption{The principle of Exaggeration exemplified on the NAO robot.}
\label{fig:nao_exagg}
\end{figure}

\begin{figure}[htbp]
\centering
\includegraphics[width=0.8\linewidth]{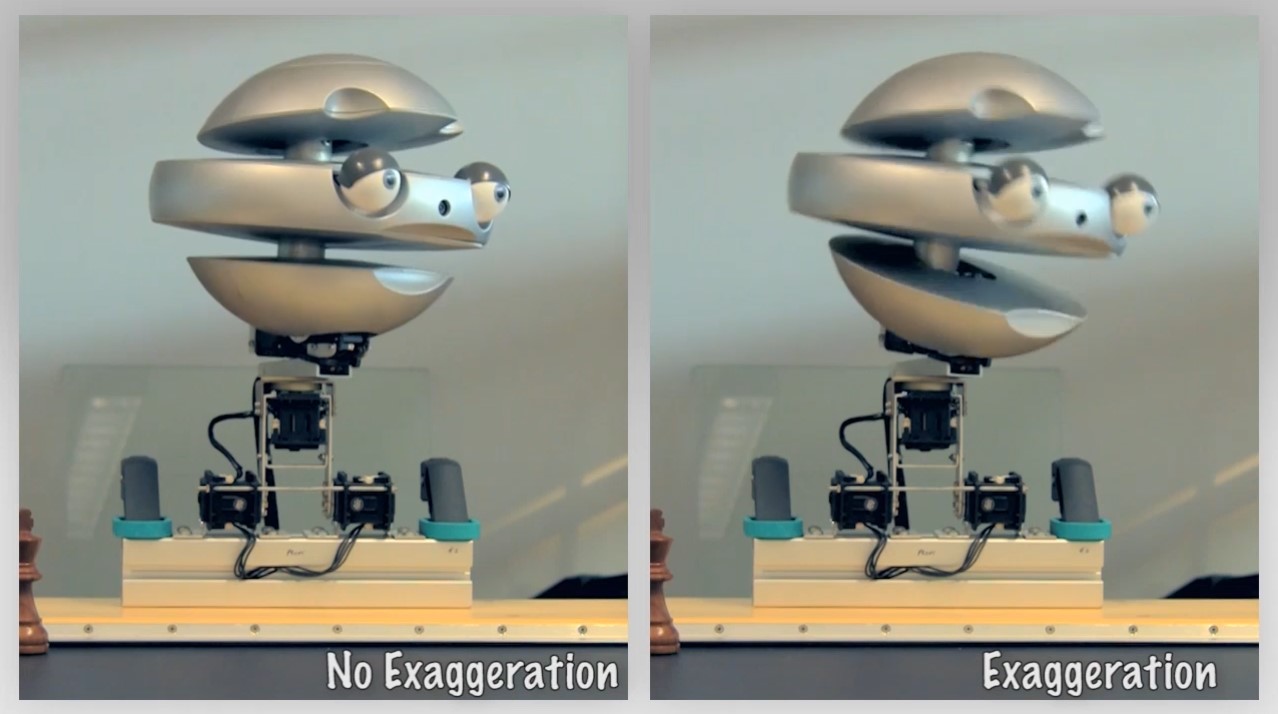}
\caption{The principle of Exaggeration exemplified on the EMYS robot.}
\label{fig:emys_exagg}
\end{figure}

Figure \ref{fig:nao_exagg} presents a snapshot from the video\footnotemark[\getrefnumber{foot:poa}] illustrating how this principle looks like on the NAO robot, while Figure \ref{fig:emys_exagg} show the same for the EMYS robot.

\subsection{Secondary Action and Idle Behavior}

\begin{figure}[htbp]
\centering
\includegraphics[width=1.0\linewidth]{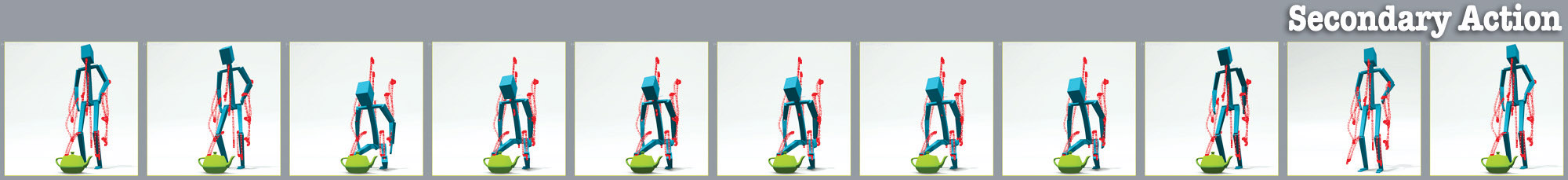}
\caption{An animation sequence denoting the principle of Secondary Action. The red marks represent the trajectory of the most relevant joints.}
\label{fig:secondary}
\end{figure}

During a conversation, people often scratch some part of their bodies, look away or adjust their hair.
In Figure \ref{fig:secondary} we can see a character that is crouching to approach the teapot, and in the meanwhile scratches its gluteus.
Using secondary action in robots will help to reinforce their personality, and the illusion of their life.

A character should not stand stiff and still, but should contain some kind of Idle motion, also known as \textit{keep-alive}.
Idle motion in robots can be implemented in a very simplistic manner.
Making them blink their eyes once and a while, or adding a soft, sinusoidal motion to the body to simulate breathing (lat. \emph{anima}) contribute strongly to the illusion of life.

In the case of facial idle behaviour such as eye-blinking, during a dramatic facial expression these will often go unnoticed or may even disrupt the intended emotion.
It is better to perform them at the beginning or end of such expressions, rather than during.
Similarly, blinking also works better if performed before and between gaze-shifts.

\subsection{Asymmetry}

This principle was derived from the traditional principle of Solid Drawing.
Although the traditional principle seemed not to relate with robots, it actually states some rules to follow on the posing of characters.

It states that a character should neither stand stiff and still, nor does it stand symmetrically.
We generally put more weight in one leg than on the other, and shift the weight from one leg to the other.
It also suggests the need for the idle behavior, and how it should be designed.

The concept of asymmetry stands both for movement, for poses and even for facial expression.
The only case in which we want symmetry is when we actually want to convey the feeling of stiffness.

\begin{figure}[htbp]
\centering
\includegraphics[width=1.0\linewidth]{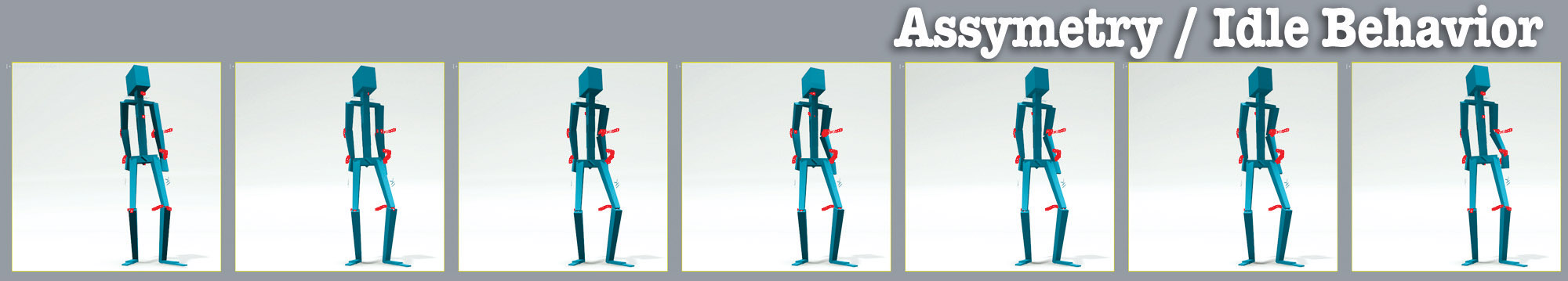}
\caption{An animation sequence denoting the principles of Asymmetry and Idle Behavior. The red marks represent the trajectory of the most relevant joints.}
\label{fig:solid}
\end{figure}

Figure \ref{fig:solid} shows a character portraying another Principle - Idle Behavior, while also standing asymmetrically.
This Idle Behavior is performed by the simulation of breathing and by slightly waving its arms like if they were mere pendulums.

\subsection{Expectation}

This principle was adapted from the original Appeal.
If we want a viewer or user to love a character, then it should be beautiful and gentle.
If we are creating an authoritative robot, it should have more dense and stiff movements.
Even if one wants to make viewers and users feel pity for a character (such as an anti-hero), then the character's motion and behaviour should generate that feeling, through clumsy and embarrassing behaviours.

\begin{figure}[htbp]
\centering
\includegraphics[width=1.0\linewidth]{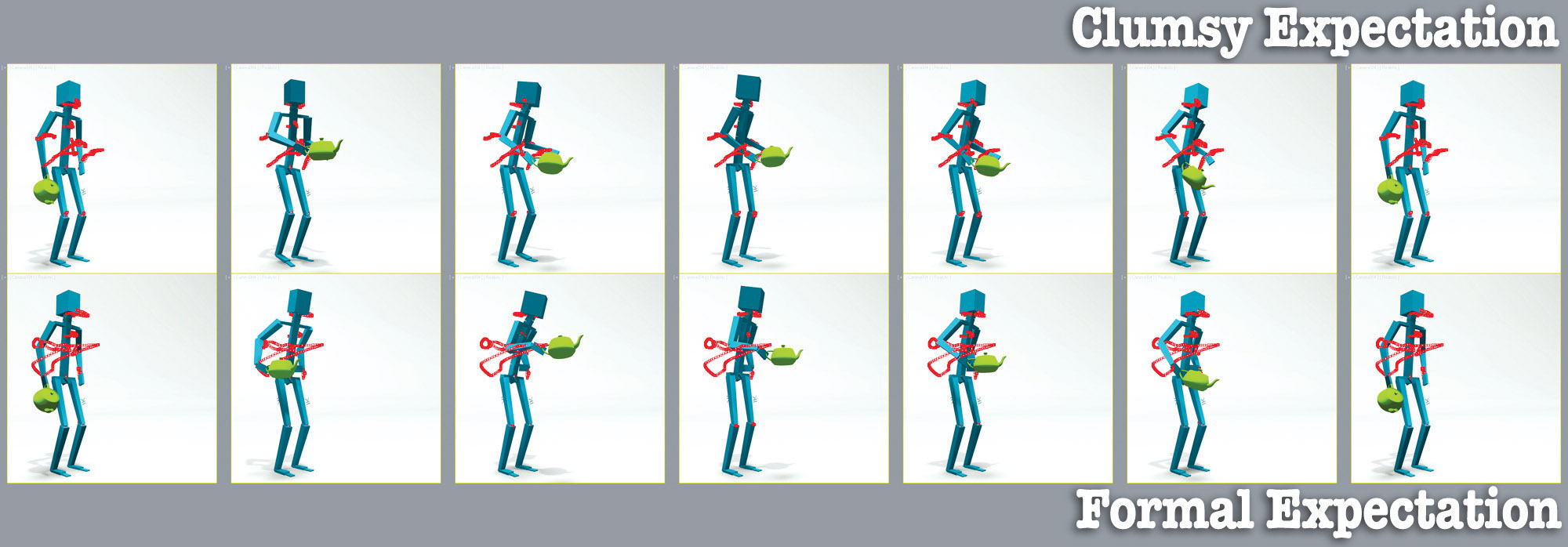}
\caption{An animation sequence denoting the principle of Expectation. The red marks represent the trajectory of the most relevant joints. Notice how the clumsy version balances the teapot around instead of holding it straight, and waves around its left ar instead of holding it closer to its body, delivering a feeling of discourtesy.}
\label{fig:appeal}
\end{figure}

Figure \ref{fig:appeal} shows two characters performing the same kind of behavior, but one of them is performing as a formal character like a butler, while the other is performing as a clumsy character like an anti-hero.
In this case the visual appearance of the character was discarded.
However, if we had a robotic butler, we would expect him to behave and move formally, and not clumsy.

The expectation of the robot drives a lot of the way users interpret its expression.
It relates to making the character understandable, because if users expect the robot to do something that it doesn't (or does something that they are not expecting) they fill fail to understand what they are seeing.

Wistort refers to Appeal as 'Delivering on Expectations' \cite{Wistort2010}, and his arguments have inspired us to agree.
He considers that the design and behavior of a robot should meet, so if it is a robotic dog, then it should bark and wag its tail.
But if it is not able to do that, then maybe it should not be a dog.
The Pleo robot\footnote{\protect\url{www.pleoworld.com} \urlDate} for example, was designed to be a toy robot for children.
So the design of it as a dinosaur works very good, as it does not cause any specific expectation in people - as people do not know any living dinosaurs, and as such, they don't know if Pleo should be able to bark or fetch, so they don't expectation him to be able to do any of that.

\subsection{Timing}

Timing can help the users to perceive the physical world to which the robot belongs.
If the movement is too slow, the robot will seem like it is walking on the moon. 

However, timing can also be used as an expression of engagement.
Some studies have revealed a correlation between acceleration and perceived arousal.
A fast motion often suggests that a character is active and engaged on what it's doing \cite{Saerbeck2010, Takayama2011}. 

\begin{figure}[htbp]
\centering
\includegraphics[width=1.0\linewidth]{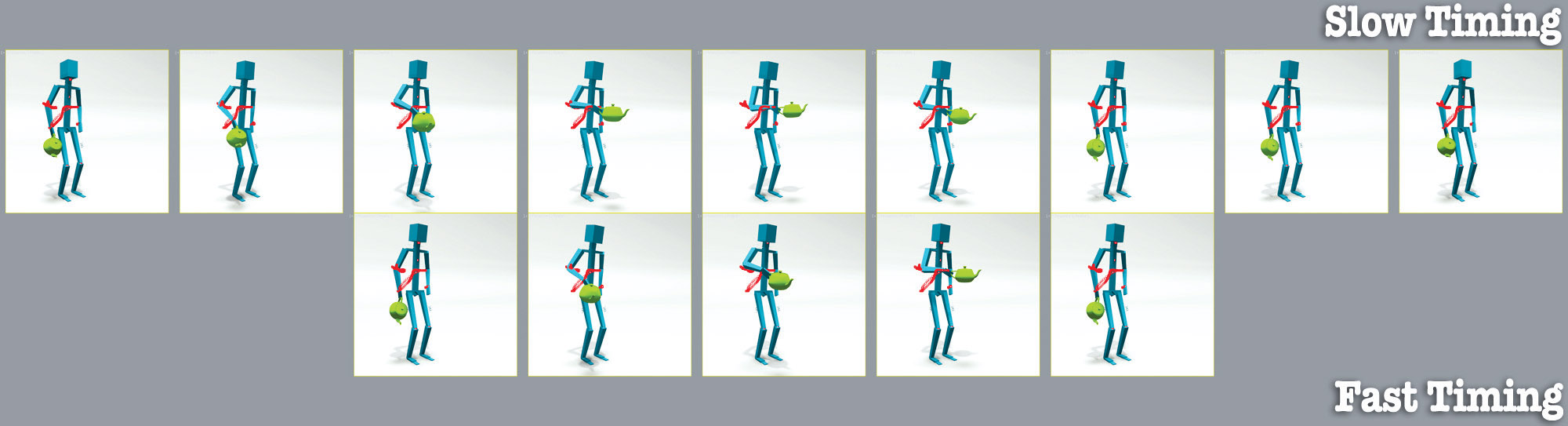}
\caption{An animation sequence denoting the principle of Timing. The red marks represent the trajectory of the most relevant joints.}
\label{fig:timing}
\end{figure}

Being able to scale the timing is useful to be able to express different things using the same animation, just by making it play slower or faster.
In Figure \ref{fig:timing} we get a sense that the top character is not engaged as much as the lower character, because we see it taking longer to perform the action.
It may even feel like the character is bored with the task.
In the fast timing case we are showing less frames of the same animation, to give the impression of it being performed faster.
In reality, that would be the result, as a faster paced animation would require less frames to be accomplished using a fixed time-step.

As a principle of robot animation, timing is something that should be carefully addressed when synthesizing motion e.g. using a motion-planner. 
Such synthesizer will typically solve for a trajectory that meets certain world-space constraints, while also complying with certain time-domain constraints such as the kinematic limits that the robot is allowed to perform.
In many cases, a very conservative policy is chosen, i.e., the planner is typically instructed to move the robot very slowly in order to keep as far away as possible from its kinematic limits.
However, such a rule may be adding some level of unwanted expressiveness to the motion.
We therefore argue that when using such planners it is important to consider, within the safety boundaries of the robot's kinematic limits, ways of generating trajectories that can exploit the time-domain in a more expressive way.

\subsection{Follow-Through and Overlapping Action}

This principle works like an opposite of anticipation.
After an action, there is some kind of reaction - the character should not stop abruptly.

We should start by distinguishing these two concepts here.
\textit{Follow-through} animation is generally associated with inertia caused by the character's movement.
An example of follow-through is when a character punches another one, and the punching arm doesn't stop immediately, but instead, even after the hit, both body and arm continue to move a bit due to inertia (unless it is punching an 'iron giant').
Overlapping is an indirect reaction caused by the character's action.
An example of overlapping is for example the movement of hair and clothes which follow and overlap the movement of the body.

Using follow-through with robots requires some precaution because we do not want the inertial follow-through to hurt a human or damage any other surroundings.
Follow-through might also cause a robot to loose balance, so it seems somewhat undesirable.
Many robot systems actually will try to defend themselves against the follow-through caused by its own movements, so why would we want it?

In first instance, we consider that follow-through should better not be used in most robots, especially for the first reason we mentioned (human and environment safety).
However, when it can be included at a very controlled level, namely on pre-animated motion, it might be useful to help mark the end of an action, and as such, to help distinguish between successive actions.
Unlike anticipation, however follow-through is much more likely to be perceived by humans as dangerous, because it can give the impression that the robot slightly lost control over its body and strength. 
We would therefore imperatively refrain from using it on any application for which the perception of safety is highest, such as in health-care or assistive robotics.

Overlapping animation depends mostly on the robot's embodiment and aesthetics.
It might serve as a tip for robot design, by including fur, hair or cloth on some parts of the robot, that can help to emphasize the movement \cite{Suguitan2019}.
As such, we find no need to include overlapping animation into the animation process of robots per se, because whatever overlapping parts that the robot might have, should be 'animated' by natural physics.
Therefore if one wishes to use it, it should be considered as an animation effect that is drawn by the design of the robot's embodiment, and thus should be developed initially at the robot design stage.
\section{Dimensions of Kinematronics}
We start by introducing the concept of \textit{Kinematronics}, which refers to all the high-level mechanical and electronic systems that allow a robot to portray animate expression either kinematically (through physical movement) or electronically (through screens, lights and sounds).
The term is derived from \textit{kinematics}, which would refer only to the mechanical, physical components, and is composed with the concept of "elec\textit{tronics}" to include the non-mechanical forms of expression.

\begin{figure}[htbp]
\centering
\includegraphics[width=1.0\linewidth]{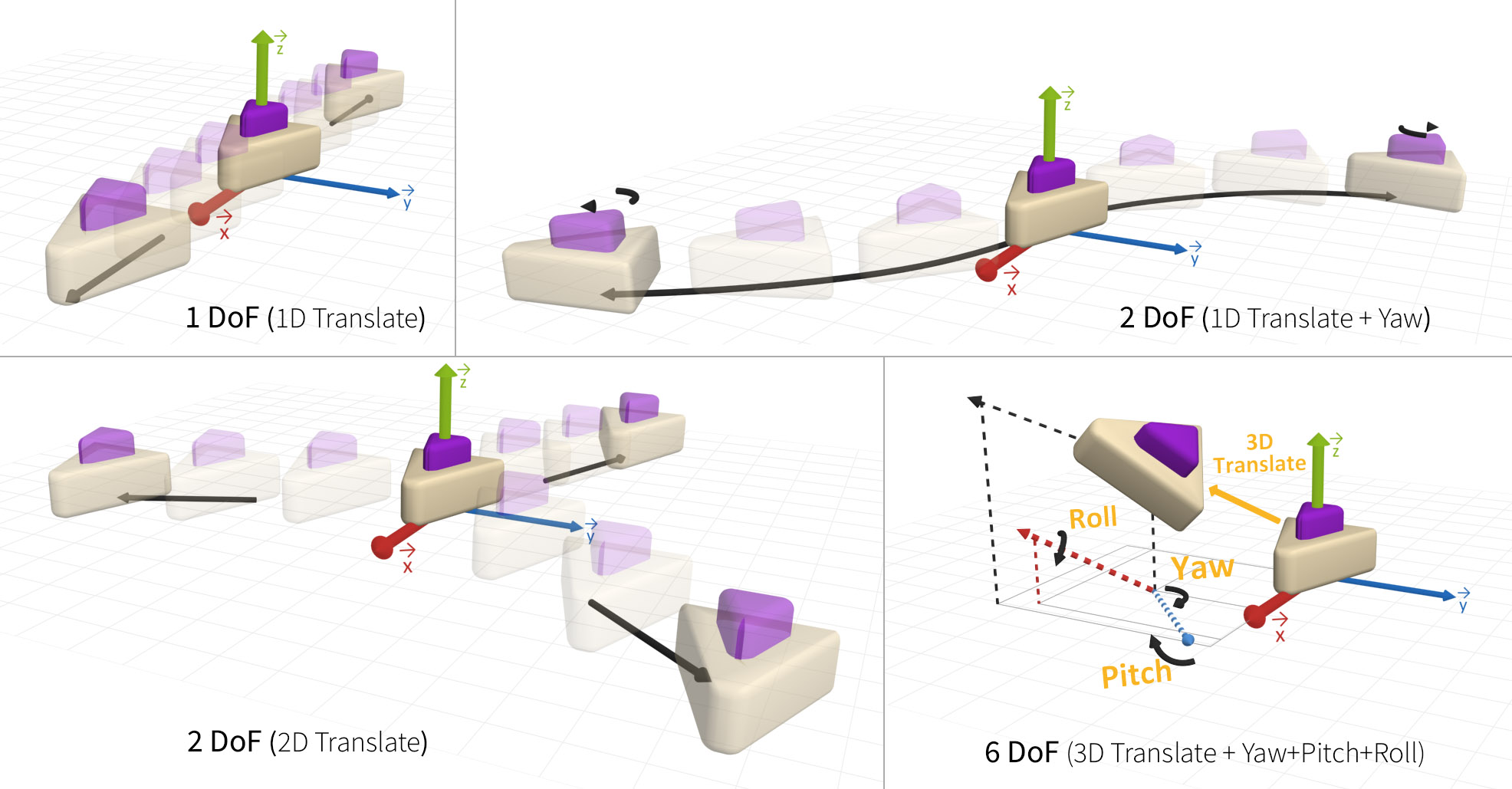}
\caption{The Spatial Expression of Kinematronics. \textbf{\textit{top-left}}: movement in a single direction represents 1 DoF. \textbf{\textit{top-right}}: movement in one translational direction plus one rotational direction (in this case, Yaw), amounts to  2 DoFs. \textbf{\textit{bottom-left}}: movement in two translational directions also amount to 2 DoFs. \textbf{\textit{bottom-right}}: the most complex 6-DoF example, in which translational movement can be performed in 3D, and rotational movement can also be performed along the three rotation axes. Using intrinsic rotations is recommended, i.e., the coordinate system for the rotations is attached to the moving body and therefore changes after each elemental rotation. Elemental intrinsic rotations are performed in the order Yaw-Pitch-Roll.}
\label{fig:spacialexpression}
\end{figure}

Robots may take many forms and shapes, and provide various means of both interacting with the world, and of conveying expressivity.
We start by defining an \textit{expressive \gls*{dof} (degree of freedom)}, further referred to merely as a \gls*{dof}, to be a one dimensional expressive channel that can be individually controlled through a given range or set of values.
Each expressive DoF in a robot can be controlled individually during interaction in order to convey a significant and intentional expression.
Based on the set and types of \glspl*{dof} a robot has, and their individual and aggregate role on providing expressivity, we have defined four different dimensions of kinematronics:
\begin{description}
\item[\textbf{Stationary Expression}] refers to motion performed by \glspl*{dof} that are purely mechanical and that do not yield any intentional movement in space.
Examples of such expressions are facial and postural expressions.
The term \textit{stationary} is chosen because these are mostly found in stationary robots such as desk-top robots, which do not move around by themselves.
We do include full-body postures into this type of expression, as long as they are not meant to move the robot in space.
An example of that would be a standing humanoid robot, which enacts a full-body emotive posture.
While its legs could be used for walking, i.e., for spatial function and expression, when they are used in non-locomotive expressions we do consider them to be acting as part of the stationary expression dimension.
\item[\textbf{Spatial Expression}] refers to motion that moves the robot around in space.
These are typically accomplished by either wheels or legs, but can also be performed by rotors, as in a quadcopter drone.
Note that besides walking, a legged robot, for example, may also have the ability to perform controlled movements in height, allowing it to e.g. climb up stairs, jump, or crouch.
Comparing to the the stationary expression dimension, the accounted number of DoFs may seem lees intuitive than for the other dimensions, as it does not related to the number of legs or wheels that the robot has, nor to their articular structure.
A legged or wheeled robot can move in 1D, 2D or 3D, while also being able to perform motion that rotates about a given set of axes.
For example, a 1D-capable robot could be able to move either back and forth, or side to side.
A 2D-capable robot could move back-and-forth, and additionally either rotate left and right (yaw), or strafe to the sides, or travel up and down.
A 3D-capable robot can typically move in 2D plus rotate about the vertical axis (Yaw).
This dimension therefore accounts for the total number of axes about which the robot can perform spatial movement, be them translational or rotational axes.
It can thus account for zero to 6 DoFs, given the robot's ability to travel along its local X, Y or Z axes, or to perform yaw, pitch or roll movements.
Please refer to figure \ref{fig:spacialexpression} for any further clarifications.
\item[\textbf{Display Expression}] refers to expressions portrayed through some form of electronic light display.
This can include simple monochrome LEDs, multi-color/RGB LEDs, a monochrome (LCD) screen or an RGB screen.
Ultimately it can also include some kind of light projection system.
If no layer of expressive control is defined for the display, we consider each individually controlled LED to be one DoF (even if it is multi-colored), and each individual screen/projector to also be one DoF (regardless of its pixel resolution).
However, if the LEDs are disposed in a particular expressive way, such that they all relate to the same expressive channel, that should always be controlled as a whole (e.g. each of NAO's eyes is composed of 8 LEDs), then we consider them all to be a single, aggregated DoF.
Similarly, of a robot necessarily includes a particular type of expressive display application, such as a face, then we consider the display element to have as many DoFs as that application.
Note that in a case where e.g. the application allows to individually control the opening of each eye, that would amount to 2 DoFs.
If one can control the opening and frowning of each eye, then it has 4 DoFs.
If however the application has only a set of pre-defined expression, without any further control, then we consider it to have only one DoF, which corresponds to the discrete list of expressions.
This type of specification for display expressions allows us to abstract from the technical aspect of how the displays and lights are physically implemented, and instead specify the type and amount of expressive signals can be individually portrayed through it.
\item[\textbf{Audible Expression}] refers to any audible form of intentional expressivity that a robot may have, from simple beeps, to 4 or 8-bit audio effects (sampled or generated), or a more sophisticated speech system. 
Speech may either be pre-recorded (from humans), pre-synthesized, or synthesized during interaction using a TTS. 
Outputting speech will typically require a more modern 16- or even 24-bit audio output system.
Similarly to the case of the display expression level, we consider that each individual audio player/controller accounts to one DoF.
That means that the speech system is one DoF, and any other audio-output adds as many DoFs as the number of audio signals it can control and play simultaneously.
\end{description}
\begin{figure}[htbp]
\centering
\includegraphics[width=1.0\linewidth]{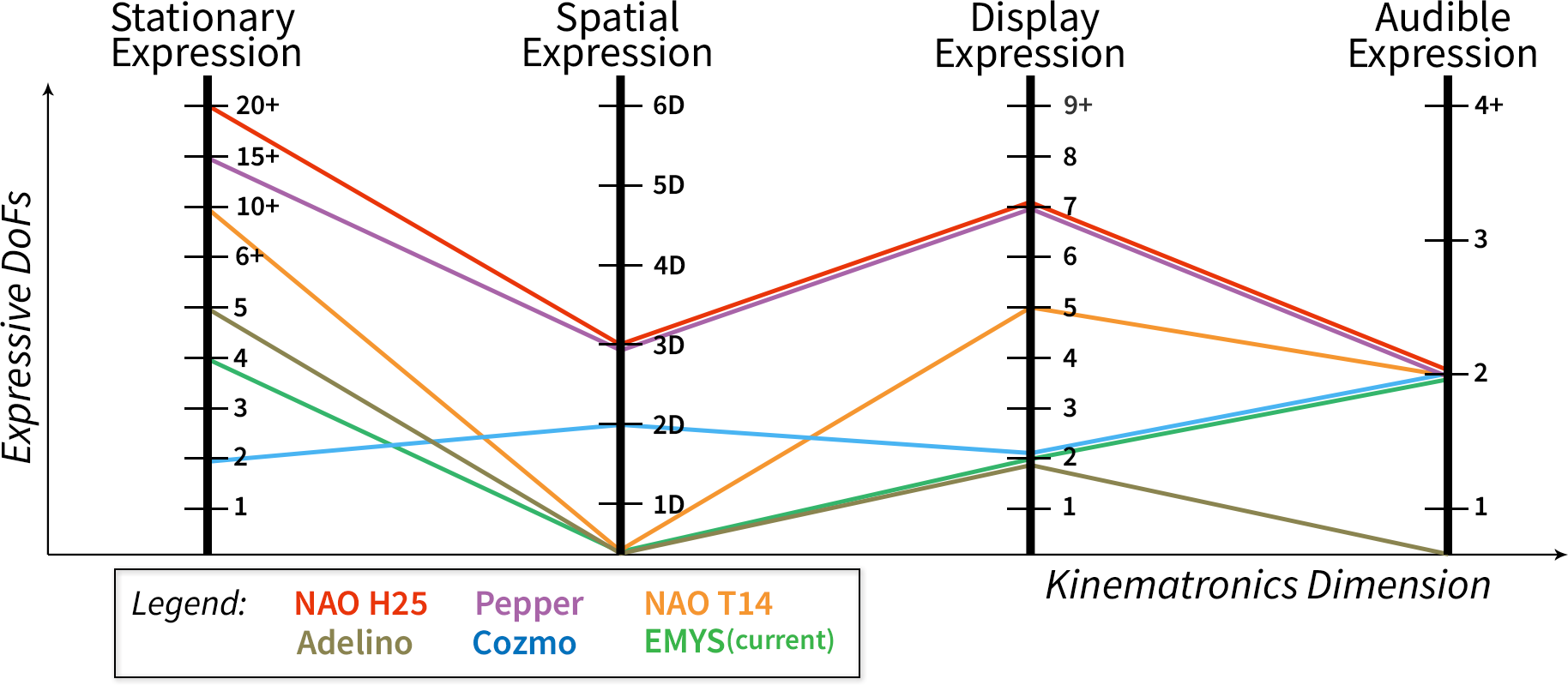}
\caption{The four kinematronics dimensions, along with an illustration of how several existing robots would be represented.}
\label{fig:kinematronics}
\end{figure}
In Figure \ref{fig:kinematronics} we can see examples of how some several robots would be placed within the kinematronics dimensions.
In particular, taking the humanoid \textit{NAO H25\footnote{\protect\url{http://doc.aldebaran.com/2-5/family/nao\_h25/index\_h25.html} \urlDate}} robot as example, we see it contains at least 25 stationary DoFs that can be used for expression.
Although its legs may be used for locomotion, there are many cases in which they are used purely for expressive postures.
Regarding spacial expression, NAO is capable of 3D motion, given that it can walk forward and backward, strafe sideways, and also perform yaw rotation.
As to display expression, and while in total, the robot has many individual LEDs, we consider the amount of display-expression DoFs to be 5: one for each eye and ear and one on the head.
Finally, for audible expression, NAO is capable of both talking and playing audio files.
While it physically contains two speakers (one on each ear), what matters expressively is that its audio-player typically allows to play only one file/sound at a time.
Be it music, expressive or warning sounds, they will all be played through the same controller.
Therefore, it contains 2 audible-expression DoFs: the TTS, and the audio-player.
The figure also compares the NAO T14\footnote{\protect\url{http://doc.aldebaran.com/2-1/family/nao\_t14/index\_t14.html} \urlDate},
Pepper\footnote{\protect\url{http://doc.aldebaran.com/2-5/home\_pepper.html} \urlDate}, 
Adelino\footnote{\protect\url{https://vimeo.com/232300140} \urlDate}, Cozmo\footnote{\protect\url{https://www.anki.com/en-us/cozmo} \urlDate}, and the latest version of the EMYS\footnote{\protect\url{https://emys.co/} \urlDate} robot.

It is extremely important to note that these dimensions do not portray how expressive a robot is or can be.
Due to design factors, a robot with e.g. few static expression DoFs such as the EMYS or the Adelino, may be considered more expressive than a high-DoF robot such as the NAO. The purpose of these dimensions are solely to enumerate and provide a specification for the various expressive channels that can be found in robots, and does not provide any hints for comparing the overall expressiveness between them.
\section{The \textit{Nutty Workflow} and \textit{Pipeline} for Robot Animation}
In order to implement social robots that are based on the concept and processes of robot animation, one must properly introduce these into the design and development workflow.
In this section we introduce general concepts on how a system architecture should be laid out and used, which is presented as the \textit{workflow} for the design and development process, and a \textit{pipeline} that can inform the design, development and execution of the animation engine.
The pipeline and workflow presented here are deeply inspired on the work developed previously with the Nutty Tracks animation engine, which was used as a sandbox to explore and develop new robot animation techniques for interactive applications \cite{Ribeiro2013,Ribeiro2016,Ribeiro2017}.
As such, we refer to these as the \textit{Nutty Workflow} and the \textit{Nutty Pipeline}.

Both the workflow and pipeline presented here aim specifically at allowing the type of animation capabilities previously mentioned throughout this chapter.
As such, these should not stand as general workflows and pipelines for the whole field of HRI and social robotics.
Instead, it presents the elements that should (or are suggested to) be present to achieve highly animate social robots, that exhibit the illusion of life, and whose design and development was carried out with animation theories and practices in mind.
Further modifications should be carried out in order to accommodate any other requirements.

\subsection{Concept Design}
First, if developing a new robot, its concept design must carefully consider all the expressive capabilities and the kinematronic dimensions needed.
We will not deeply explore this concept there, as it is also subject of study in other works.
In particular we refer to the work by Hoffman \& Ju which explores the initial stage of designing a robot with its expressive movement in mind \cite{Hoffman2014}.
This stage should include both hand-drawn concepts, along with 3D animated concepts, and even pre-visualization prototypes that allow the designers and developers to virtually simulate how the robot would behave during specific use cases of interaction with humans.
Such pre-vizualization can be developed using game-development engines such as the Unity\footnote{\protect\url{https://unity3d.com} \urlDate} or the Unreal Engine\footnote{\protect\url{https://unrealengine.com} \urlDate}.
This initial concept stage will help to inform developers both about the aesthetical design of the robot, on its kinematic structure, such as number of joints, and range of motion, and also about the use of display expression elements.
Sound design \cite{sonnenschein2013sound} can also be explored at this stage, and can be used both on rendered 3D animations of the robots, and on the interactive pre-vizualization.
If the robot will be performing speech, it is also a good idea to pre-visualize how it will look with the robot at this stage, as that may seriously impact the design of any facial features that should become animated while the robot is speaking.
\subsection{The Nutty Workflow}
\begin{figure}[htbp]
\centering
\includegraphics[width=1.0\linewidth]{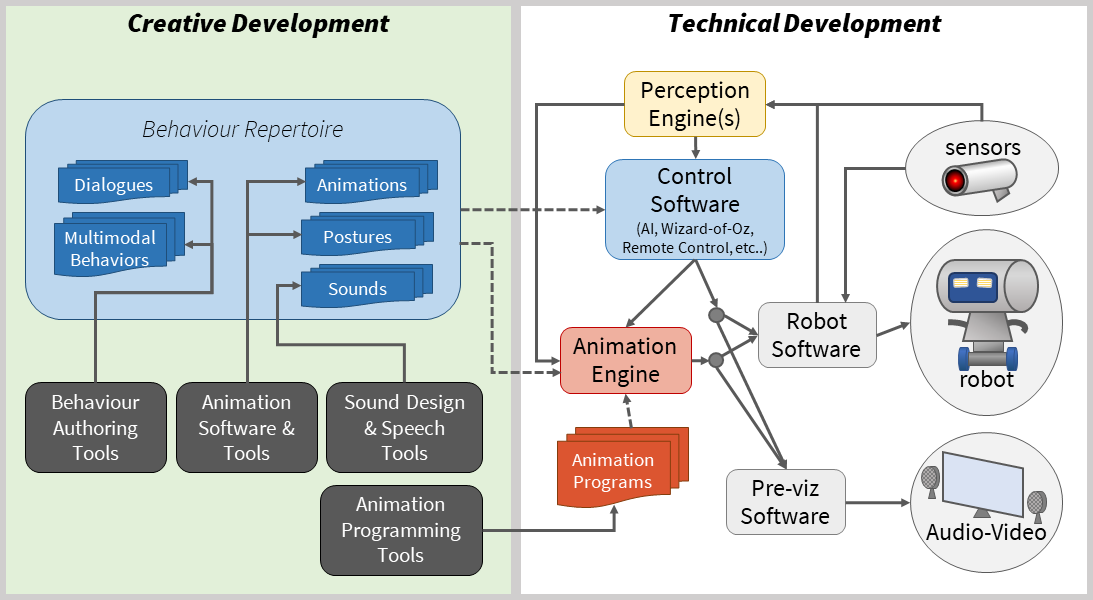}
\caption{The \textit{Nutty} robot animation workflow.}
\label{fig:workflow}
\end{figure}

Figure \ref{fig:workflow} illustrates the \textit{Nutty Workflow}.
We split the workflow in two main areas: the creative development and the technical development.
The idea is that both are intrinsically part of the model and should holistically be considered as a whole.
In the creative development area, we find most of the behaviour-authoring related to the social robot, including the development of pre-designed animations and postures, sound design, dialogues and composite multi-modal behaviours, which allow to sequence and synchronously play back a set of e.g. animations and postures, along with dialogue and sound effects.
This area is expected to include non-technical developers such as animators or psychologists.
As such it is important to carefully consider the type of tools used, to make sure they can produce properly specified assets that can be used further in the software.
We will addressed and elaborate on such tools later in section \ref{sec:tools}.

The technical development area consists of the software architecture that is typically expected for a social robot.
That includes, on the hardware part, both the \textbf{Robot} and \textbf{Sensors}, along with the lower-level \textbf{Robot Software} that controls and communicates with both.
Note that while the robot will likely contain sensors already, other external sensors may be used, such as external cameras (RGB or RGB-D) for object and user tracking and recognition, or even for localization of the robot.
As such it is useful to include a dedicated \textbf{Perception Engine} that can handle the input signals and translate them to symbolic, meaningful inputs for the Control Software and the Animation Engine.
The \textbf{Control Software} is illustrated as a single component, but may be split into various sub-components depending on the application.
This should handle the actual application-domain knowledge and control, which allows the robot to perform a given task or application.
Alternatively, it can consist of remote control tools, such as a Wizard-of-Oz, or a tele-operation panel.
The output of Control Software should be discrete and well-specified commands, given to either the robot software directly (e.g. shutdown, reset, etc..), or to the Animation Engine.

Because the focus of this workflow is robot animation, we do place the \textbf{Animation Engine} as a separate component.
This engine should be able to handle all the commands that control the various kinematronic abilities of the robot.
Depending on the aim of the application, it may have various levels of complexity.
There are explained further in Section \ref{sec:pipeline}.

Note that the animation engine has the ability of running \textbf{Animation Programs}.
These programs differ from a static animation file, in that they contain a sequence of rules that allow the generation, transitioning and blending of various expressive modalities, along with the computation of ad-hoc motion such as the ones that are produced through inverse kinematics or path planning (Section \ref{sec:pipeline}).
While a more traditional architecture would delegate such techniques to the actual robot software, we claim that including all motion control in the animation engine allows to seamlessly use \textbf{Pre-Viz Software} in place of the real robot during much of the development.
In particular, such Pre-Viz aims at allowing the creative developers to work on the robot's behaviour and expressivity, in an interactive way, in order to ensure that the final behaviour of the robot during an interaction will match the intended, authored behaviour, as close as possible.

\subsection{The Nutty Pipeline for programmable robot animation engines}
\label{sec:pipeline}

\begin{figure}[htbp]
\centering
\includegraphics[width=1\linewidth]{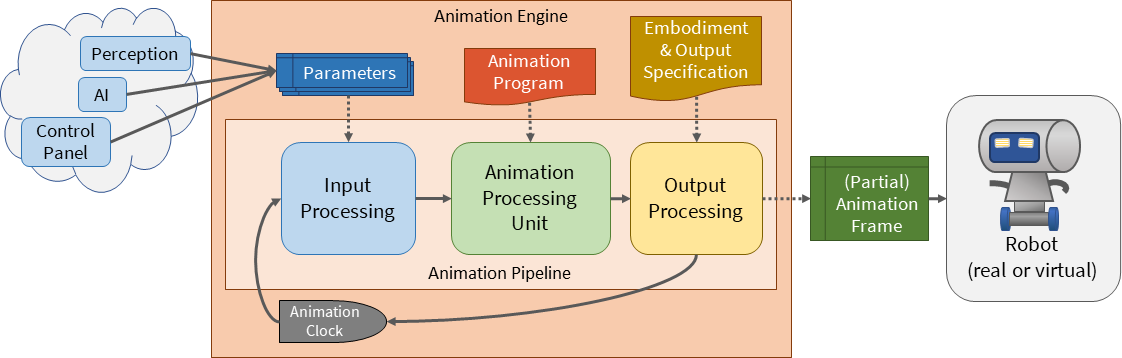}
\caption{A \textit{Nutty}-based animation engine, including the \textit{Nutty Pipeline}. At each clock cycle, the Input parameters along with the selected Animation Program are provided to the Animation Program Processor, which outputs a (Partial) Animation Frame contains the motion parameters for each programmable DoF.}
\label{fig:pipeline}
\end{figure}

A programmable robot animation engine in \textit{Nutty} terms, is a program that continuously runs a sequence of steps at a given rate, in order to produce on-line motion for a robot, based on interactive parameters specified by an AI or tele-operation module, and on user- and environment-based perception data.
The \textbf{Nutty Pipeline} lies within the animation engine, and configures the steps that run on each animation cycle.
The choice of those steps specifies how the motion is effectively produced, given a set of inputs, rules, and various types of motion generators.
The concept of the programmable animation pipeline is deeply inspired by programmable graphics pipelines such as the one provided by OpenGL \cite{RTR4}.
It means that the actual execution pipeline is not fixed, but instead, can be programmed to specify both the execution layout, the steps that will run, and how they are parameterized.
It allows an animation engine to be used with different embodiments and applications, by introducing the new concept of \textit{Animation Program}.
Figure \ref{fig:pipeline} illustrates how the Nutty Animation Pipeline fits within a \textit{Nutty}-based animation engine.

The \textbf{Input} to the pipeline consists of parameters that are provided by other components such as the \textit{AI} or the \textit{Perception Engine}.
Those input parameters can be very diverse taking several forms such: gaze-target coordinates; expressive posture to exhibit; a pre-designed animation to play; an array of emotional-state values, and even particular custom commands such as \textit{reset posture}, or \textit{activate idle-motion}.
There is no fixed specification for the animation pipeline input, which may need to be designed and developed for each particular situation. 

The \textbf{Output} of the pipeline generates an \textit{Animation Frame} that is compliant with the currently selected embodiment and output module.
Although the embodiment and output are typically enforced to work together, it is important to distinguish them.
The embodiment specification defines the available DoFs and their layout, or hierarchical structure.
The output typically connects with either the Robot Software or the Pre-Viz's API, in order to render the animation frame, either on the physical embodiment or on its virtual representation.

An \textbf{Animation Frame} (AF) is a data structure, containing both header information and a matrix of motion parameters for each programmatically animatable DoF.
The header may contain information such as the embodiment designation and the frame's delta-time.
For each DoF, the motion parameters may have various data types, depending on the kinematronic dimension of the DoF.
For non-integer numeric values, it may be as simple as a single, absolute set-point (no derivatives), or include 1\textsuperscript{st}, 2\textsuperscript{nd} or 3\textsuperscript{rd}-order derivatives (velocity, acceleration and jerk).
However it may also contain discrete or enumerate values, which are more appropriate for e.g. a display-expression component with pre-defined expressions.
We also distinguish between an Animation Frame and a \textbf{Partial Animation Frame} (PAF) in that the partial animation frame may contain only part of the whole list of DoFs (or in some cases, even none - an \textit{empty} animation frame).
This allows the pipeline to output only the signals that have been modified in each cycle, allowing to control different DoFs at different rates, and to perform blending and other operations using only a selected set of DoFs.
When we refer to an AF is, it always contains parameters for all the DoFs (i.e., a \textit{full} animation frame), while a reference to a PAF means that it may contain all or only part of the DoFs, and even be an empty frame, with no DoFs (which therefore produces no effect).

\begin{figure}[htbp]
\centering
\includegraphics[width=0.9\linewidth]{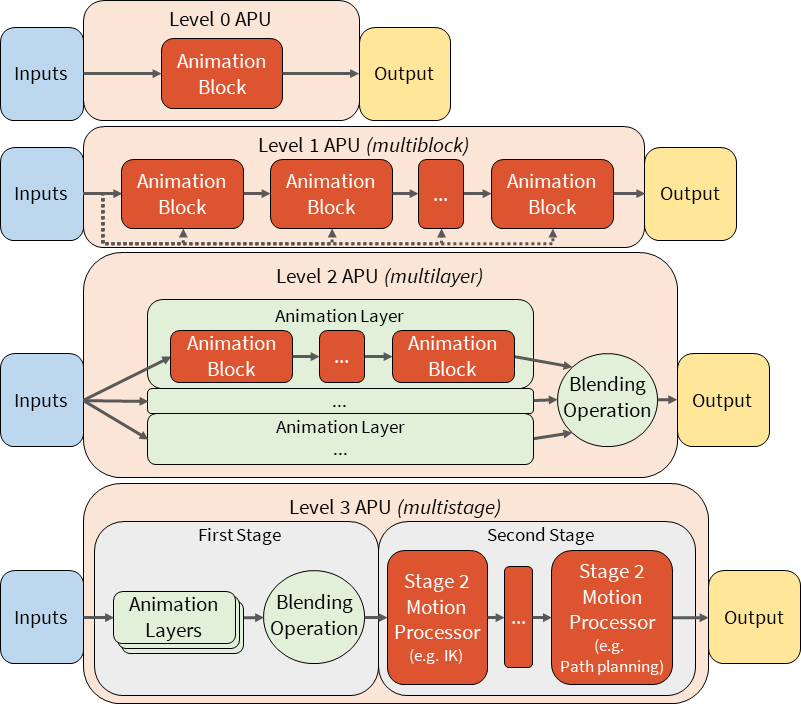}
\caption{The four types of \textit{Nutty} APUs.}
\label{fig:apu}
\end{figure}

The central component of the Animation Engine and Pipeline is the \textbf{Animation Processing Unit} (APU), which executes an \textbf{Animation Program}.
In the Nutty pipeline, an Animation Program takes a similar role as a \textit{Shader} program in the \textit{OpenGL} pipeline \cite{RTR4}. 
The APU can be developed to run at different levels of complexity, depending on the requirements of the robot-animated application, as illustrated in Figure \ref{fig:apu}.

The main building-block of the APU is called an \textbf{Animation Block} (AB).
Multiple variants of ABs are created for different purposes.
Each of these blocks takes in a set of input parameters, and generates a PAF through a specific method such as playing an animation file, or generating a motion signal through a mathematical formula. 
We further distinguish an \textit{operator} AB as one that takes in a PAF that already contains a signal and modifies it, versus a \textit{source} AB, which provides a source for the signal and generates it.
In many cases the AB will also manage an internal state, such as in the case of an animation file player, for which, given the delta-time as input, the AB calculates the new time-position within the animation, and outputs the respective frame.
That allows each AB to control how the signal is produced in the time-domain, along successive cycles of the animation engine.
Given that they output PAFs and not necessarily AFs, an AB may also be some sort of single-dimensional motion generator such as a sine-wave or 1D Gaussian noise-generator.
The pre-loaded Animation Program will specify the type of APU that is required, describe the required ABs, and specify how they are laid out into sequences, layers and stages.

The following list provides an overview of various complexity levels for APUs.
A given animation engine can be developed to support only up to a specified level of complexity, or support all of them.
Later, the AF will let the engine know what kind of layout is required to be set-up.
\begin{description}
\item[The \textbf{Level 0} APU] is the simplest form of APU, and contains a single Animation Block. 
Conceptually, a Level 0 APU is also interchangeable with an AB, as both contain a single execution step.
\item[The \textbf{Level 1} APU] supports multi-block processing, or a sequence of Level 0 APUs.
Each AB may output to another AB and therefore it allows for more complex animation, that is achieved by sequential composition of ABs.
\item[The \textbf{Level 2} APU] supports multi-block, multi-layer processing.
Each sequence of AB blocks composes a single layer and is equivalent to a Level 1 APU.
The various layers are blended using a specified Blending Operation, in order to produce a final, single PAF.
\item[The \textbf{Level 3} APU] supports multi-block, multi-layer and multi-stage processing. 
At the moment we define only two stages. 
The \textbf{First Stage} consists of a Level 2 APU.
The \textbf{Second Stage} allows to include more complex and intensive motion-generation processors such as inverse kinematics (IK) or path-planning. 
The Stage 2 processors are meant to be used as post-processing steps, and should be applied to several - or all - of the DoFs simultaneously. Nutty Tracks \cite{Ribeiro2013} is an example of a Level 3 programmable animation engine\footnote{\protect\url{https://vimeo.com/67197221} \urlDate}$^{,}$\footnote{\protect\url{https://vimeo.com/232300140} \urlDate}.
\end{description}
Note that depending on the requirements of the animation engine, one may create other forms of APUs, such as a multi-stage, multi-block APU that does not supports layers, or a multi-layer, single-block APU that does not support sequential composition.

\newcommand{\round}[1]{\lfloor #1\rceil}
\newcommand{\sgn}{\text{sgn}}
\newcommand{\Eta}{\text{H}}
\newcommand{\abs}[1]{\lvert\ #1\rvert}

\section{The Nutty Motion Filter (NMF)}
\label{sec:filter}
When we take and adapt methods or techniques from CGI animation to robots, it is common to run into a particular pitfall regarding the generated motion signal.
In CGI, objects can move around freely with no physical or kinematic constraints.
As such it is \textit{easy} to elaborate techniques that produce various kinds of motion, and to shape the motion into the expected end-results, following on simple interpolation techniques, and even using stepped motions (ones that are discontinuous).
The fact is that virtual motion is, by nature, discrete, so it is always rendered in discontinuous steps, no matter how small those are, even if any derivatives are also calculated.

In robotics however, the motors are physical and therefore enforce certain kinematic constraints which, if not met, may result in errant and jerky motion.
If one attempts to render a stepped motion on a robotic servo, the resulting movement will necessarily be continuous, moving from its initial position all the way to the final position, no matter how fast that motion might be.
Even if we \textit{try} to ignore it, inertia and other external forces will always be playing a part on the resulting motion.
Therefore motion generated for robotic use must comply to different norms than the one produced for purely virtual applications.

In particular, a motion signal generated for a robot should be at least $C^2$ continuous, i.e, containing a derivative of order 2 or more.
For servos and motors that power articulated structures in simpler robots, such a $C^2$ signal is typically enough (i.e., motion explicitly contains a limited acceleration component).
In the case of more complex robots, and in particular for motion in space, the motion signal that drives e.g. the path of a robot must be $C^3$ (i.e., containing jerk, which is the derivative of acceleration), or even more (jounce is the 4th order derivative of motion, i.e., the derivative of jerk).
Furthermore, in addition to angular limits, the mechanics and motors used will typically enforce a physical limit on each of the derivatives' value, which, if violated, may result in either physical damage, or in disorderly motion.

	\begin{figure}[htbp]
	\centering
	\includegraphics[width=0.6\linewidth]{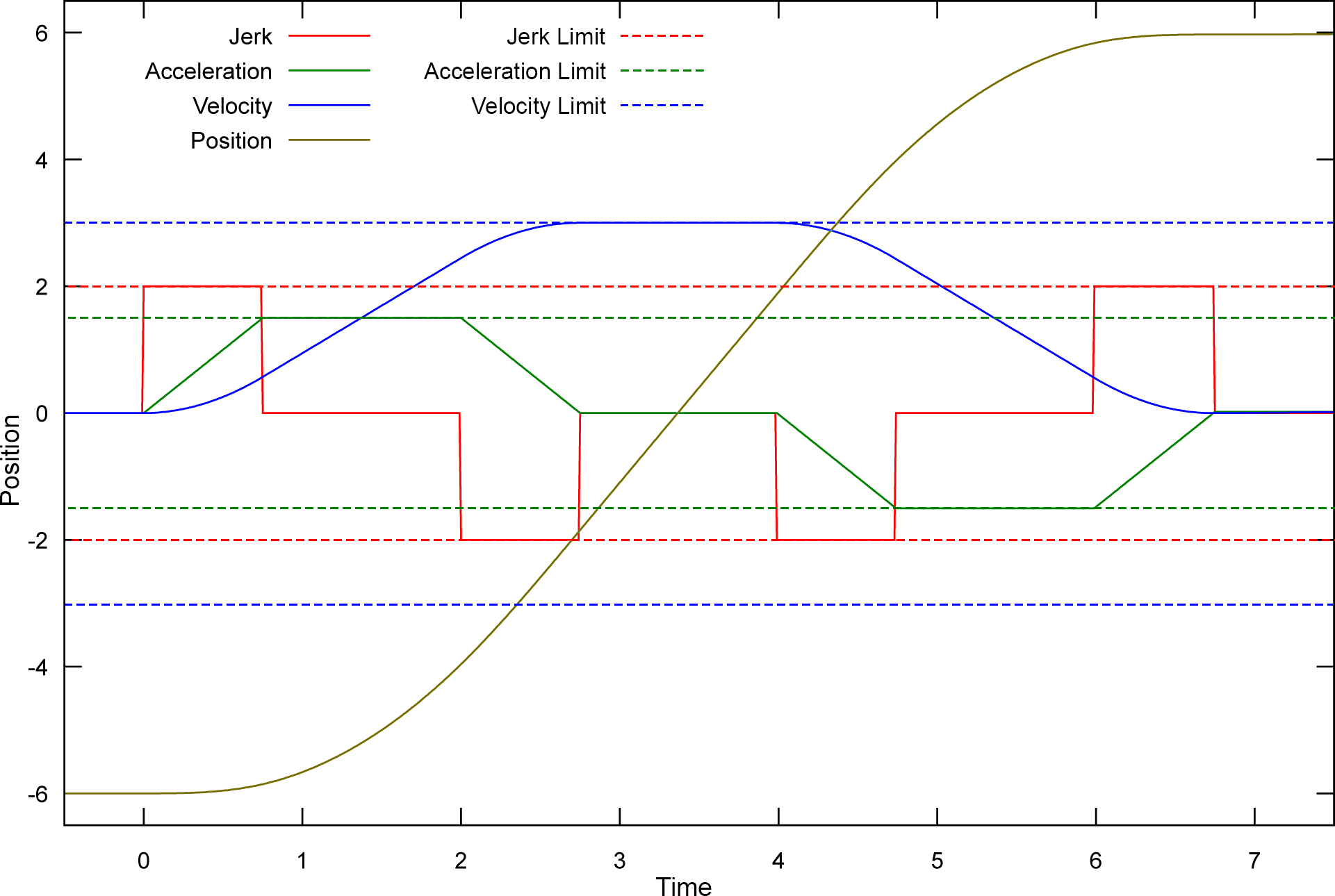}
	\caption{A diagram illustrating jerk, acceleration and velocity of a $C^3$ continuous motion signal that moves from -6 to 6.\protect\footref{foot:jerk}.}
	\label{fig:motion_derivatives}
\end{figure}	
\footnotetext{\label{foot:jerk}\protect\url{https://en.wikipedia.org/wiki/Jerk_(physics)} \urlDate}

Figure \ref{fig:motion_derivatives} illustrates a simple motion signal with $3^{rd}$ order derivatives between two positions (-6 and 6), along with the limits of each of the derivatives.
The signal input is referred to as the set-point, and in this case is a stepped signal, which is the most basic type of signal that can be used for motion control.
Recall that a stepped signal is very undesirable for robots, but is, in fact, the type of signal that is typically produced by a CGI application.
A CGI application typically runs at a high frame-rate (e.g. 60Hz), which generates motion in small steps of $\frac{1}{60}$ seconds, which therefore becomes unnoticeable on screen.
Therefore, it is generally not required to calculate all the derivatives that ensure the smoothness of the motion.
If such a stepped signal is, however, applied to a robotic servo, it is likely to cause a lot of audible noise, along with jittery motion, given that, despite the stepped input, the motor will in fact have to move through the intermediate positions between the current one and the set-point, and that achieving that motion (velocity) will lead it to accelerate and de-accelerate between each step.
Despite this issue, various authors in the field of HRI have actually used simple position-based motion controllers to control small, expressive robots motion controllers \cite{Gray2010,Hoffman2012,Ribeiro2012}.
As long as the generated motion is \textit{slow} enough, guaranteed to \textit{seem} smooth, and produced at a rate of at least 30Hz, the jittery effect may become mitigated or at least acceptable.

Throughout our work we have, at times, took that same, simplistic approach.
However in the long term, we feel the need for a proper motion filter that can be used as a bridge between any discontinuous, stepped motion such as the one typically produced in CGI, and the $C^3$ continuous motion required by robotics. 
Furthermore, because we place such a strong focus on the character animation aspect, we also considered it desirable to have a motion filter that would allow some kind of tweaking, in order to adapt the resulting motion not only to the robot's embodiment, but also to its character's traits, such as mood, personality or emotion.

The Nutty Motion Filter (NMF), presented in this section, solves that problem by:
\begin{itemize}
	\item Taking as input any $C^0$ motion, in real-time, in a sample-by-sample basis (i.e, one sample at a time), in irregular time intervals;
	\item Outputting a $C^1$, $C^2$ or $C^3$ signal corresponding to the input one saturated by its velocity, acceleration and/or jerk limits, at a steady output rate;
	\item Providing character parameters that can be tweaked to shape the motion produced, namely in terms of smooth in/out, smooth damping, or if and how the motion should produce follow-through such as overshooting or controlled damped oscillation;
\end{itemize}

In addition to the contribution contained in this section, we have also made available an online simulator of the Nutty Motion Filter\footnote{\protect\url{http://www.tiagoribeiro.pt/nutty/motionfilter.html}}, which will allow the reader to further test and visualize the motion produced by the NMF, using various different parametrizations and trajectories.

\subsection{NMF Definition}

The Nutty Motion Filter is defined as the function $X(x(t), t(i), s)$, where $x(t):\mathbb{R}^{+}_0\to[P_{min}, P_{max}]$ is the motion signal history, i.e., the previous positions that were output from the filter. 
The parameters $P_{min}$ and $P_{max}$ represent the minimum and maximum values respectively. 
In e.g. a hinge joint, these would represent the angular limits of the joint. 
$x(0)$ is the initial position of the signal and must be specified.
The function $t(i):\mathbb{N}_0\to\mathbb{R}^{+}_0$ (shortened to $t_i$) represents the time at each sample $i$, such that $0 \leq t_{i-1} < t_i$, and $t_i-t_{i-1} = \Delta t$, where $\Delta t$ is a fixed time-step, calculated from the sample rate $R$, such that $\Delta t=\frac{1}{R}$. 
The sample rate should be chosen based on the requirements and capabilities of both the robotic and computational systems, and must be at least equal to the desired output rate. 
Therefore 30-100Hz are typically acceptable sample rates.
Note that from this definition, $i$ refers to the current sample, and therefore the current time is represented by $t_i$, while the time of the last sample is $t_{i-1}$ and so on.
The set-point $s$ is the new target position, and is used to calculate the \textit{induced velocity} $\dot{x}(t_i)$ as specified in Equation \ref{eq:filter_input}.

Finally, $x(t_i)$ represents the output that will be computed of the filter at the current time (not in the history yet), while $s$ therefore represents the input.
As such, $\dot{x}(t_i)$ must be calculated from $s$ instead of $x(t_i)$.
\begin{equation}
\label{eq:filter_input}
\dot{x}(t_k) = \left\{
\begin{array}{ll}
\frac{s-x(t_{i-1})}{\Delta t} & , \textit{if}\ k = i \\
& \\
\frac{x(t_{k})-x(t_{k-1})}{\Delta t} & otherwise \\
\end{array}
\right.
\end{equation}

\begin{figure}[htbp]
	\centering
	\begin{subfigure}[b]{0.3\textwidth}
		\includegraphics[width=\textwidth]{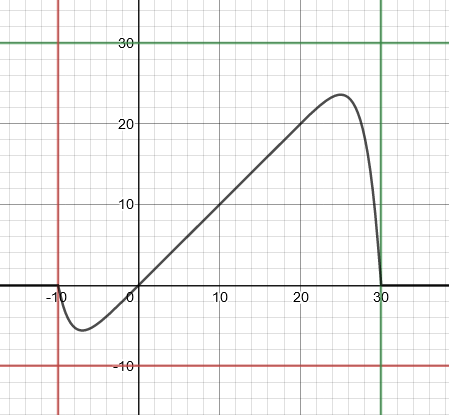}
		\caption{Output of $\Omega$ using $\beta=5$.}
		\label{fig:position_sat_b5}
	\end{subfigure}
	~ 
	\begin{subfigure}[b]{0.3\textwidth}
		\includegraphics[width=\textwidth]{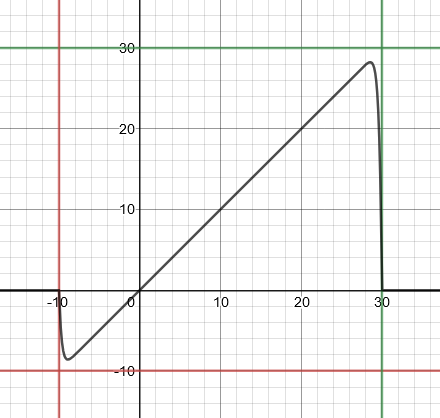}
		\caption{Output of $\Omega$ using $\beta=20$.}
		\label{fig:position_sat_b30}
	\end{subfigure}	
	~
	\begin{subfigure}[b]{0.3\textwidth}
		\includegraphics[width=\textwidth]{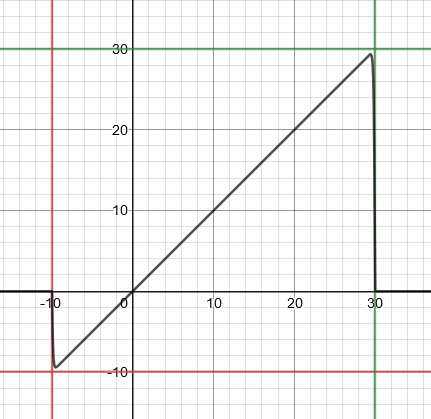}
		\caption{Output of $\Omega$ using $\beta=100$.}
		\label{fig:position_sat_b100}
	\end{subfigure}
	\caption{Comparison of the output saturation function $\Omega$ given the minimum and maximum limits of $[-10, 20]$, using three different exponents $\beta\in[5, 20, 100]$.}
	\label{fig:position_sat_compare}
\end{figure}

We start by dealing with the problem of limiting the position output of the motion using Equation \ref{eq:position_limit}.
This output saturation function $\Omega(\dot{x}, x, P_{\mathit{max}}, P_{\mathit{min}}, \beta)$ takes the induced velocity and the current output position and prevents the induced velocity from moving the signal beyond the minimum and maximum values $P_{\mathit{max}}$ and $P_{\mathit{min}}$.
As seen in the equation, the induced velocity is reduced by $\Omega$ as the current output position approaches either the minimum or the maximum limits, while through the central portion of the motion range, the velocity is untouched.
This approach differs from a hard limiter on the output (clamping), by providing some control over the motion before it reaches the position limit.
Instead of causing a hard break, we can induce a de-acceleration up to a complete stop, when the output motion is approaching its limits.
However the saturation is only applied when the induced velocity moves the signal \textit{towards} the limit, i.e., if the current position is above its center (given by $\alpha$), then the velocity is only saturated when it is positive, and if the current position is below the center, the velocity will only be saturated when it is negative.
Without this remark, the velocity would become stuck at zero upon hitting the edge of the motion range, as this saturation function would not allow it to move away from the it.
The $\beta$ parameter controls the exponent of this de-acceleration, thus allowing to control how close to the limit the output is allowed to get before being saturated.
As $\beta$ increases, the saturation becomes more similar to a hard clamping function. 
The effect of different values for the $\beta$ parameter is illustrated in Figure \ref{fig:position_sat_compare}.

\begin{equation}
\label{eq:position_limit}
\begin{split}
\Omega(\dot{x}, x, P_{\mathit{max}}, P_{\mathit{min}}, \beta) &= 
	\left\{
	\begin{array}{ll}
		\dot{x}\cdot\Bigg(1-\bigg(\frac{x-P_{\mathit{min}}-\alpha}{\alpha}\bigg)^{2\beta}\Bigg), & \mathit{if} (x>\alpha\ \&\ \dot{x}>0)\ |\ (x<\alpha\ \&\ \dot{x}<0) \\
		 & \\
		 \dot{x}, & otherwise \\
	\end{array}
	\right.\\
	\\
\alpha &= \frac{P_{\mathit{max}} - P_{\mathit{min}}}{2}
\end{split}
\end{equation}

Additionally we define the derivative saturation function $\lambda(x):\mathbb{R}\to\mathbb{R}$.
This saturation function is individually applied to each of the motion signal's derivatives in order to enforce the physical limits that are imposed by the robot's embodiment, i.e., enforce that their absolute value does not exceed a given value limit $k$.
This function may e.g. apply hard limits (described in Equation \ref{eq:filter_limit_func_regular} for exemplification purposes), or provide smooth limits as described in Equation \ref{eq:filter_limit_func}, where $k \in \mathbb{R}^{+}_{0}$ is the absolute limit value, such that $\abs{\lambda(x,k)} \leq k, \forall x \in \mathbb{R}$.
The latter one (Equation \ref{eq:filter_limit_func}) was chosen for the NMF, as it progressively saturates the input signal while it is approaching its limit, in order not to induce a hard break when the limits are reached, thus alleviating the motion oscillation that would be introduced through the use of the hard limiter.

In fact, using the tanh-limiter, the real limit is never reached, given that the input would have to be infinite for it to happen ($\lim_{x\to\infty} tanh(x) = 1$).
Being based on the hyperbolic tangent, this saturation function produces a signal that is also continuously differentiable (contrary to $\lambda'$, which is $C^0$).

\begin{equation}
\label{eq:filter_limit_func_regular}
\lambda'(x, k) = min(k, max(-k, x))
\end{equation}

\begin{equation}
\label{eq:filter_limit_func}
\lambda(x, k) = \frac{k}{2}\cdot \tanh(x/\frac{k}{2})
\end{equation}

Using the equations of motion directly to calculate the final motion upon saturating the signal would still lead, however, to some oscillation, especially in our case, where the filter digests set-points in real-time, unknowingly of when the set-point and motion will come to a rest.
The length and amplitude of such oscillation would depend on the filter order and the physical limit values.

We have however devised an additional velocity transfer function $H(v)$ (also referred to as stabilization function), presented in equation \ref{eq:transfer_function} that softly brings the motion to a rest once it starts to approach its latest given set-point.
The transfer function is applied to the \textit{saturated induced velocity}.
This stabilization function uses two hyper-parameters $\{\sigma,\rho\}$, representing \textit{smoothness} and \textit{responsiveness} respectively, that allow to tweak the filter, changing how quickly it responds and how much it is allowed to oscillate.
We call these the \textit{character parameters}, as different configurations for them will shape the motion differently.
As such we argue that they can be used to model different character traits, even when the same physical limits are enforced.
The \textit{smoothness} parameter $\sigma$ will ease out the oscillations. However, depending on other filter parameters such as the physical limits, fully easing out might become too slow and make the motion seem too muddy and flat.
That is where the \textit{responsiveness} parameter $\rho$ comes in, which allows to precipitate the easing out, so that it may still be smooth, but faster, and thus, more responsive.
While these concepts of \textit{smoothness} and \textit{responsiveness} may seem antagonistic in the context of a motion signal, they will be better explained further through illustrative examples.

\begin{equation}
\label{eq:transfer_function}
\Eta(v) = \frac{v}{2} \cdot \Bigg(\tanh\bigg(\Big(\frac{\abs{v}}{1-\rho}\Big)^{1-\sigma}-\pi\bigg)+1\Bigg), 0 \leq\sigma\leq 1, 0\leq\rho<1\\
\end{equation}

Based in the output saturation function $\Omega$ from Equation \ref{eq:position_limit}, on the derivative saturation function $\lambda$ from Equation \ref{eq:filter_limit_func}, and the stabilization function $H$ from Equation \ref{eq:transfer_function}, we present below the final equations for either a $C^3$, $C^2$ or a $C^1$ NMF filter. Recall also the definition of $\dot{x}(t_k)$ from Equation \ref{eq:filter_input}.
Higher order filters can also be inferred, based on these equations.

Equation \ref{eq:filter_c3} contains the $C^3$, or $3^{rd}$ order NMF variant, defined as $\chi_3(x, t, s)$. 

\begin{equation}
\label{eq:filter_c3}	
\begin{split}		
\chi_3(x, t_i) & = x(t_{i-1}) + \lambda(\psi_3(x, t_i), \mathit{velocity\_limit}) \\
\psi_3(x, t_i) & = \dot{x}(t_{i-1}) + \lambda(\frac{\xi(x, t_i)-\dot{x}(t_{i-1})}{\Delta t}, \mathit{acceleration\_limit}) \\
\xi(x, t_i) & = \ddot{x}(t_{i-1}) + \lambda(\frac{\frac{v\cdot\Eta(v)-\dot{x}(t_{i-1})}{\Delta t}-\ddot{x}(t_{i-1})}{\Delta t}, \mathit{jerk\_limit}) \\
v &= \Omega(\dot{x}(t_i), x(t_{i-1}), P_{\mathit{max}}, P_{\mathit{min}}, \beta) \\
\end{split}
\end{equation}

Equation \ref{eq:filter_c2} contains the $C^2$, or $2^{nd}$ order NMF variant, defined as $\chi_2(x, t, s)$. 
\begin{equation}	
\label{eq:filter_c2}
\begin{split}
\chi_2(x, t_i) & = x(t_{i-1}) + \lambda(\psi_2(x, t_i), \mathit{velocity\_limit}) \\
\psi_2(x, t_i) & = \dot{x}(t_{i-1}) + \lambda(\frac{v\cdot\Eta(v)-\dot{x}(t_{i-1})}{\Delta t}, \mathit{acceleration\_limit}) \\
v &= \Omega(\dot{x}(t_i), x(t_{i-1}), P_{\mathit{max}}, P_{\mathit{min}}, \beta) \\
\end{split}
\end{equation}

Finally, equation \ref{eq:filter_c1} contains the $C^1$, or $1^{st}$ order NMF variant, defined as $\chi_1(x, t, s)$.
\begin{equation}
\label{eq:filter_c1}
\begin{split}
\chi_1(x, t_i) &= x(t_{i-1}) + \lambda(v\cdot\Eta(v), \mathit{velocity\_limit}) \\
v &= \Omega(\dot{x}(t_i), x(t_{i-1}), P_{\mathit{max}}, P_{\mathit{min}}, \beta) \\
\end{split}
\end{equation}

\newcommand{\wTn}{\text{$W_3^{\lambda'}$}}
\newcommand{\wT}{\text{$W_3$}}

\newcommand{\xATn}{\text{$X_3^{A\lambda'}$}}
\newcommand{\xAT}{\text{$X_3^A$}}
\newcommand{\xASn}{\text{$X_2^{A\lambda'}$}}
\newcommand{\xAS}{\text{$X_2^A$}}
\newcommand{\xAFn}{\text{$X_1^{A\lambda'}$}}
\newcommand{\xAF}{\text{$X_1^A$}}

\newcommand{\xBTn}{\text{$X_3^{B\lambda'}$}}
\newcommand{\xBT}{\text{$X_3^B$}}
\newcommand{\xBSn}{\text{$X_2^{B\lambda'}$}}
\newcommand{\xBS}{\text{$X_2^B$}}
\newcommand{\xBFn}{\text{$X_1^{B\lambda'}$}}
\newcommand{\xBF}{\text{$X_1^B$}}

\newcommand{\xCTn}{\text{$X_3^{C\lambda'}$}}
\newcommand{\xCT}{\text{$X_3^C$}}
\newcommand{\xCSn}{\text{$X_2^{C\lambda'}$}}
\newcommand{\xCS}{\text{$X_2^C$}}
\newcommand{\xCFn}{\text{$X_1^{C\lambda'}$}}
\newcommand{\xCF}{\text{$X_1^C$}}

\newcommand{\xDTn}{\text{$X_3^{D\lambda'}$}}
\newcommand{\xDT}{\text{$X_3^D$}}
\newcommand{\xDSn}{\text{$X_2^{D\lambda'}$}}
\newcommand{\xDS}{\text{$X_2^D$}}
\newcommand{\xDFn}{\text{$X_1^{D\lambda'}$}}
\newcommand{\xDF}{\text{$X_1^D$}}
\newcommand{\xET}{\text{$X_3^E$}}

\newcommand{\xA}{\text{$X^{A}$}}
\newcommand{\xB}{\text{$X^{B}$}}
\newcommand{\xC}{\text{$X^{C}$}}
\newcommand{\xD}{\text{$X^{D}$}}

\newcolumntype{R}[1]{>{\raggedright\arraybackslash}m{#1}}
\newcolumntype{C}[1]{>{\centering\arraybackslash}m{#1}}
\newcolumntype{L}[1]{>{\raggedleft\arraybackslash}m{#1}}

\newcolumntype{s}{>{\columncolor[HTML]{AAACED}} c}
\arrayrulecolor[HTML]{DDDDDD}

\newcommand{\NonTanhCell}{\cellcolor[HTML]{DDDDDD}}

\subsection{Usage and Examples}
In order to demonstrate and exemplify the usage of the NMF, we will be defining a set of example filters and example input signals, for which we will then illustrate the transfer function of the example filters along with the output that results from applying them to a given example input signal. 
Throughout this section, the graphs presented show the position output that is produced by incrementally calculating the filter at each time-step $t \in [0, T_{max}]$, at a 60Hz sample rate (steps of $\frac{1}{60}s$).
The figures also display the resulting velocity, acceleration and jerk (when applied).
Recall that the filter is calculated on a per-sample basis, and has no look-ahead information on the trajectory (which allows it to be used in real-time applications).
Therefore on each moment, the filter knows only what is the current set-point, and what were the previous output positions and derivatives, thus the graphs presented are accurately representative of the output that would be produced by each filter in a real-time application.

\subsubsection{Example Filters}
We start by defining a set of example filters in Table \ref{tab:filter_examples}, organized into groups (\textit{Regular}, $A$, $B$ $C$, $D$ \& $E$) based on their hyperparameters definition, i.e., the set of character parameters and physical limits.
Within each group, there are $1^{st}$, $2^{nd}$ or $3^{rd}$ order variants, and either may use the Tanh limiter function (Equation \ref{eq:filter_limit_func}), or the Non-tanh limiter function (Equation \ref{eq:filter_limit_func_regular}).
Each example filter is designated by a name in the format $X_\alpha^{\beta}$, where $\alpha$ is the order of the filter, and $\beta$ is its hyperparameter group, followed by the symbol $\lambda'$ in case it does not use the Tanh-limiter (the $\lambda$ symbol is omitted otherwise).
Although we chose and strongly recommend to use the Tanh-limiter with the NMF, we will be demonstrating both versions in order to illustrate how it impacts the output of the filter.
The \textit{Regular} filter represents one in which our stabilizing transfer function $H$ is bypassed, and therefore it contains no character parameters.
\begin{table}[ht]
	\centering
	\begin{tabular}{cccc|C{35pt}R{40pt}|C{35pt}|C{45pt}|C{35pt}}
		\toprule
		\rowcolor{white}
		Hyperparameter &
		Filter & 
		\multirow{2}{*}{Limiter} & 
		\multirow{2}{*}{Order} &
		\multicolumn{1}{|c}{\multirow{2}{*}{Smooth}} &
		\multicolumn{1}{c|}{\multirow{2}{*}{Responsive}} &
		\multicolumn{3}{|c}{Kinematic Limits} \\
		Group & Name & & & & & Velocity & Acceleration & Jerk
		 \\
		\toprule
		\multirow{2}{*}{Regular} & \NonTanhCell \wTn & \NonTanhCell Non-tanh & \multirow{2}{*}{$3^{rd}$} & 
		\multirow{2}{*}{-} & \multirow{2}{*}{-} &
		\multicolumn{3}{c}{\multirow{2}{*}{\begin{tabular}{C{35pt}C{45pt}C{35pt}} 20 & 100 & 10 000 \end{tabular}}}  \\
		& \wT & Tanh &&&&  \\
		\midrule
		
		\multirow[b]{2}{*}{A} & \NonTanhCell \xATn & \NonTanhCell Non-tanh & \multirow{2}{*}{$3^{rd}$} & 
			\multirow{4}{*}{1.0} & \multirow{4}{*}{1.0} & 
			\multicolumn{3}{c}{\multirow{8}{*}{\begin{tabular}{C{35pt}C{45pt}C{35pt}} 20 & 100 & 10 000 \end{tabular}}}  \\
		& \xAT & Tanh & && &&& \\
		\cline{2-4}
		\multirow[t]{2}{*}{\textit{slow \& smooth}} & \xAS & Tanh & $2^{nd}$ && &&& \\
		\cline{2-4}
		& \xAF & Tanh & $1^{st}$ && &&& \\
		
		\cline{1-6}
		\multirow[b]{2}{*}{B} & \NonTanhCell \xBTn & \NonTanhCell Non-tanh & \multirow{2}{*}{$3^{rd}$} & 
			\multirow{4}{*}{0.1} & \multirow{4}{*}{0.0} &&& \\ 
		& \xBT & Tanh & && &&& \\
		\cline{2-4}
		\multirow[t]{2}{*}{\textit{slow \& vivid}} & \xBS & Tanh & $2^{nd}$ && &&& \\
		\cline{2-4}
		& \xBF & Tanh & $1^{st}$ && &&& \\
		
		\midrule
		\multirow[b]{2}{*}{C} & \NonTanhCell \xCTn & \NonTanhCell Non-tanh & \multirow{2}{*}{$3^{rd}$} & 
			\multirow{4}{*}{0.1} & \multirow{4}{*}{0.0} & 
			\multicolumn{3}{c}{\multirow{10}{*}{\begin{tabular}{C{35pt}C{45pt}C{35pt}} 90 & 700 & 50 000 \end{tabular}}}  \\
		& \xCT & Tanh & && &&& \\
		\cline{2-4}
		\multirow[t]{2}{*}{\textit{fast \& vivid}} & \xCS & Tanh & $2^{nd}$ && &&& \\
		\cline{2-4}
		& \xCF & Tanh & $1^{st}$ && &&& \\
		
		\cline{1-6}
		\multirow[b]{2}{*}{D} & \NonTanhCell \xDTn & \NonTanhCell Non-tanh & \multirow{2}{*}{$3^{rd}$} & 
			\multirow{4}{*}{0.95} & \multirow{4}{*}{1.0} & &&  \\
		& \xDT & Tanh & && &&& \\
		\cline{2-4}
		\multirow[t]{2}{*}{\textit{fast \& smooth}} & \xDS & Tanh & $2^{nd}$ && &&& \\
		\cline{2-4}
		& \xDF & Tanh & $1^{st}$ && &&& \\
		\cline{1-6}
		\multirow[b]{1}{*}{E} & \multirow{2}{*}{\xET} & \multirow{2}{*}{Tanh} & \multirow{2}{*}{$3^{rd}$} & 
			\multirow{2}{*}{0.95} & \multirow{2}{*}{0.2} & &&  \\
		\textit{fast \& very smooth} &&&&&&&&\\
		\toprule
	\end{tabular}
	\caption{Definition of filters used in the examples throughout the current section. As a mnemonic, the subscript of the filter name represents its order (1, 2 or 3), while the superscript represents its hyperparameters group (3 distinct sets $A$, $B$ $C$ $D$ \& $E$), along with an additional $\lambda'$ in case it uses the Non-tanh limiter. Additionally the Non-tanh filters were shaded to improve readability.}
	\label{tab:filter_examples}
\end{table}

\subsubsection{Example Input Signals}
\begin{figure}
	\centering
	\begin{subfigure}[b]{0.45\textwidth}
		\includegraphics[width=\textwidth]{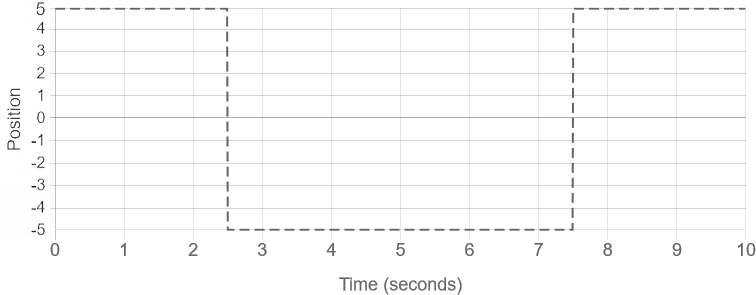}
		\caption{Example \textit{linear} input $\Phi_L$ as a highly discontinuous input.}
		\label{fig:traj_line}
	\end{subfigure}
	~ 
	\begin{subfigure}[b]{0.45\textwidth}
		\includegraphics[width=\textwidth]{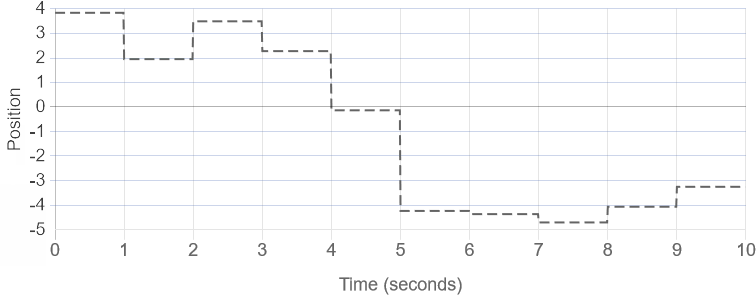}
		\caption{Example input $\Phi_R$ as a \textit{random} input. }
		\label{fig:traj_random}
	\end{subfigure}	
	
	\begin{subfigure}[b]{1.0\textwidth}
		\includegraphics[width=\textwidth]{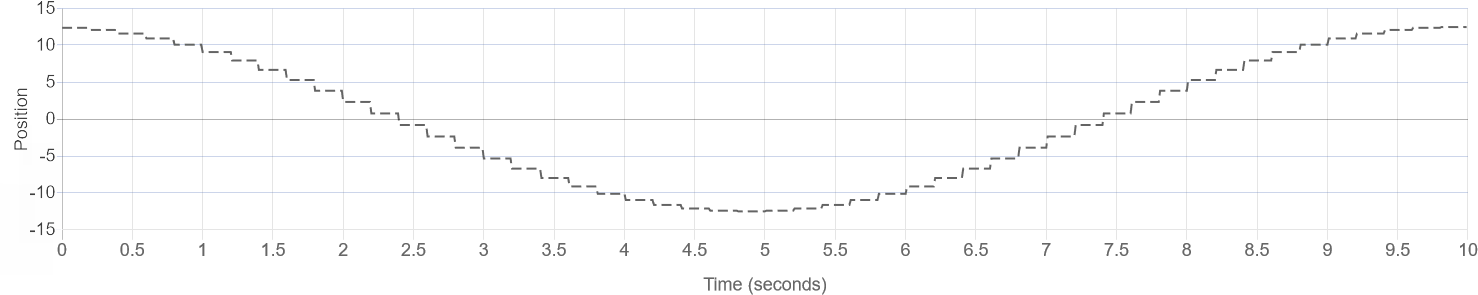}
		\caption{Example input $\Phi_C$ as a stepped \textit{circular} trajectory (in one dimension only).}
		\label{fig:raj_circle}
	\end{subfigure}
	\caption{Three example input trajectories, used to demonstrate the use of the Nutty Motion Filter.}
	\label{fig:trajectories}
\end{figure}

Figure \ref{fig:trajectories} shows three different input trajectories that were used to demonstrate the filter across different conditions. 
The first input $\Phi_L$ illustrates a very simple linear trajectory in which the set-point for the position is moved instantaneously from 5 to -5 at time $t=2.5$, and then back to 5 at $t=7.5$.

 It is intended to show how a filter responds to a large change in the input signal.
The second input $\Phi_R$ illustrates a case in which the trajectory set-point is randomly adjusted at each second. It is intended to show how a filter responds both to small and large changes in the input signal.
The third input $\Phi_C$ illustrates a case in which the final trajectory was a circle. In this case we see only one of the two dimensions of the circular trajectory. This trajectory however, was discretized into 50 points, thus producing a small step at every 0.2 seconds for the 10-second long trajectory. It is intended to emulate what in CGI would be seen as a smooth signal, but is, however, a stepped input.

\subsubsection{Example Filters' Transfer Function Response}
\begin{figure}[htbp]
	\centering
	\begin{subfigure}[b]{0.45\textwidth}
		\includegraphics[width=\textwidth]{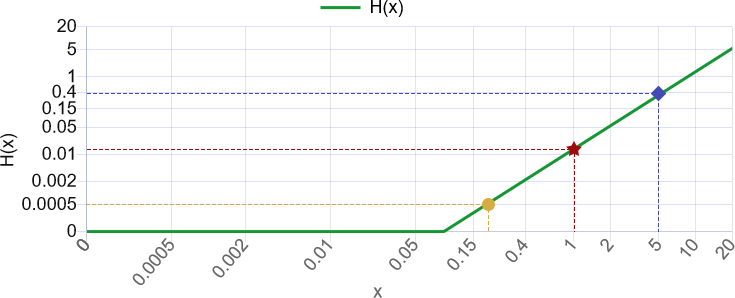}
		\caption{Transfer function of group $A (\sigma=1.0, \rho=1.0)$.}
		\label{fig:transfer_A}
	\end{subfigure}
	~ 
	\begin{subfigure}[b]{0.45\textwidth}
		\includegraphics[width=\textwidth]{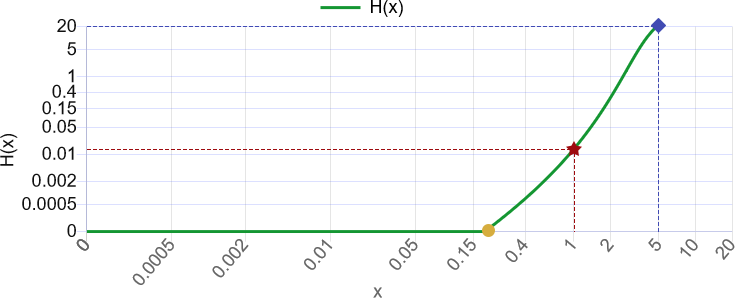}
		\caption{Transfer function of group $B (\sigma=0.1, \rho=0.0)$.}
		\label{fig:transfer_B}
	\end{subfigure}
	
	\begin{subfigure}[b]{0.45\textwidth}
		\includegraphics[width=\textwidth]{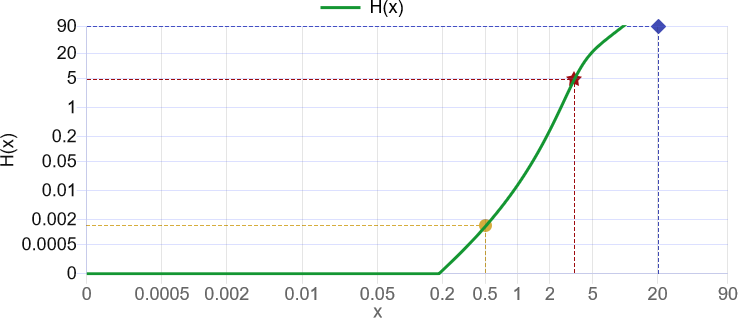}
		\caption{Transfer function of group $C (\sigma=0.1, \rho=0.0)$.}
		\label{fig:transfer_C}
	\end{subfigure}
	~
	\begin{subfigure}[b]{0.45\textwidth}
		\includegraphics[width=\textwidth]{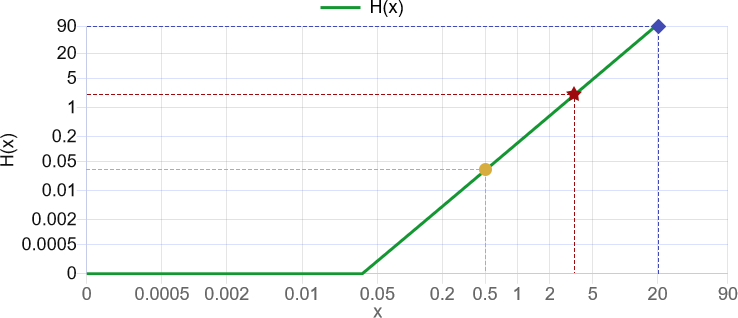}
		\caption{Transfer function of group $D (\sigma=0.95, \rho=1.0)$.}
		\label{fig:transfer_D}
	\end{subfigure}
	\caption{Plots of the different transfer functions specified by each hyperparameter group $A$, $B$ $C$ \& $D$. The domain of these graphs is $x\in[0, \mathit{VelocityLimit}]$, thus corresponding to $[0, 20]$ for $A$ and $B$, and to $[0, 90]$ for $C$ and $D$. Also note that filters in groups $B$ and $C$ share the same character parameters, which results in the same transfer function, confined only to a different domain. All graphs include three distinct points in the $x$ axis as an aid to interpret how they differ.}
	\label{fig:transfer_functions}
\end{figure}

Figure \ref{fig:transfer_functions} contains four plots that illustrate the transfer function for different hyperparameter groups.
The top graphs refer to groups $A$ and $B$, which both share one set of \textit{slower} physical limits, while the lower graphs refer to groups $C$ and $D$, which share the other set of \textit{faster} physical limits.
The transfer function of $B$ and $C$ are actually equal (same character parameters \{$\sigma$, $\rho$\}), however they consider different physical limits.
This is reflected in the plots, as each shows the output of $H(x)$ being $x\in[0, \mathit{VelocityLimit}]$. As such, $x\in [0, 20]$ for groups $A$ and $B$, while $x\in [0, 90]$ for $C$ and $D$.

\subsubsection{Output Examples using the Nutty Motion Filter}
Through this section we will be comparing various graphs in order to illustrate how the filter's response changes both given a different set of character parameters, physical limits and input trajectory.
We will also take the opportunity to demonstrate how the tanh-based limiter differs from a non-tanh-based limiter, and even to demonstrate how a given filter would behave without the use of our stabilizer function.

\begin{figure}[!htbp]
	\centering	
	\begin{subfigure}[b]{0.54\textwidth}
		\includegraphics[width=\textwidth]{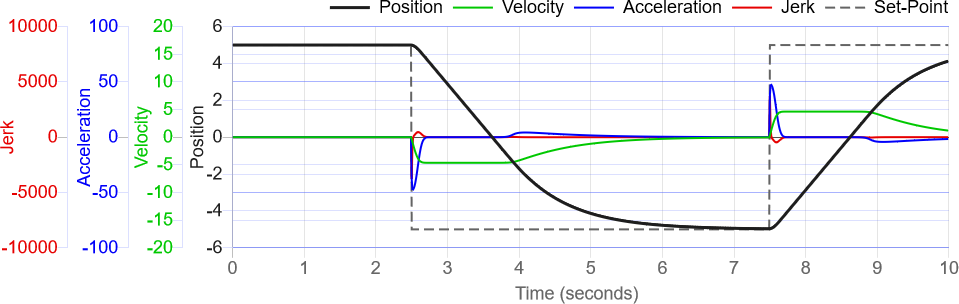}
		\caption{Output of Filter \xAT (with tanh limiter).}
		\label{fig:filterAT}
	\end{subfigure}
	~ 
	\begin{subfigure}[b]{0.42\textwidth}
		\includegraphics[width=\textwidth]{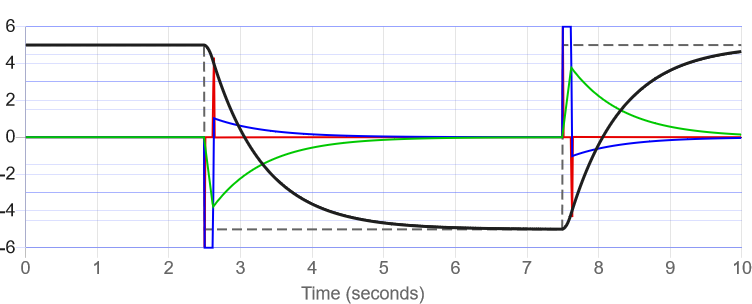}
		\caption{Filter \xATn (with non-tanh limiter).}
		\label{fig:filterATn}
	\end{subfigure}
	
	\begin{subfigure}[b]{0.54\textwidth}
		\includegraphics[width=\textwidth]{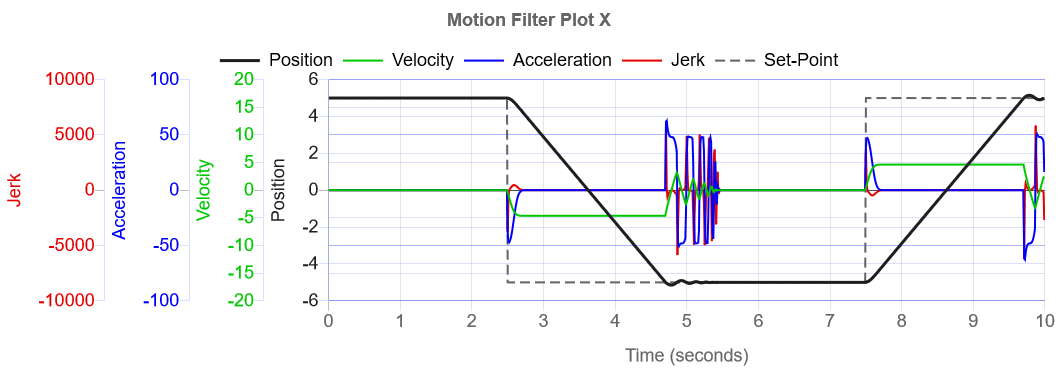}
		\caption{Output of Filter \wT ($H(x)=x$, with tanh limiter).}
		\label{fig:filterWT}
	\end{subfigure}
	~ 
	\begin{subfigure}[b]{0.42\textwidth}
		\includegraphics[width=\textwidth]{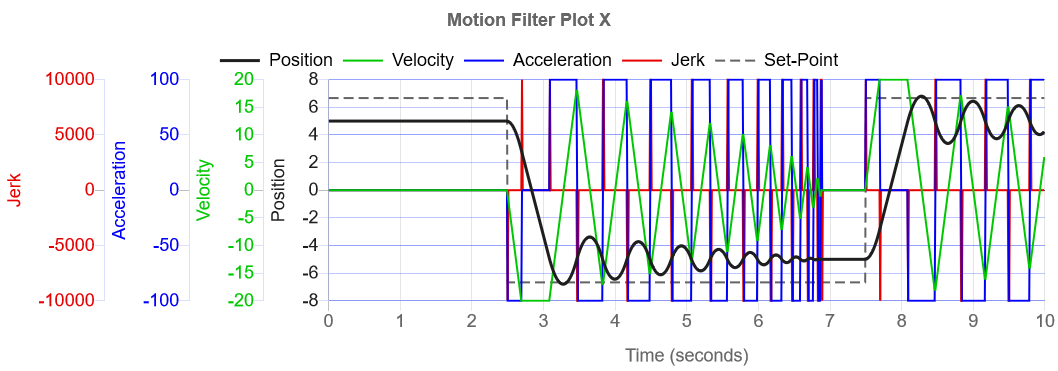}
		\caption{Filter \wTn ($H(x)=x$, with non-tanh limiter).}
		\label{fig:filterWTn}
	\end{subfigure}
	\caption{\textbf{Top row}: The output of filter \xAT (a) compared to filter \xATn (b). \textbf{Bottom row}: The output produced by the NMF equations if the transfer function was bypassed, i.e., making $H(x)=x$, while using the tanh-based limiter on the left (c), and a non-tanh based limiter on the right (d). All four plots are produced from the \textit{simple} input signal $\Phi_L$.}
	\label{fig:filterATn_compare}
\end{figure}

Using the simpler $\Phi_L$ input, Figure \ref{fig:filterATn_compare} shows on the top left (a), the output of \xAT, with the tanh-limiter, in comparison with \xATn, on the top right (b), which uses the non-tanh limiter.
The hyperparameters for this filter group ($A$) make it what we would call a \textit{slow} character, given that the motion takes some time to respond, and then again to become fully stationary when the set-point has rested.
Observing the derivatives' curves, the difference between the two variants becomes clear.
In the first case, they never hit their maximum value, and are all smooth, as they are based on the hyperbolic tangent.
The output of the tanh-based variant becomes, however, slower, because the velocity was in general, confined to a lower value than in the non-tanh version.
This illustrates the implications of the tanh-limiter on the output motion - the system will, in general, produce an output that is slower than physically allowed, by creating what we call a \textit{headroom}\footnote{Borrowed from the concept of headroom used in digital audio. \protect\url{https://en.wikipedia.org/wiki/Headroom_(audio_signal_processing)} \urlDate}, that allows to smoothly accommodate cases in which a very large change is induced by the input signal.
Due to this feature of the tanh-limiter, we differentiate the maximum value of a derivative, from its maximum sustained value.

Taking as example the first derivative, that maximum sustained velocity will be the absolute value at which the velocity tends to hold as constant (about 5 in Figure \ref{fig:filterAT}), in contrast with the real maximum absolute velocity (20 in the same Figure), which is the hyperparameter used to parameterize the filter, and includes the headroom.

For a reference, on the bottom row we also present the output of the signal that would be produced if we bypassed our transfer function, i.e., making $H(x)=x$.
In this case we see on the bottom left (c) that the signal actually responds quickly with some slight oscillation when using the tanh-based limiter, and results in severe oscillation when using the non-tanh limiter (bottom right (d)).
Without the transfer function, the output only starts to stabilize after it has reached the set-point. 
Therefore, we can observe, on the left-side, that the \wT\ filter accelerates until it reaches a maximum velocity and continues that trajectory until it reaches the set-point. 
Only then does it attempt to stabilize the output.
Because it was going too fast and even overshot it, some oscillation was produced, which however, was mitigated by the use of the tanh-limiter.
However, on the \xAT filter we see that the output starts to de-accelerate much earlier in order to allow the output to stabilize smoothly without oscillating.
These graphs therefore show that although bypassing our transfer function is a possibility, we would have no control over how fast or smoothly the filter responded, except by tweaking the physical limits, which would be an undesirable requirement. 

\begin{figure}[!htbp]
	\centering
	\begin{subfigure}[b]{0.54\textwidth}
		\includegraphics[width=\textwidth]{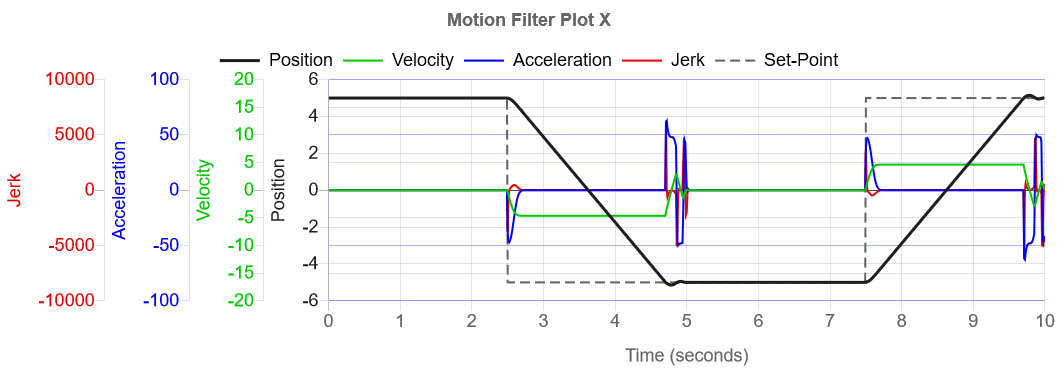}
		\caption{Output of Filter \xBT (with tanh limiter).}
		\label{fig:filterBT}
	\end{subfigure}
	~ 
	\begin{subfigure}[b]{0.42\textwidth}
		\includegraphics[width=\textwidth]{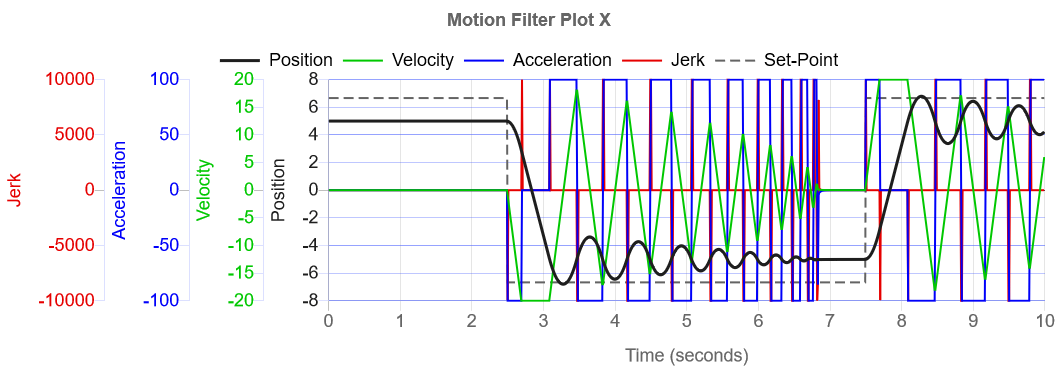}
		\caption{Output of Filter \xBTn (with non-tanh limiter).}
		\label{fig:filterBTn}
	\end{subfigure}

	\begin{subfigure}[b]{0.54\textwidth}
		\includegraphics[width=\textwidth]{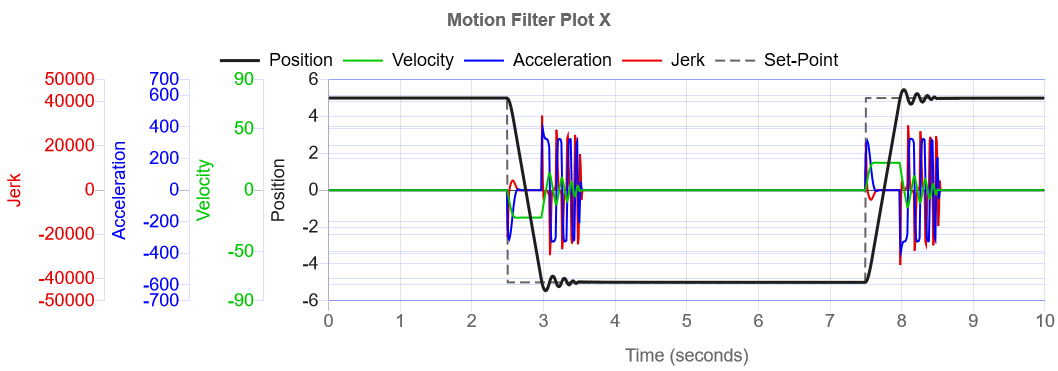}
		\caption{Output of Filter \xCT (with tanh limiter).}
		\label{fig:filterCT}
	\end{subfigure}
	~ 
	\begin{subfigure}[b]{0.42\textwidth}
		\includegraphics[width=\textwidth]{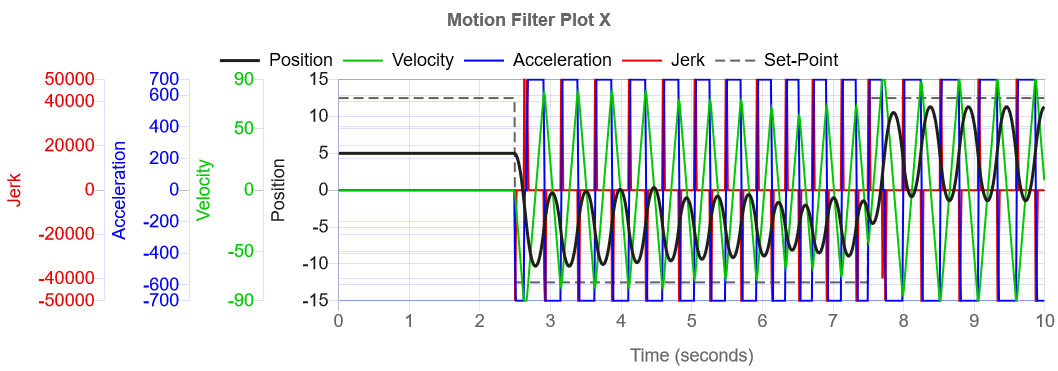}
		\caption{Output of Filter \xCTn (with non-tanh limiter).}
		\label{fig:filterCTn}
	\end{subfigure}

	\begin{subfigure}[b]{0.54\textwidth}
		\includegraphics[width=\textwidth]{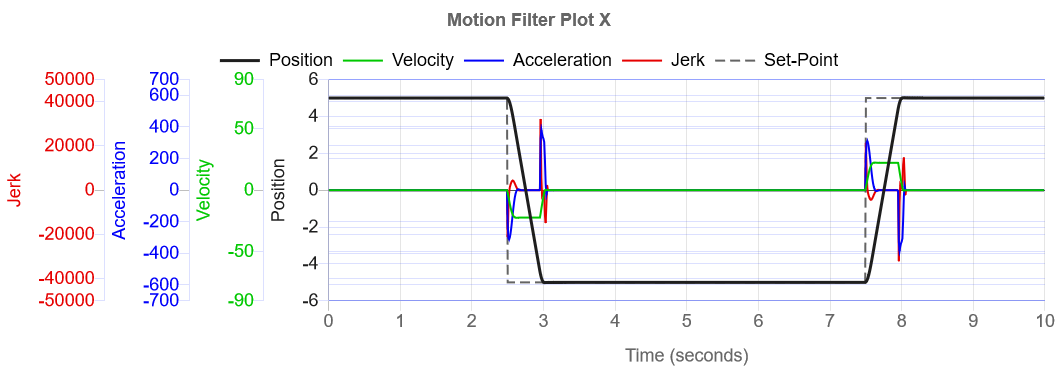}
		\caption{Output of Filter \xDT (with tanh limiter).}
		\label{fig:filterDT}
	\end{subfigure}
	~ 
	\begin{subfigure}[b]{0.42\textwidth}
		\includegraphics[width=\textwidth]{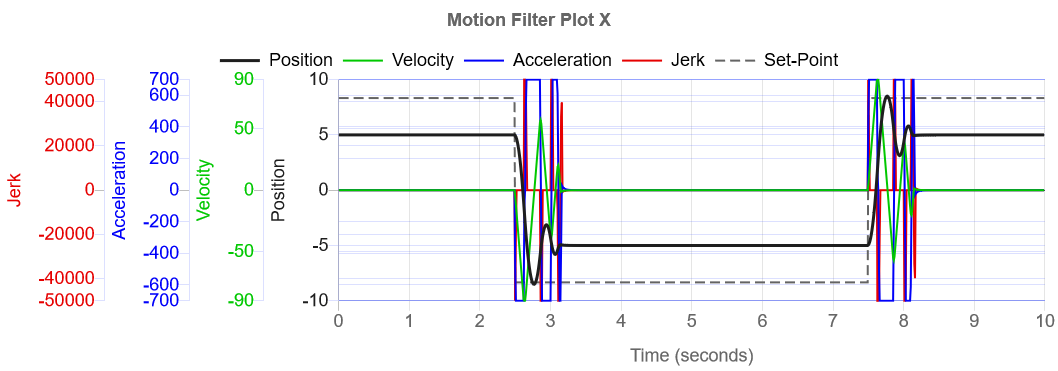}
		\caption{Output of Filter \xDTn (with non-tanh limiter).}
		\label{fig:filterDTn}
	\end{subfigure}
	\caption{Output of the $3^{rd}$ order filter of groups $B$, $C$ and $D$, using the \textit{simple} input signal $\Phi_L$.}
	\label{fig:filterDT_compare}
\end{figure}

Figure \ref{fig:filterDT_compare} show a comparison of the same input signal using the $3^{rd}$ order variants of the $B$, $C$ and $D$ filter groups, again with both their tanh and non-tanh based variants.
In this set of examples we have varied the character parameters and physical limits.
When we parameterize the filter to provide a more vivid response as in filters \xBT (Fig. \ref{fig:filterBT}) and \xCT (Fig. \ref{fig:filterCT}), we do encounter a slight oscillation effect.
This oscillation is, however, introduced due to our choice of the parameters, and is therefore a controlled oscillation, i.e., one that would allow the character to exhibit some overshooting and follow-through animation, in order to convey a sense of weight and inertia.
If that oscillation is fully undesirable, we may parameterize the filter further to produce a fast and steady response, as seen in filter \xDT (Fig. \ref{fig:filterDT}).

On the right side of the figure, the non-tanh limiter shows a faster response in comparison with the tanh-based limiter version, but then after the output overshoots, it struggles to stabilize the signal quickly, thus leading to the oscillation.
It becomes clearer why the tanh-limiter became our choice for the NMF as it allows us to tweak the \textit{shape} of the output signal (as seen on the left), without introducing that \textit{undesirable} oscillation.

\begin{figure}[htbp]
	\centering
	\begin{subfigure}[b]{0.39\textwidth}
		\includegraphics[width=\textwidth]{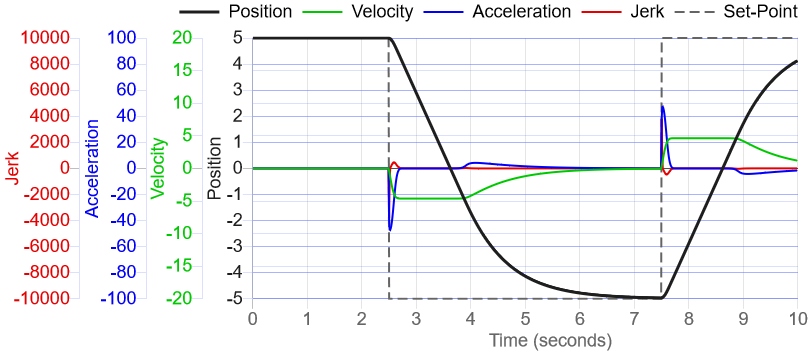}
		\caption{Output of Filter \xAT.}
		\label{fig:filterAT2}
	\end{subfigure}
	~ 
	\begin{subfigure}[b]{0.28\textwidth}
		\includegraphics[width=\textwidth]{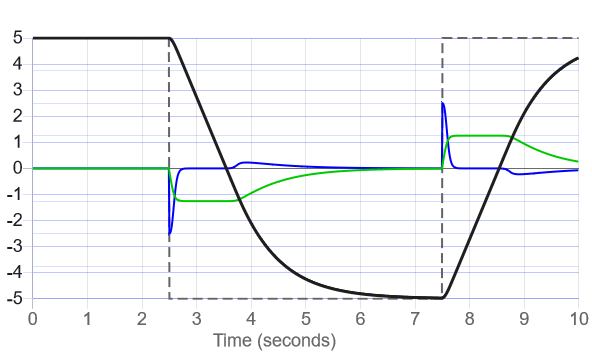}
		\caption{Output of Filter \xAS.}
		\label{fig:filterAS}
	\end{subfigure}
	~
	\begin{subfigure}[b]{0.28\textwidth}
		\includegraphics[width=\textwidth]{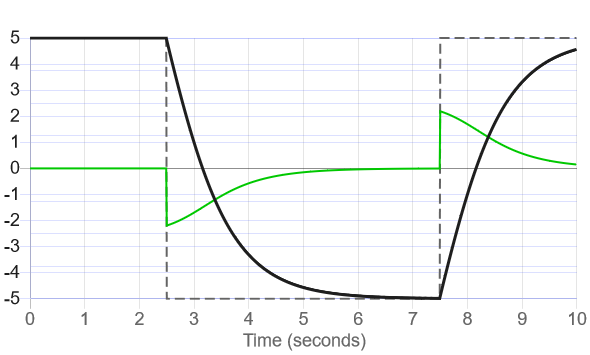}
		\caption{Output of Filter \xAF.}
		\label{fig:filterAF}
	\end{subfigure}

	\begin{subfigure}[b]{0.39\textwidth}
		\includegraphics[width=\textwidth]{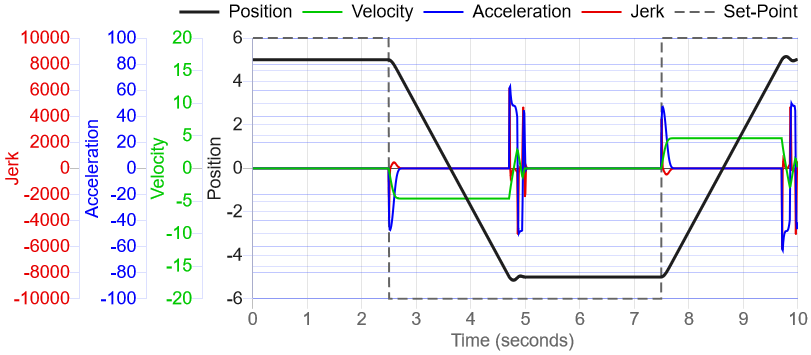}
		\caption{Output of Filter \xBT.}
		\label{fig:filterBT2}
	\end{subfigure}
	~ 
	\begin{subfigure}[b]{0.28\textwidth}
		\includegraphics[width=\textwidth]{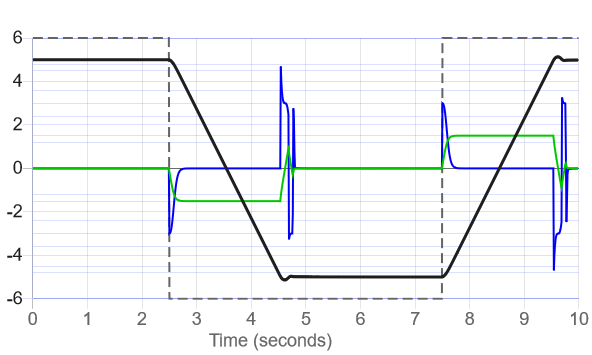}
		\caption{Output of Filter \xBS.}
		\label{fig:filterBS}
	\end{subfigure}
	~
	\begin{subfigure}[b]{0.28\textwidth}
		\includegraphics[width=\textwidth]{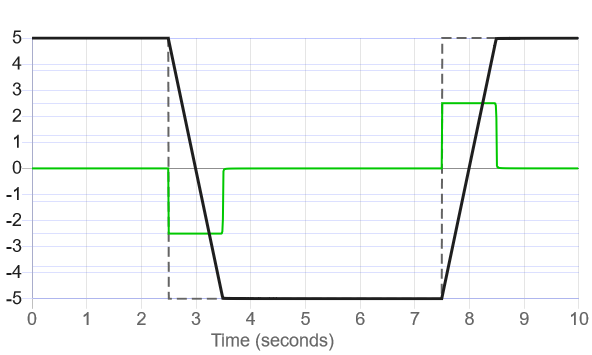}
		\caption{Output of Filter \xBF.}
		\label{fig:filterBF}
	\end{subfigure}

	\begin{subfigure}[b]{0.39\textwidth}
		\includegraphics[width=\textwidth]{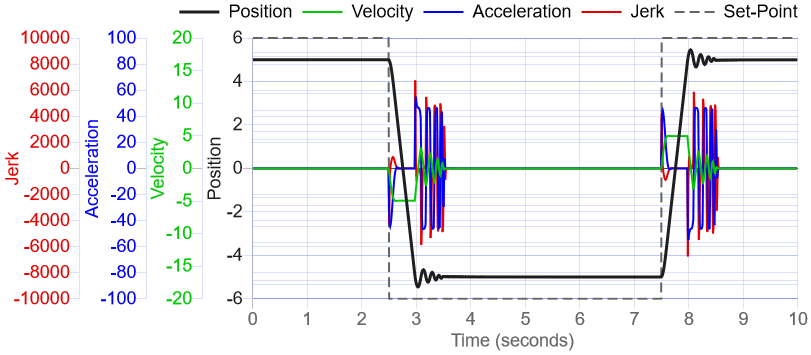}
		\caption{Output of Filter \xCT.}
		\label{fig:filterCT2}
	\end{subfigure}
	~ 
	\begin{subfigure}[b]{0.28\textwidth}
		\includegraphics[width=\textwidth]{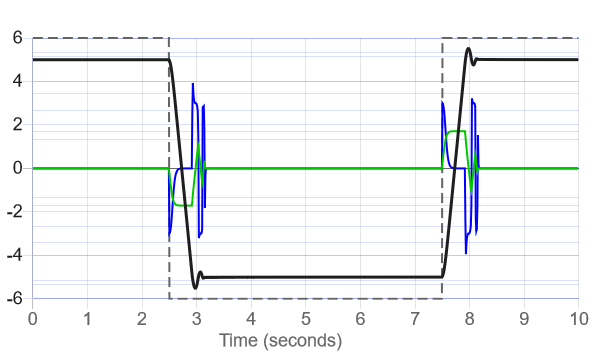}
		\caption{Output of Filter \xCS.}
		\label{fig:filterCS}
	\end{subfigure}
	~
	\begin{subfigure}[b]{0.28\textwidth}
		\includegraphics[width=\textwidth]{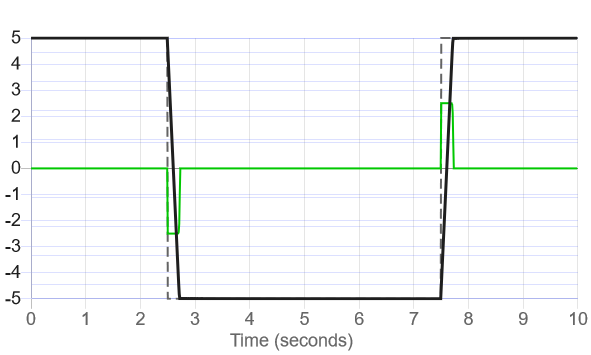}
		\caption{Output of Filter \xCF.}
		\label{fig:filterCF}
	\end{subfigure}

	\begin{subfigure}[b]{0.39\textwidth}
		\includegraphics[width=\textwidth]{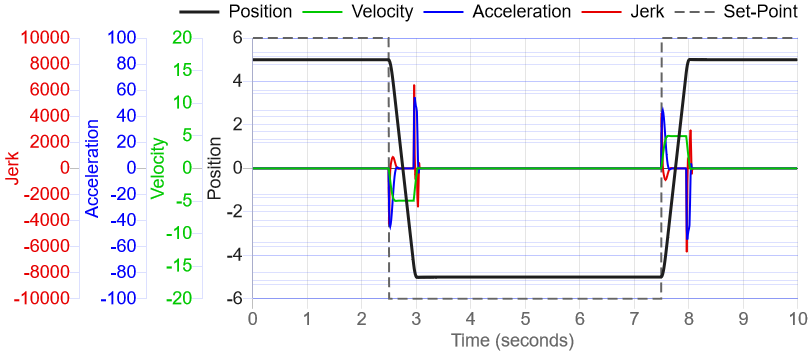}
		\caption{Output of Filter \xDT.}
		\label{fig:filterDT2}
	\end{subfigure}
	~ 
	\begin{subfigure}[b]{0.28\textwidth}
		\includegraphics[width=\textwidth]{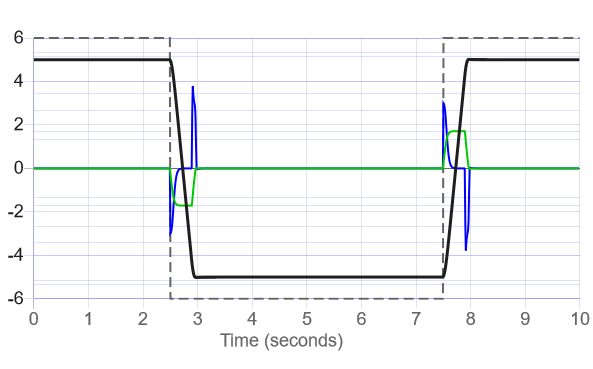}
		\caption{Output of Filter \xDS.}
		\label{fig:filterDS}
	\end{subfigure}
	~
	\begin{subfigure}[b]{0.28\textwidth}
		\includegraphics[width=\textwidth]{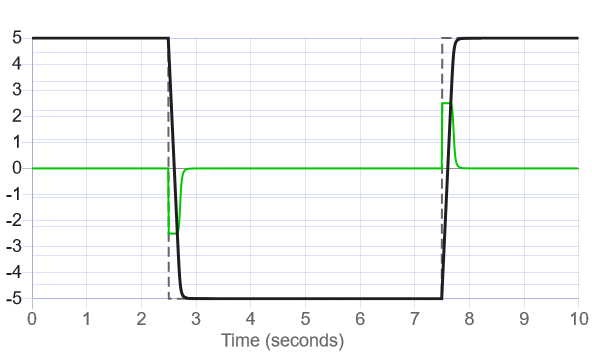}
		\caption{Output of Filter \xDF.}
		\label{fig:filterDF}
	\end{subfigure}

	\caption{Comparison of the $3^{rd}$, $2^{nd}$ and $1^{st}$ order filters for hyperparameter groups $A$ (first row), $B$ (second row) $C$ (third row) and $D$ (last row), using the \textit{simple} input signal $\Phi_L$.}
	\label{fig:filterOrders_compare}
\end{figure}

In Figure \ref{fig:filterOrders_compare} we can see a comparison between three different filter orders for each of the hyperparameter groups $A$, $B$, $C$ and $D$ using the same simple $Phi_L$ input.
This figure allows to verify that the filter's response shape remains consistent across different orders, given the same character parameters and physical limits.
What also observe that the maximum sustained velocity increases as the filter order decreases, thus suggesting that we should use the least order filter that the embodiment allows, in order to take the best advantage of the embodiment's kinematic capabilities.

In order to better conclude about how various filter parameters respond to different trajectories, we include some additional sets of plots.

\begin{figure}[ht]
	\centering
	\begin{subfigure}[b]{1\textwidth}
		\includegraphics[width=\textwidth]{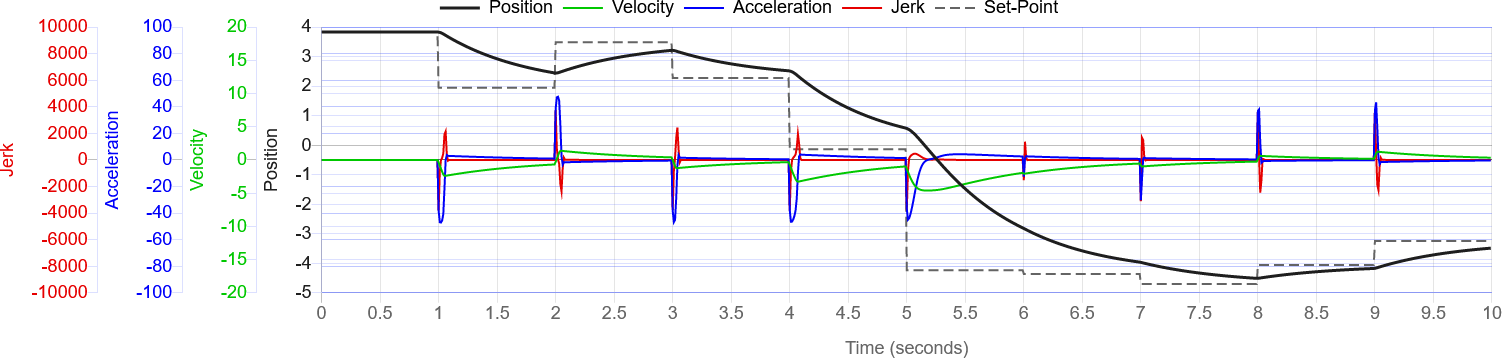}
		\caption{Output of Filter \xAT using the \textit{random} input signal $\Phi_R$.}
		\label{fig:filterA_random}
	\end{subfigure}

	\begin{subfigure}[b]{1\textwidth}
		\includegraphics[width=\textwidth]{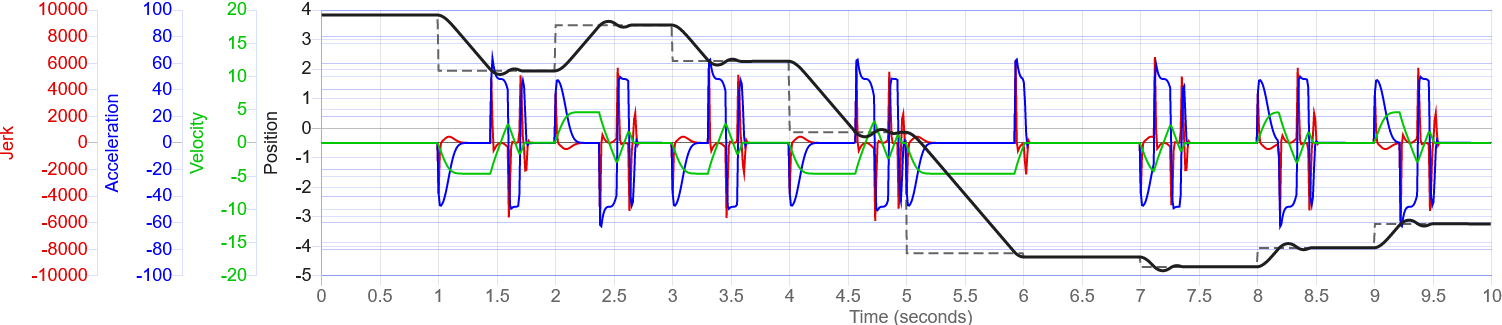}
		\caption{Output of Filter \xBT using the \textit{random} input signal $\Phi_R$.}
		\label{fig:filterB_random}
	\end{subfigure}

	\begin{subfigure}[b]{1\textwidth}
		\includegraphics[width=\textwidth]{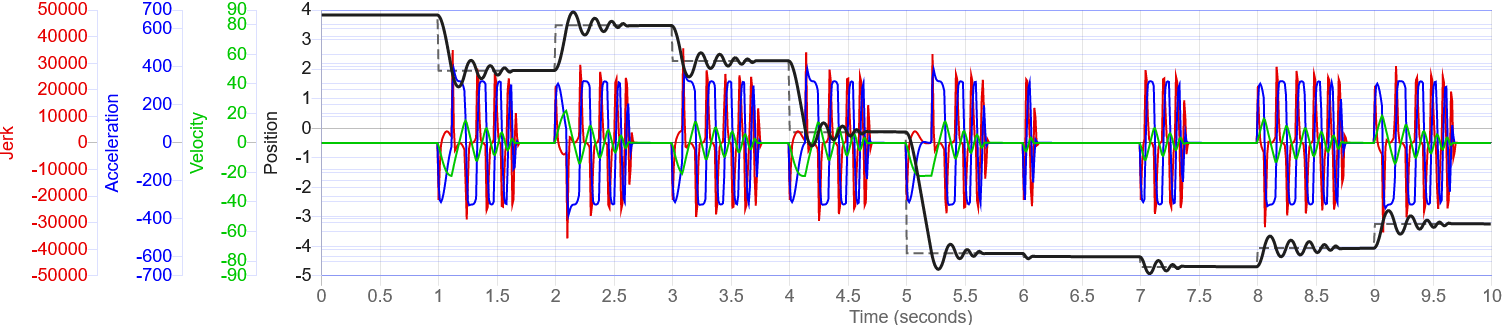}
		\caption{Output of Filter \xCT using the \textit{random} input signal $\Phi_R$.}
		\label{fig:filterC_random}
	\end{subfigure}

	\begin{subfigure}[b]{1\textwidth}
		\includegraphics[width=\textwidth]{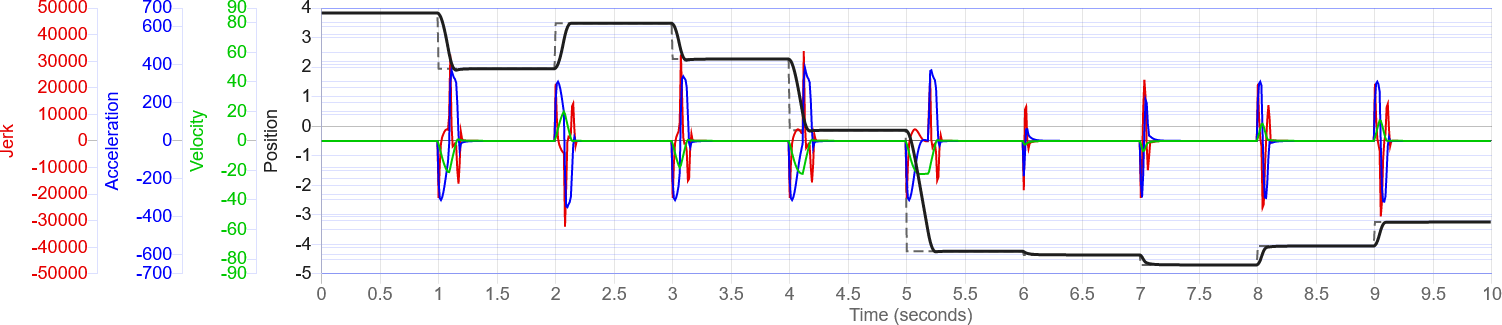}
		\caption{Output of Filter \xDT using the \textit{random} input signal $\Phi_R$.}
		\label{fig:filterD_random}
	\end{subfigure}
	\caption{Comparison of hyperparameter groups $A$, $B$, $C$ and $D$, using the \textit{random} input signal $\Phi_R$.}
\label{fig:filterRandom_compare}
\end{figure}

Figure \ref{fig:filterRandom_compare} shows the plot for each hyperparameter group $A$, $B$, $C$ and $D$, using the \textit{random} input trajectory $\Phi_R$.
In this trajectory we see how each filter responds both to large and small set-point changes. 
Filter \xAT is too slow to actually reach the set-points through half of the trajectory. 
Filter \xBT performs better there, but adds some overshooting which can be seen as a small bump. 
Filter \xCT is too loose, and although it reaches the set-points quickly, it introduces not only overshooting but also some oscillation. 
Filter \xDT illustrates what we consider as a fast and steady response, with nearly no overshooting.

\begin{figure}[!htbp]
	\centering
	\begin{subfigure}[b]{1\textwidth}
		\includegraphics[width=\textwidth]{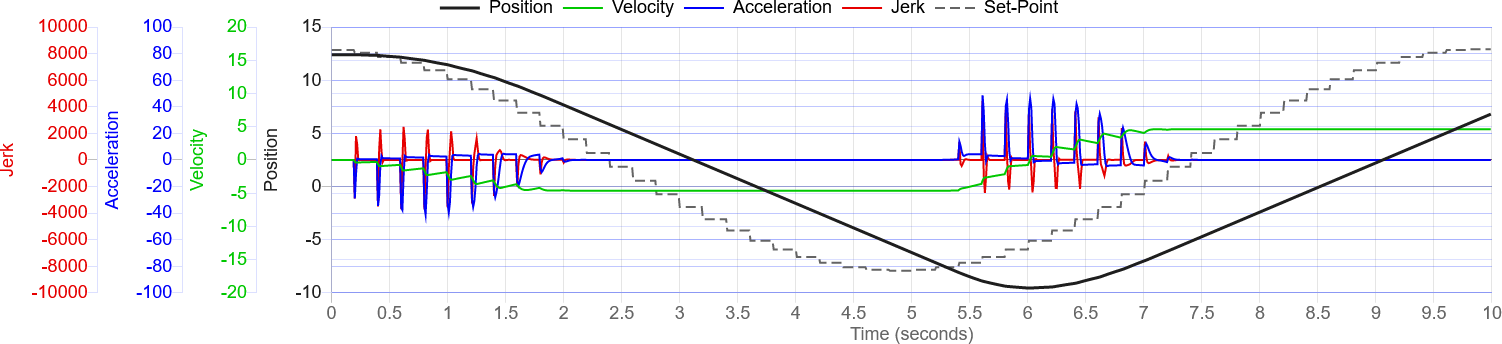}
		\caption{Output of Filter \xAT using the \textit{circular} input signal $\Phi_C$.}
		\label{fig:filterA_circular}
	\end{subfigure}
	
	\begin{subfigure}[b]{1\textwidth}
		\includegraphics[width=\textwidth]{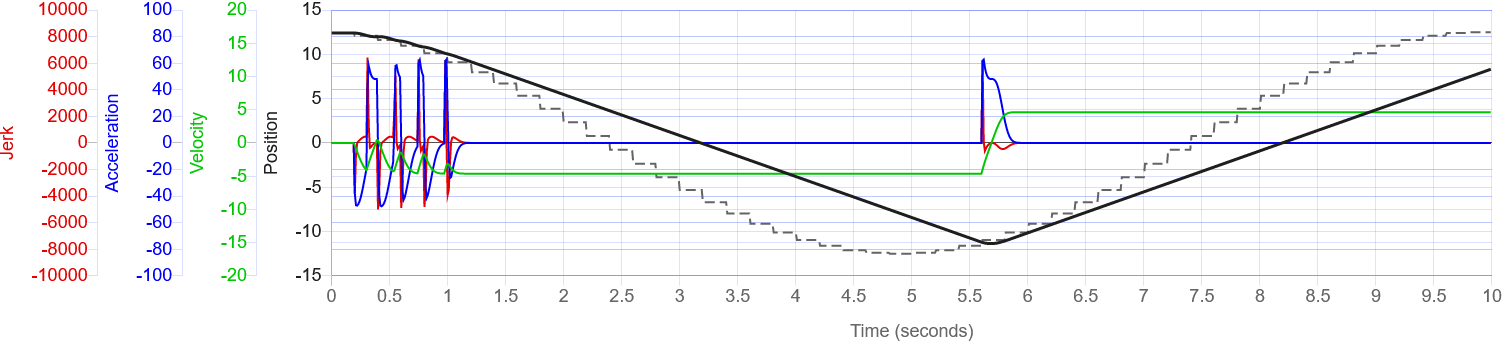}
		\caption{Output of Filter \xBT using the \textit{circular} input signal $\Phi_C$.}
		\label{fig:filterB_circular}
	\end{subfigure}
	
	\begin{subfigure}[b]{1\textwidth}
		\includegraphics[width=\textwidth]{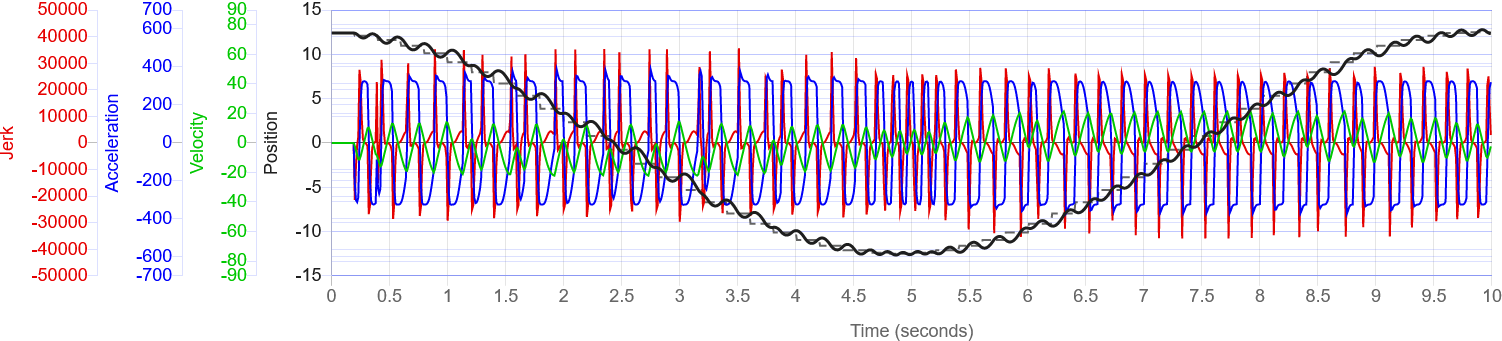}
		\caption{Output of Filter \xCT using the \textit{circular} input signal $\Phi_C$.}
		\label{fig:filterC_circular}
	\end{subfigure}
	
	\begin{subfigure}[b]{1\textwidth}
		\includegraphics[width=\textwidth]{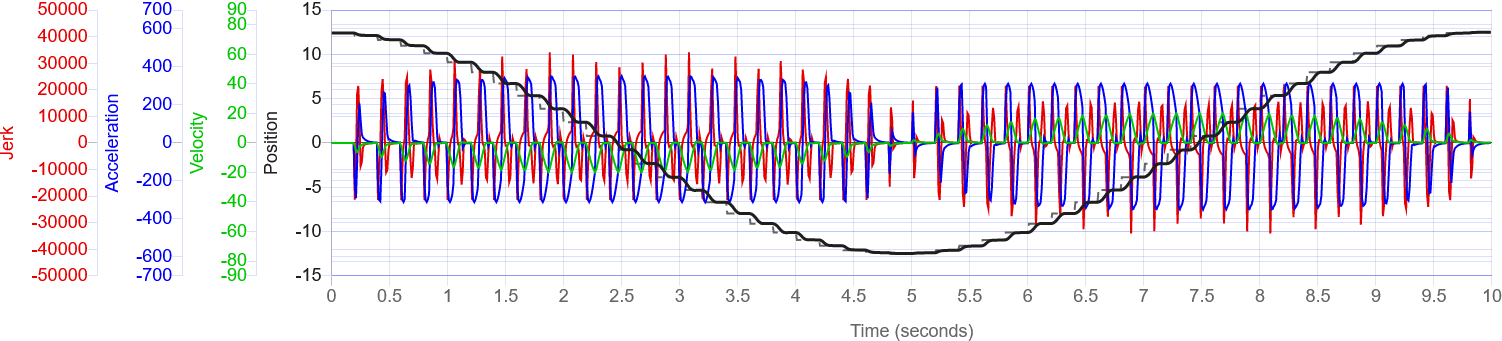}
		\caption{Output of Filter \xDT using the \textit{circular} input signal $\Phi_C$.}
		\label{fig:filterD_circular}
	\end{subfigure}

	\begin{subfigure}[b]{1\textwidth}
		\includegraphics[width=\textwidth]{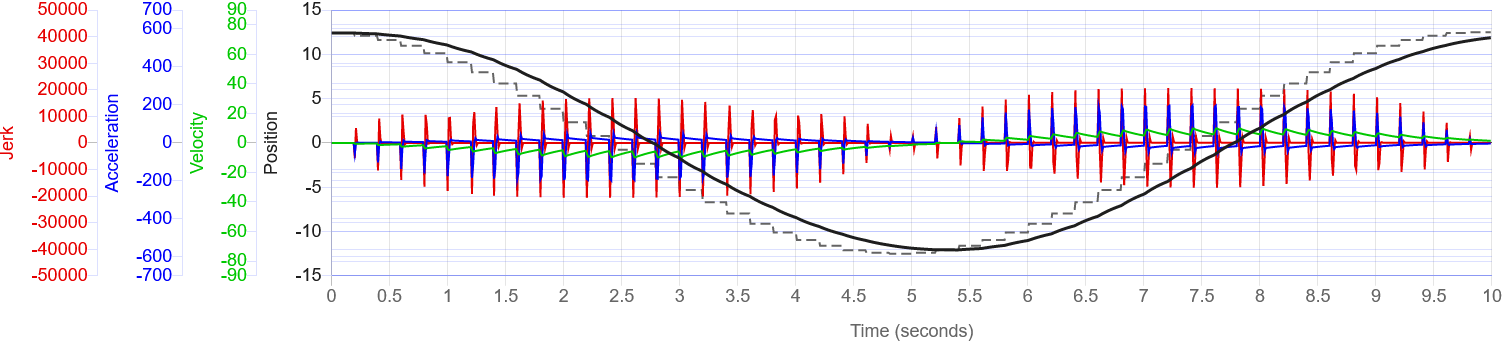}
		\caption{Output of Filter \xET using the \textit{circular} input signal $\Phi_C$.}
		\label{fig:filterE_circular}
	\end{subfigure}
	\caption{Comparison of the hyperparameter groups $A$, $B$, $C$, $D$ and $E$, using the \textit{circular} input signal $\Phi_C$.}
	\label{fig:filterCircular_compare}
\end{figure}

Finally, Figure \ref{fig:filterCircular_compare} shows the plot for each hyperparameter group $A$, $B$, $C$, $D$ and additionally $E$, using the \textit{circular} input trajectory $\Phi_C$.
These plots how each filter responds to an input signal that is continuously changing in small steps - which in some cases is actually a challenge.
Most of the remarks from the previous set of plots (Figure \ref{fig:filterRandom_compare}) also apply here.
However we have added the additional \xET filter as a version of \xDT with a lower \textit{responsive} parameter.
The purpose of this final set of plots is to show that not only will the selected character parameters depend on the intended shape and smoothness of the signal response, but must also consider how the input signal will be generated and fed to the filter. 

\subsection{Comments and Remarks}
Throughout this section we have presented the Nutty Motion Filter, which uses a composition of various transfer functions to allow an open-loop motion control system to smoothly interpolate and stabilize a given input signal even when that signal is highly discontinuous.
Various choices were made which however, should not fully enclose this filter as a sealed solution.
We have described the filter as various separate components, although one may implement a more optimized version by combining them in a different way.
In addition, various options may be considered regarding e.g. the limiter function. 
While the tanh-based limiter became our choice, it was a balanced decision, i.e., it provides steady, controllable results, with no tweaking required.
For other particular applications, it is important to emphasize that one may choose, explore and develop other types of limiter functions that better suite their requirements.

A final and important remark that must not be left uncommented is on the use of limiters with mobile robots that are operating in 2D trajectories.
While it may seem obvious for a person with a strong background in robotics, we want to clarify that in cases where the X and Y directions are controlled separately, they must typically be limited together, in order to provide a limiter on the robot's actual linear velocity (which is made up of x and y)
This means that given a 2D vector $[\dot{x}, \dot{y}]$, representing the velocities in both dimensions, one would have to calculate the magnitude of the resulting linear velocity vector, apply the limiter to that resulting vector only, and then proportionally saturate each of the two components, in order to limit both dimensions in a way that the resulting linear velocity does not exceed the specified limit.
The same principle would apply for any other degrees of freedom that jointly operate to perform 2D motion (or even more).

\section{Animation Tools for Social Robots}
\label{sec:tools}
When including creative artists such as animators into the development workflow, one of the first question that arises is the tools that the artists can use to author and develop expressive behaviour for the robot.
Typically those artists are commissioned to produce only pre-authored animation files that can be played back by the animation engine.
This may be achieved by either developing a custom-build GUI that allows them to directly develop on the system's tools, data types and configurations, or to allow the artists to use their familiar animation tools such as
3dsmax\footnote{\label{foot:3dsmax}\protect\url{https://www.autodesk.com/products/3ds-max/overview} \urlDate}, 
Maya\footnote{\label{foot:maya}\protect\url{https://www.autodesk.com/products/maya/overview} \urlDate}, 
SideFX Houdini\footnote{\label{foot:houdini}\protect\url{https://www.sidefx.com/products/houdini} \urlDate} or 
even the open-source Blender software\footnote{\label{foot:blender}\protect\url{https://www.blender.org} \urlDate}.
These existing animation packages allow to export animation files using general-purpose formats such as Autodesk FBX\footnote{\protect\url{https://www.autodesk.com/products/fbx/overview}  \urlDate}.
That requires the animation engine to support loading such formats, and to convert them into the internal representation of pre-animated motions.
Alternatively, and as most of those software support scriptable plug-ins, one may develop such a plug-in that allows to export the motion data into a format that is designed specifically for the animation engine.

Upon our introduction of the programmable animation engine, and of animation programs, it also becomes necessary to understand how the animators can contribute to such animation programming, alongside with their participation in the motion design.

\subsection{Animation Design Tools and Plug-ins}

We argue that for simple cases, developing an e.g. FBX import for the actual animation engine run-time environment is a good choice.
In this case the learning curve for the animators is almost inexistent, given that they will be working on their own familiar environment.
They will only need to adapt to specific technical directions such as maintaining a properly named and specific hierarchy for the joints and animatable elements, so that those can be properly imported later on.
When the nature of the project or application does not allow to rely on third-party, or proprietary software, then the only option may be to develop a custom animation GUI, which poses as the most complex and tedious one.
However our feeling has been that the creation of plug-ins for existing, third-party animation software provides a good balance between development effort, usability, user-experience and results.

\begin{figure}
	\centering
	\includegraphics[width=1\linewidth]{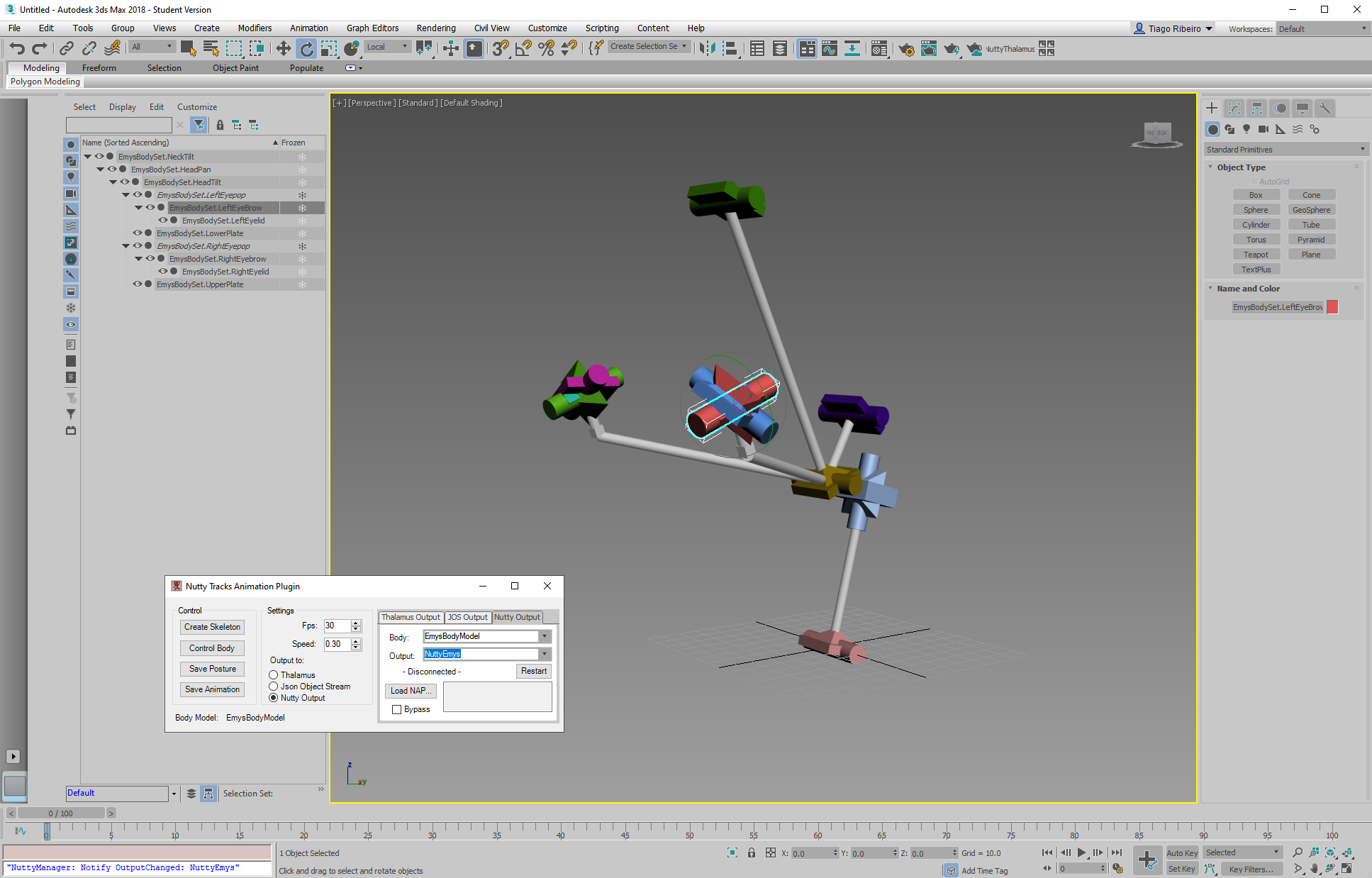}
	\caption{A screenshot of the Nutty Tracks plug-in for Autodesk 3dsmax, illustrating the skeletal animation rig created by the plug-in. An animator can generate this rig through the simple click of a button, and then use the plug-in to export the final animation to a Nutty-compatible animation file.}
	\label{fig:3dsmaxplugin}
\end{figure}

The creation of plug-ins for existing animation software includes the same advantages and requirements as in the first case, 
of developing an animation-format importer for the engine.
Animators will be familiar with the software, 
but may have to comply with certain technical directions in order for the plug-in to be able to properly fetch and export the motion data.
Figure \ref{fig:3dsmaxplugin} shows an example of the Nutty Tracks plug-in for Autodesk 3dsmax.
By having the EMYS embodiment already loaded in the Nutty Tracks engine, the plug-in can create an animatable rig for the robot, through the click of a single button, based on the embodiment's hierarchical specification including rotation axes, joint limits, etc.
Optionally it may even include the actual geometry of the robot for a more appealing experience.
From here on an animator may animate each of the gizmos that were created for each of the robot's animatable DoFs, using his or her typical workflow and techniques.

\begin{figure}
	\centering
	\includegraphics[width=1\linewidth]{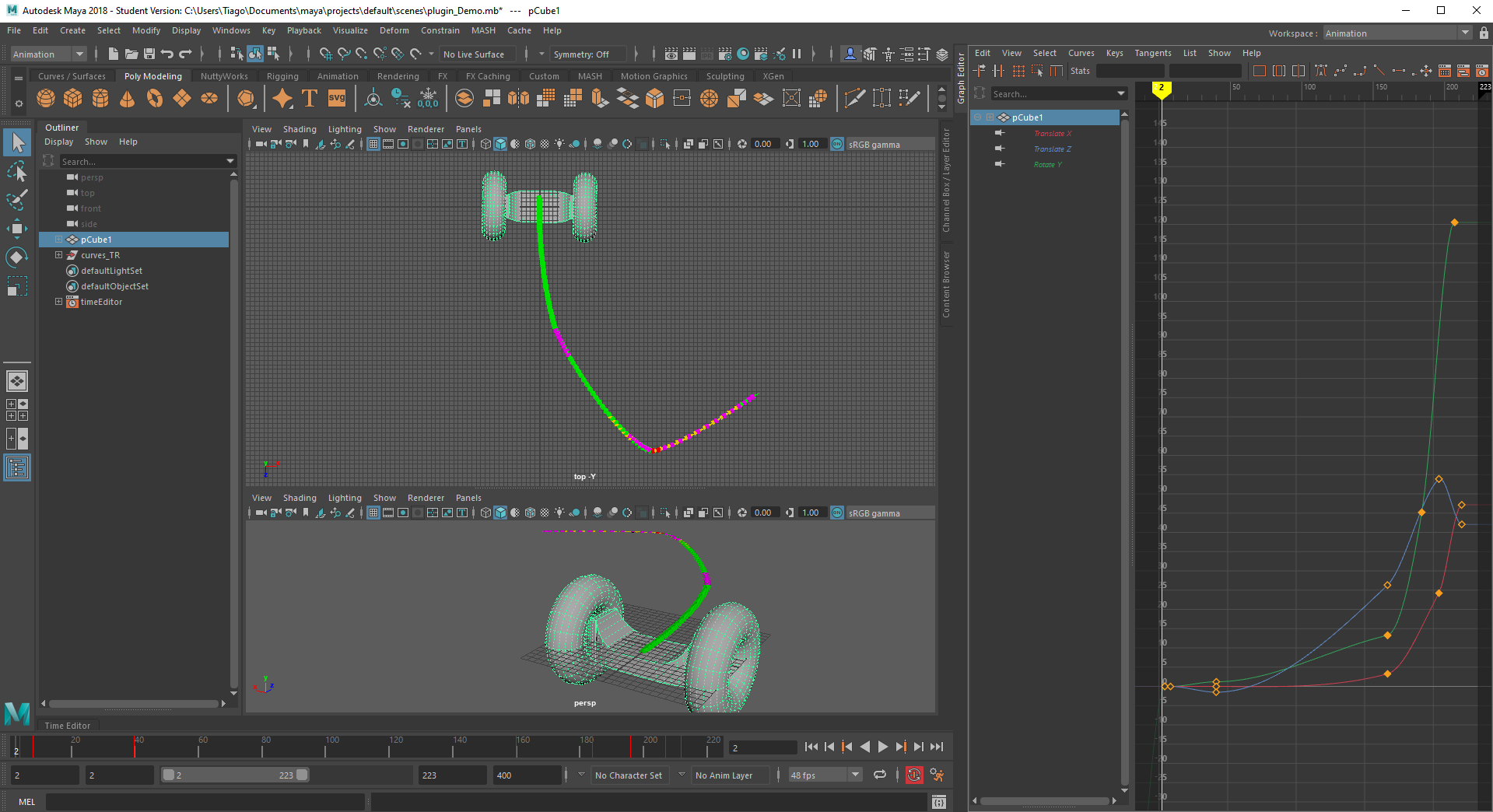}
	\caption{A screenshot illustrating the robot-animation trajectory-helper feature implemented through a plug-in into Autodesk Maya. This feature draws the motion trajectory as a path directly into the scene of the animation software, and highlights the points of the trajectory that break any of the robot's kinematic limits.}
	\label{fig:mayaplugin}
\end{figure}

However, the development of such a plug-in also allows to augment the creative development workflow, 
by adding visual guides directly into the viewports of the animation software, 
in order to represent technical constraints that are required specifically for robots, such as kinematic ones (e.g. velocity, acceleration, jerk limits).
Figure \ref{fig:mayaplugin} shows an example of a plug-in developed for Autodesk Maya, to show the \textit{trajectory-helper} of a given mobile robot platform, which highlights the points in the trajectory that break some of the robot's kinematic constraints. 
In this case, green means that the trajectory is within the limits, while the other colors each represent a certain limit violation, such as maximum velocity exceeded (orange), or maximum acceleration exceeded (pink) or maximum jerk exceeded (red).
Based on this visual guide, the animator knows where the trajectory must be corrected, and is able to readily preview how the fix will look like, while making any further adjustments to the motion in order to ensure the expected intention or expression is properly conveyed without exceeded the physical limits of the robot.

Other useful features may be to perform automatic correction of such constraints, while rendering the result directly within the animation environment, 
thus allowing the animators to fix the motion that results from enforcing such constraints, in a more interactive way.
From what we have gathered however, animators are typically not happy to have a tool that can change and control their animations.
Instead, the preferred option is to keep the artist-animated version of robot untouched by the plug-in, and to create an additional copy of the same robot model.
This copy, which we call the \textit{ghost}, will, in turn, not be animatable or even selectable by the animator, but instead, will be fully controlled by the plug-in.
Therefore, when the animator is previewing the playback of its animation, the plug-in will take that motion and process it in order to enforce the kinematic limits.
The resulting corrected motion is however applied only to the ghost, which therefore moves along with the animated robot.
If at any point, the animated motion did exceed the limits, the ghost will be unable to properly follow the animated model due to the signal saturation, 
which allows the animator to have a glimpse not only of where the motion is failing to comply with the limits, but also how it would look like if the limits were enforced.
In some cases the animator might actually feel that the result is acceptable, even if the originally designed motion would report limit violations on a trajectory-helper solution such as the one of Figure \ref{fig:mayaplugin}.
Note that in the case of the \textit{ghost-helper} technique, whenever the final animation is exported, it should be exported from the \textit{ghost} robot, which contains the corrected motion, and not the animated robot which does not.

In summary, the two major robot-animation features we have presented, and that can be provided through the use of animation software plug-ins, are the \textit{trajectory-helper}, as presented in Figure \ref{fig:mayaplugin}, and the \textit{ghost-helper}, described in the previous paragraph.
Depending on the animator's preferences, and the scripting capabilities of the animation environment, either one or both of the features can be used.
The ghost-helper seems to provide a more agile solution, as the animators aren't required to fix all the limit violations.
As long as they accept the motion provided through the ghost, the problem is considered to be solved, thus allowing them to complete animations quicker than using the trajectory-helper.
The trajectory-helper however allows an animator to better ensure that all the points of the trajectory are smooth and natural, and especially that the automatic correction (achieved e.g. through signal saturation) will not introduce any other unexpected phenomena. 
This feature is especially important when animating multiple robots\footnote{\protect\url{https://gagosian.com/exhibitions/2018/urs-fischer-play/} \urlDate}, to ensure that each of the individual auto-corrections do not place the robots in risk of colliding.

Without the ability to preview or at least evaluate the animated motion directly within the animation environment, the animators would need to jump between their software, and a custom software that solves and reports on those issues, while providing typically a mediocre or even no visual feedback on what is happening, and what needs to be fixed.
Besides making it a more complex workflow, that option also hinders and breaks the animator's own creative process.

Finally, an additional feature that can be developed through plug-ins for existing animation software is the ability to directly play the animations through the robot software or interactive pre-visualisation system.
This allows the animators to include testing and debugging into their workflow, by being able to see what will happen with their animations once they become used during interaction with the users and the environment.

\subsection{Animation Programming Tools}

Animators working with social robot application are required to learn some new concepts about how motion works on robots, in order to identify what can or cannot be done with such physical characters, as opposed to what they are used to do in fully virtual 3D characters.
Besides having to adapt to certain technical requirements when building their characters and animation rigs, they may also need to learn how to interact with some other pieces of software that will allow them to pre-visualize how the designed motion will look on the robots during actual interactions.

At some point the character animators will acquire so many new competencies and knowledge that they become actual \textit{robot animators}, an evolution of animators that besides being experts on designing expressive motion for robots, may also have learned other technical skills as part of the process.
One such skill is what we call animation programming.
The difference between a non-robot-programming animator, and a programming-robot animator is akin to the difference between a texture artist and a shader artist (or lighting artist) in the digital media industry.
The texture artist is a more traditional digital artist that composes textures that are \textit{statically} used within digital media.
A shader artist is able to take such textures, or other pattern-generators, and configure the shaders (i.e., programs) to adapt and change according to the environment parameters and applications.
The shaders are, in that sense, \textit{programmable} textures.
Similarly, and animation programs are \textit{programmable} animations.

Animation programs can, at a very basic level, be specified by some kind of mark-up code.
However, we believe the best option to be taking inspiration from currently existing tools.
Both Autodesk's Slate material editor\footnote{\protect\url{https://knowledge.autodesk.com/support/3ds-max/learn-explore/caas/CloudHelp/cloudhelp/2017/ENU/3DSMax/files/GUID-7B51EF9F-E660-4C10-886C-6F6ADE9E8F56-htm.html}  \urlDate}, and the Unreal Material Editor\footnote{\protect\url{https://docs.unrealengine.com/en-us/Engine/Rendering/Materials/Editor/Interface}  \urlDate}, are well-established artist-friendly shader-programming interfaces.
Houdini\footref{foot:houdini} is also known for its visual graph-based visual effects programming system.
Pure Data\footnote{\protect\url{http://puredata.info}  \urlDate} allows visual, sound and performance artists to develop their own musical instruments, visual effects processor, or any other kind of interactive system, using a visual block-graph paradigm that allows to simultaneously run the program while also allowing it to be composed, all in real-time.
As such, we argue for the creation of similar, artist-friendly, animation-programming editors.

These new animation programming tools can be built from scratch as standalone GUI application (e.g. Nutty Tracks), or using game development tools such as the Unity Engine\footnote{\protect\url{https://www.unity3d.com} \urlDate}, which allows for the scripting of new interface tools.
In this case, because a game engine such as Unity3D also provides 3d visualization and animation tools, it could be extended by an animation programming tools in order to become a fully-fledged animation designing, programming and pre-visualization tool.

\begin{figure}
\centering
\includegraphics[width=1\linewidth]{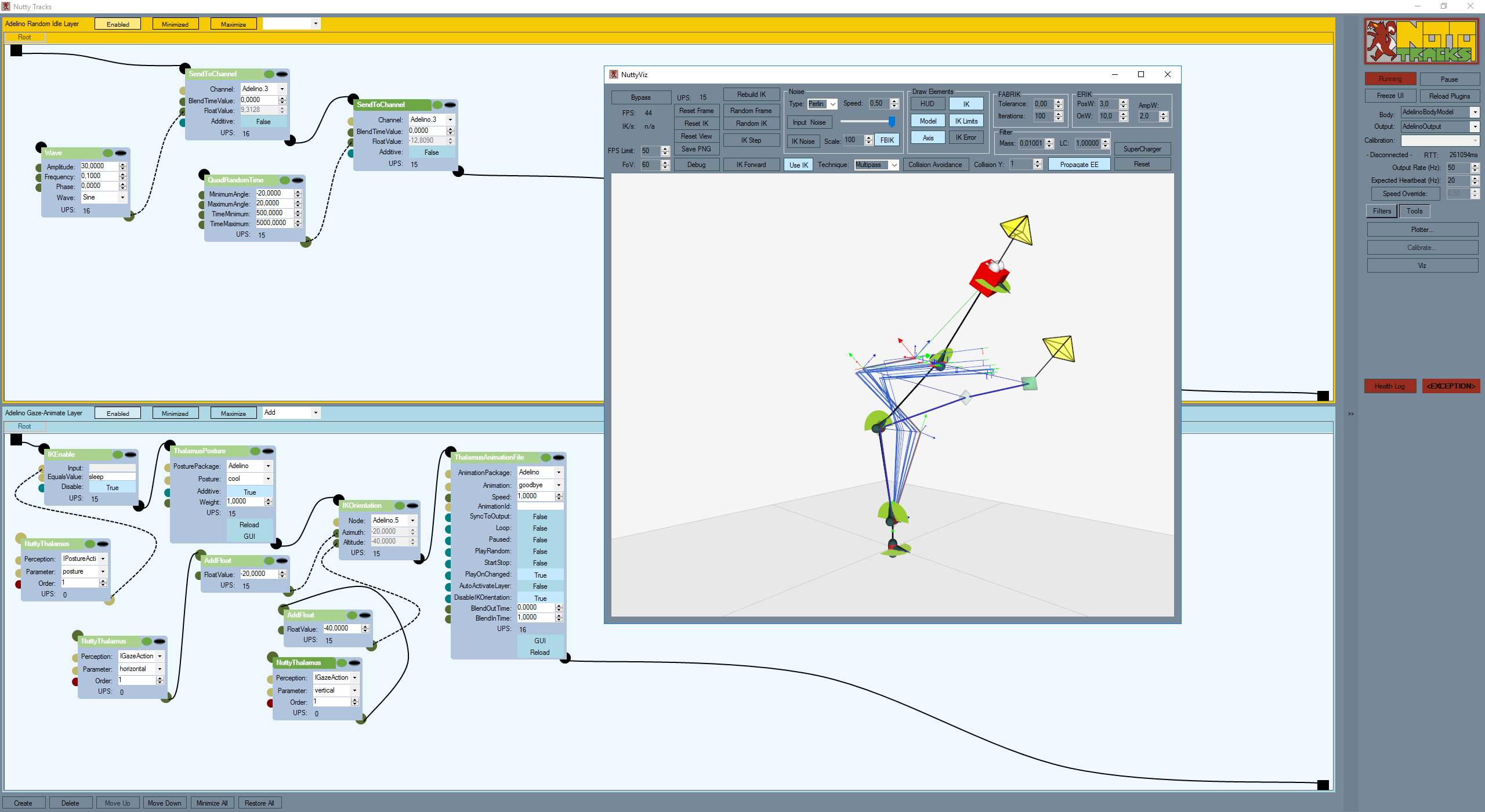}
\caption{The Nutty Tracks GUI, used for animation programming in a multi-layer, multi-block visual editor. Within the figure, we see several different Animation Blocks. Some of them take PAFs as input and/or output (black connection points), while others provide colored connection points for single-dimension signals, which are color-coded depending on the type of signal they carry (e.g. floats, integers, strings, etc..). It additionally includes an Inverse Kinematics interactive visualizer which allows an animator to tweak some of its parameters, in order to adjust the generated motion to the robot's kinematic capabilities.}
\label{fig:nutty_screenshot2}
\end{figure}

Nutty Tracks provides an example of how such an editor may be presented\footnote{\protect\url{https://vimeo.com/67197221}\urlDate}.
Its programmable animation GUI is also shown in Figure \ref{fig:nutty_screenshot2}.
It was conceptualized to allow an animator to load and pre-visualize how animations and expressive postures designed in another software (e.g. 3dsmax) will look like when procedural layers of motion are added, such as ones that generate idle-behaviour, user-face tracking, or inverse kinematics.
Such output motion is processed by the Level 3 APU in Nutty Tracks, and could not be properly visualized within the typical animation design software.
However the process of composing and tweaking the animation program using animation blocks follows a workflow that is similar to the one found on other artist-friendly applications that inspired us.

Despite such effort, it will still be the case that such an animation program editor will pose as a truly novel tool for the animators, with a steep learning curve.
An animator may e.g. be familiar with the concept of an animation layer, which does not match the one used in the visual animation program editor.
The idea of composing programmable animations using operator- and generator-blocks may have a paralell with certain motion control nodes found in some animation software, but the way they are used and composed may also not seem intuitive or obvious for the traditional 3D animator.
As such, it is required that these tools are developed with a user-centered design perspective, in close collaboration with the end-users, who are the actual animators, and to ensure the GUI provides an understandable translation between the animator's mindset, and the underlying mechanics and pipeline of the animation engine.

Throughout this paper we have presented our perspective on how robot animation can become an integral process in the development of social robots, based on theories and practices that have been created through the last century, in the fields of both traditional and 3D computer-graphics character animation.
We have introduced and described the 12 principles of robot animation, as a foundation that aims at aiding the transfer of the previous character animation practices into the new robot animation ones.
In the traditional character animation workflow, characters and their motions are designed to be faithfully played-back on screens.
One of the most relevant steps in this transition is the ability to not only design, but also program how animations should be shaped, merged and behave during interaction with human users.
We must therefore introduce new techniques and methods that allow such artistically crafted animations to become not only interactive (such as in video-games), but to interact in the real world, with real users.
Such new techniques and methods will be provided by new tools and workflows that are designed with artists in mind, and that aim at the technical requirements imposed by robotics.
One such technique is the development and use of the Programmable Animation Engines based on the Nutty APUs.
Upon establishing such techniques, such artists may become a new type of animators which we call robot animators.
These are not only experts in traditional character animation, but also know how animation must be designed for robots, and how it should be adapted and shaped during real-world interactions.
By following and implementing such paradigms, we expect that social robots may become more akin to animated characters, in a sense that they are able to interact with users in social settings while properly exhibiting the illusion of life.

\section*{ACKNOWLEDGEMENTS}
This work was supported by national funds through FCT - Funda\c{c}\~{a}o para a Ci\^{e}ncia e a Tecnologia with references UID/CEC/50021/2019 and SFRH/BD/97150/2013.

\bibliographystyle{acm}
\bibliography{main}

\end{document}